%% file: main.tex
\documentclass[letterpaper]{article} %
\usepackage{aaai2027}  %
\usepackage[hyphens]{url}  %
\usepackage{graphicx} %
\usepackage{natbib}  %
\usepackage{caption} %
\usepackage{algorithm}
\usepackage{algorithmic}
\usepackage{tabularx}
\usepackage{enumitem}
\usepackage{subcaption}
\usepackage{multirow}
\usepackage{tablefootnote}
\usepackage{threeparttable}

\usepackage{newfloat}
\usepackage{listings}
\DeclareCaptionStyle{ruled}{labelfont=normalfont,labelsep=colon,strut=off} %
\floatstyle{ruled}
\newfloat{listing}{tb}{lst}{}
\floatname{listing}{Listing}

\usepackage{xcolor}

\newcommand{\ASim}{\texttt{A-Sim}}
\newcommand{\AXBar}{\texttt{A-XBAR}}
\newcommand{\ATWI}{\texttt{A-TWI}}
\newcommand{\WPRNG}{\texttt{W-PRNG}}
\newcommand{\AVac}{\texttt{A-Vac}}

\usepackage{booktabs}

\usepackage{amsmath}
\usepackage{amssymb}
\usepackage{bm}
\usepackage{multirow}
\usepackage{mathtools}

\usepackage[disable]{todonotes}
\usepackage{dblfloatfix}
\usepackage{orcidlink}

\nocopyright %

\title{Constrained Co-Design for Photonic Bayesian Neural Networks}
\author{
    Hendrik Borras\equalcontrib\textsuperscript{\rm 1}\orcidlink{0000-0002-2411-2416},
    Xiao Wang\equalcontrib\textsuperscript{\rm 1}\orcidlink{0009-0001-5064-7940},
    Bernhard Klein\textsuperscript{\rm 1}\orcidlink{0000-0003-0497-5748},
    Robin Janssen\textsuperscript{\rm 1}\orcidlink{0009-0006-4356-3059},\\
    Frank Brückerhoff-Plückelmann\textsuperscript{\rm 2}\orcidlink{0009-0006-5748-7069},
    Wolfram Pernice\textsuperscript{\rm 2}\orcidlink{0000-0003-4569-4213},
    Holger Fröning\corresponding\textsuperscript{\rm 1}\orcidlink{0000-0001-9562-0680}
}
\affiliations{

    \textsuperscript{\rm 1}Hardware and Artificial Intelligence Lab,\\
    Institute of Computer Engineering,\\Heidelberg University, Germany

    \textsuperscript{\rm 2}Neuromorphic Quantumphotonics\\
    Kirchhoff-Institute for Physics,\\Heidelberg University, Germany

    \{hendrik.borras,xiao.wang,bernhard.klein,robin.janssen,holger.froening\}@ziti.uni-heidelberg.de\\
    \{frank.brueckerhoff-plueckelmann,wolfram.pernice\}@kip.uni-heidelberg.de

}

\begin{document}

\maketitle

\input{sections/01_abstract.tex}

\input{sections/02_introduction.tex}

\input{sections/03_background.tex}

\input{sections/03b_operator.tex}

\input{sections/05_experiments.tex}

\input{sections/05a_coupled.tex}

\input{sections/06_discussion.tex}

\section*{Acknowledgments}
The authors acknowledge support by the state of Baden-Württemberg through bwHPC
and the German Research Foundation (DFG) through grant INST 35/1597-1 FUGG.

\bibliography{references}

\clearpage
\onecolumn 
\appendix
\input{sections/07_appendix.tex}

\end{document}

%% file: sections/01_abstract.tex
\begin{abstract}

Classical neural networks frequently produce overconfident predictions on ambiguous or out-of-distribution (OOD) data, a liability that grows with each AI system deployed in safety-critical real-world scenarios.
Bayesian neural networks (BNNs) provide a principled framework for uncertainty-aware prediction by replacing deterministic parameters with probability distributions, but repeated sampling increases latency, memory traffic, and energy consumption.
Photonic probabilistic computing offers a promising alternative by exploiting intrinsic optical stochasticity for fast and parallel sampling.
However, photonic BNNs are not ideal samplers: analog constraints on quantization, programming error, dynamic range, and representable mean and variance restrict the variational families that can be implemented in hardware.

In this work, we study which hardware-imposed constraints limit scalable photonic BNN inference, how these constraints can be represented, and which ranges can be tolerated by photonic BNNs beyond small proof-of-concept networks.
We formulate photonic BNN inference as constrained stochastic variational inference and perform a systematic ablation study over stochasticity location, stochasticity modality, quantization, programming error, and mean/variance bounds.
From these results, we derive concrete co-design guidelines that distinguish hardware constraints that can be compensated by training from those requiring hardware or architecture intervention.
We validate these guidelines under coupled, hardware-realistic constraints on Dirty-MNIST, CIFAR-10, and CINIC-10, using Fashion-MNIST and SVHN as OOD benchmarks, showing that hardware-aware training recovers predictive performance and uncertainty quality whenever the required variational family remains representable, whereas violations of representational limits require targeted hardware modifications.

\end{abstract}

%% file: sections/02_introduction.tex
\section{Introduction}

\todo[inline, color=yellow]{regarding wording:  standard deviation, variance and scale}
\todo[inline, color=yellow]{Should we introduce TWI as alias for the reference? We refer to that paper so many times. -> HB: Yes, we should do that basically for all references we quote in table 1.}

Reliable machine learning requires more than accurate predictions---it requires knowing when those predictions are unreliable. 
In autonomous systems, medical diagnosis, scientific instrumentation, and edge sensing, this ability is essential for safe and trustworthy decision making. 
Reliable uncertainty estimation, however, remains challenging, particularly under distribution shift and for out-of-distribution (OOD) inputs~\cite{ovadia2019trust}. 
Consequently, uncertainty-aware models are evaluated not only by predictive performance, but also by the quality of their uncertainty estimates, including calibration, OOD detection, and the separation of different sources of uncertainty~\cite{nado2021uncertainty}.

Bayesian neural networks (BNNs) provide a principled framework for uncertainty-aware learning by replacing deterministic parameters with probability distributions~\cite{mackay1992practical}. 
Practical implementations rely on approximate inference methods such as Monte Carlo Dropout~\cite{gal2016mcdo}, Deep Ensembles~\cite{lakshminarayanan2017deepensembles}, and stochastic variational inference (SVI)~\cite{graves2011practical,blundell2015}. 
While these approaches produce high-quality uncertainty estimates, they require repeated stochastic evaluations during inference, increasing latency, energy consumption, and memory traffic on conventional digital hardware. 
These computational costs are especially problematic in real-time, safety-critical, and resource-constrained applications, where reliable uncertainty estimates are often most valuable.

Physical probabilistic computing offers an attractive alternative by treating intrinsic device stochasticity as a computational resource rather than a nuisance, eliminating the need to emulate randomness digitally.
Among emerging probabilistic computing technologies, photonics is particularly attractive because high-bandwidth optical signals enable stochastic sampling and processing at very high rates.
Recent demonstrations have shown the feasibility of photonic Bayesian inference (see Section~\ref{sec:related:photonic_hardware}), establishing photonic hardware as a platform for Bayesian neural computation~\cite{brueckerhoff2025probabilistic}.

However, photonic Bayesian hardware cannot be designed as an ideal sampler.
Like deterministic analog accelerators, it requires hardware-aware optimization to compensate for device non-idealities through techniques such as calibration, surrogate modeling, or in situ learning~\cite{rasch2023hardware, wright2022deep, pai2023in}.
Its probabilistic nature, however, introduces an additional challenge: hardware constraints directly affect the probabilistic representation.
Consequently, \textbf{photonic Bayesian hardware does not merely accelerate SVI-based BNNs; it constrains the variational distributions that can be represented in hardware.}
BNN architecture, variational parameterization, training procedure, and hardware settings must therefore be co-designed to preserve uncertainty quality under physical constraints.

Prior work has shown that photonic stochasticity can support uncertainty-aware inference, but network architectures and hardware settings were largely pre-determined by the hardware.
It therefore remains unclear which hardware constraints are benign, which can be compensated through training, and which fundamentally degrade uncertainty quality.
Moreover, the combined effect of these constraints is poorly understood: parameter range limits, quantized activations, and analog programming errors may each be tolerable in isolation, yet become detrimental when combined.
Without a systematic co-design methodology, photonic BNNs risk becoming impressive hardware demonstrations whose uncertainty quality is difficult to predict, reproduce, or scale.

This raises \textbf{three guiding questions} for the co-design of photonic BNNs:

\begin{enumerate}[label={\textbf{Q\arabic*}},leftmargin=*,align=left]
    \item Which hardware-imposed constraints limit predictive accuracy and uncertainty quality in large-scale photonic BNNs?

    \item Which ranges of quantization, programming error, and representable mean and variance constraints remain tolerable across stochasticity locations and modalities?

    \item Which constraints can be compensated by training or parameterization, and which require hardware or architecture intervention?
\end{enumerate}

To address these questions, we study the co-design of photonic BNNs under hardware constraints by treating the photonic processor as a physical implementation of a \textbf{constrained variational family}.
Our contributions are fourfold:
\begin{enumerate}

\item We formulate photonic SVI-based BNN inference as constrained variational inference, relating hardware constraints to the representable variational distributions, variational parameterization, sampling process, and neural operators.
\item We systematically analyze the sensitivity of uncertainty quality to individual hardware constraints, identifying which can be compensated by training and which fundamentally degrade performance.
\item We derive practical co-design guidelines for photonic BNNs, distinguishing between constraints that can be mitigated via training and those requiring hardware or architectural intervention.
\item We validate the proposed guidelines through an integration study based on a state-of-the-art time-wavelength-interleaving photonic architecture, demonstrating their relevance to a realistic photonic BNN implementation.
\end{enumerate}

Together, these results establish practical design principles for the co-design of variational families and scalable photonic Bayesian hardware, enabling reliable uncertainty estimation under realistic hardware constraints.

%% file: sections/03_background.tex
\section{Background and Related Work}

\subsection{Bayesian Neural Networks}

BNNs extend conventional neural networks by introducing stochastic weights or stochastic activations, thereby learning a distribution over models rather than a single deterministic parameter configuration~\cite{jospin2022handsOnBnns}.

Let $\mathcal{D}$ denote the observed data and $\mathbf{z}$ the latent variables representing all stochastic network components. Bayesian inference seeks the posterior distribution $p(\mathbf{z}\mid\mathcal{D})$, which is generally intractable for neural networks due to their high dimensionality and nonlinear structure.

\subsubsection{Stochastic Variational Inference}
Variational inference addresses this intractability by introducing a tractable variational distribution \(q_\phi(\mathbf{z})\) to approximate the true posterior $p(\mathbf{z} \mid \mathcal{D})$. The parameters of \(q_\phi(\mathbf{z})\), for example mean and standard deviation, are optimized via stochastic gradient descent by 
maximizing the Evidence Lower Bound (ELBO)~\cite{blei2017variational}:
\begin{equation}
	\mathrm{ELBO}(q) = \mathbb{E}_{q_{\phi}(\mathbf{z})}\!\left[\log p(\mathcal{D} \mid \mathbf{z})\right] - \mathrm{KL}\!\left(q_{\phi}(\mathbf{z}) \,\|\, p(\mathbf{z})\right),
\end{equation}
where the first term encourages the latent variables to explain the observed data, and the second term regularizes the variational distribution toward the prior distribution $p(\mathbf{z})$. 

Stochastic Variational Inference (SVI) maximizes the ELBO using stochastic gradient descent on mini-batches, making Bayesian inference scalable to modern datasets.~\cite{hoffman2013svi, jospin2022handsOnBnns}.

A distinctive property of SVI is that the variational distribution $q_{\phi}(\mathbf{z})$ is parameterized in closed form, e.g., as a Gaussian with mean $\mu$ and standard deviation $\sigma$ for each stochastic component. 
This parameterization opens a direct path to analog hardware deployment: by configuring a controllable noise source to match the learned $\mu$ and $\sigma$, the hardware's intrinsic physical noise instantiates the variational distribution, eliminating the need for pseudo-random number generation during inference. This direct correspondence between SVI parameters and controllable analog noise motivates the present work. Within this SVI framework, we empirically compare the suitability of weight and activation stochasticity for analog hardware implementation.

\begin{table*}[t]
    \centering
    \caption{
        Definitions of representative hardware constraints in photonic BNN accelerators and their realization in \ATWI.
    }
    \label{tab:constraints}

    \footnotesize
    \renewcommand{\arraystretch}{1.0}
    \setlength{\tabcolsep}{3pt}

    \begin{tabularx}{\textwidth}{
        @{}
        >{\raggedright\arraybackslash}p{0.15\textwidth}
        >{\raggedright\arraybackslash}X
        >{\raggedright\arraybackslash}p{0.21\textwidth}
        >{\raggedright\arraybackslash}p{0.21\textwidth}
        @{}
    }
        \toprule
        \textbf{Constraint}
        & \textbf{Mathematical formulation}
        & \textbf{\ATWI{} configuration}
        & \textbf{Relevant Systems}
        \\
        \midrule

        Input quantization
        & $x_{\mathrm{QAT}}
          =\operatorname{dequantize}(\operatorname{quantize}(x))$
        & 8-bit
        & \AVac, \AXBar, \ATWI
        \\
        \addlinespace[2.5pt]

        Constrained mean
        &
        \begin{tabular}[c]{@{}l@{}}
            Signed: $\bm{\mu}\in[-\mu_{\max},\mu_{\max}]$\\
            Unsigned: $\bm{\mu}\in[0,\mu_{\max}]$
        \end{tabular}
        & $\bm{\mu}\in[-1,1]$
        &
        \begin{tabular}[c]{@{}l@{}}
            Signed: \ATWI\\
            Unsigned: \AXBar, \WPRNG
        \end{tabular}
        \\
        \addlinespace[2.5pt]

        Constrained std
        &
        \begin{tabular}[c]{@{}l@{}}
            Upper bound: $\bm{\sigma}\in[0,\sigma_{\max}]$\\
            Lower bound: $\bm{\sigma}\geq\sigma_{\min}$
        \end{tabular}
        &
        \begin{tabular}[c]{@{}c@{}}
            Mul: $\bm{\sigma}\in[0.234,0.468]$\\
            Add: $\bm{\sigma}\in[0.520,0.789]$
        \end{tabular}
        &
        \begin{tabular}[c]{@{}l@{}}
            \WPRNG, \AXBar\\
            \ATWI
        \end{tabular}
        \\
        \addlinespace[2.5pt]

        Programming error
        &
        \begin{tabular}[c]{@{}l@{}}
            $\hat{\mu}=\mu+\epsilon_\mu$,
            $\epsilon_\mu\sim\mathcal N(0,\delta_\mu)$\\
            $\hat{\sigma}=\sigma+\epsilon_\sigma$,
            $\epsilon_\sigma\sim\mathcal N(0,\delta_\sigma)$
        \end{tabular}
        &
        \begin{tabular}[c]{@{}l@{}}
            $\delta_\mu=0.11$\\
            $\delta_\sigma=0.053$
        \end{tabular}
        & \WPRNG, \AXBar, \ATWI
        \\
        \addlinespace[2.5pt]

        Distribution type
        &
        \begin{tabular}[c]{@{}l@{}}
            Non-Gaussian sampling, e.g. Bose--Einstein statistics
        \end{tabular}
        & Gaussian
        & \AXBar
        \\
        \addlinespace[2.5pt]

        Sample correlation
        & Correlation between related samples, $\rho(\epsilon_t,\epsilon_{t+k})\neq 0$
        & Negligible after tuning
        & \ASim, \AXBar, \ATWI
        \\

        \bottomrule
    \end{tabularx}
\end{table*}

\subsubsection{Uncertainties in BNNs}

The learned stochasticity of BNNs defines a predictive distribution, which is obtained empirically through repeated sampling. 
This distribution enables the estimation of predictive uncertainty, commonly decomposed into aleatoric uncertainty arising from the data and epistemic uncertainty arising from uncertainty in the model~\cite{kendall2017uncertainties}.

For classification tasks, the total predictive uncertainty is quantified by the Shannon entropy~\cite{shannon1948entropy} of the predictive distribution, while the expected entropy of the individual predictions (Softmax Entropy) captures the aleatoric component. 
Both quantities can be estimated from the same predictive samples, and their difference yields the Mutual Information (MI), which quantifies epistemic uncertainty~\cite{Depeweg2018}. 
This decomposition enables BNNs to distinguish between data uncertainty and model uncertainty, making MI a natural signal for out-of-distribution detection.

\subsubsection{Evaluating uncertainty quantification performance of BNNs}

Standard deep learning benchmarks remain applicable to BNNs for evaluating in-distribution (ID) performance, but dedicated protocols are required to evaluate uncertainty quantification (UQ). A widely adopted benchmark for UQ is Dirty-MNIST~\cite{Mukhoti2022dirtyMNIST}. This dataset extends MNIST~\cite{lecun2010mnist} with Ambiguous-MNIST, which contains ambiguous digit images belonging plausibly to two classes, enabling the evaluation of aleatoric uncertainty. To measure epistemic uncertainty, Dirty-MNIST can be paired with Fashion-MNIST~\cite{xiao2017fashionMnist} as an OOD dataset. Performance is typically measured by the area under the receiver operating characteristic curve (AUROC), where the Softmax Entropy serves as the classification score for separating ambiguous from ID samples, and the MI for separating OOD from ID samples.

While Dirty-MNIST provides a comprehensive uncertainty benchmark, its ID classification task is relatively simple. To evaluate the scalability of BNNs, more challenging datasets such as CIFAR-10~\cite{CIFAR10} and CINIC-10~\cite{CINIC10} are commonly used for ID evaluation. Epistemic uncertainty can then be assessed by comparing against an OOD dataset, for which we use Street View House Numbers (SVHN)~\cite{SVHN}. Unfortunately, creating large-scale benchmarks with controlled aleatoric ambiguity is more challenging, and aleatoric uncertainty is therefore generally not evaluated in such settings.

\subsection{Related Work}
\label{sec:related:photonic_hardware}

Physical probabilistic computing exploits intrinsic device stochasticity as a computational resource rather than treating it as an imperfection to be suppressed.

A variety of hardware platforms have been proposed for probabilistic neural computation, including memristive Bayesian neural networks~\cite{Bonnet2023}, magnetic-tunnel-junction-based probabilistic in-memory computing~\cite{liu2022mtj}, and probabilistic synapses based on emerging two-dimensional materials~\cite{sebastian2022probabilistic}.

Probabilistic photonic hardware is particularly attractive because high-bandwidth optical signals enable stochastic sampling and processing at very high rates. 
Existing approaches can be broadly grouped according to whether stochasticity is applied to the activations or the weights.

A first class of approaches realizes stochasticity in the activations.
\citet[\ASim]{9931937} present a largely simulation-based photonic accelerator in which an amplified spontaneous emission (ASE) light source generates Gaussian noise while micro-ring resonators (MRRs) program the stochastic parameters.
Similarly, \citet[\AXBar]{brckerhoffplckelmann2024probabilistic} implement activation-space SVI on a photonic crossbar array using ASE-based stochasticity.
More recently, \citet[\ATWI]{brückerhoffplückelmann2025uncertaintyreasoningphotonicbayesian} replace the crossbar with a time-wavelength interleaving~\citep[TWI,][]{TWI_orig} architecture and demonstrate end-to-end BNN inference.
In this accelerator, the variational parameters are associated with the weights, but the resulting stochasticity is applied at the activations.
The \ATWI~accelerator serves as the reference hardware platform in the present work.
Finally, \citet[\AVac]{PBM_vacuum} propose a photonic probabilistic activation function based on quantum vacuum noise.
Since the activation probabilistically collapses to one of two output states without explicit probabilistic parameters, we interpret the approach as a hardware accelerator for a binarized variant of Monte Carlo Dropout~\cite{gal2016mcdo} rather than an implementation of SVI.

In contrast, \citet[\WPRNG]{PBNNS} apply stochasticity to the weights using a photonic random number generator cell for integration into a photonic crossbar array.
Gaussian-distributed weight samples are generated by programming the mean primarily with a Mach--Zehnder interferometer (MZI) and the standard deviation with an MRR, enabling SVI-based BNN inference across several benchmark tasks.

Adjacent to this, a small body of work exists on accelerating or compressing BNNs on digital hardware.
Here, for example, quantization-aware training \cite{borras2025uncertaintypreservingqbnnsmultilevelquantization}
or post-training quantization \cite{ferianc2021quantisation,subedar2021quantization,lin2023quantization}
is used to significantly save memory and compute.

\begin{figure*}[t]
	\centering
	\includegraphics[width=0.8\textwidth]{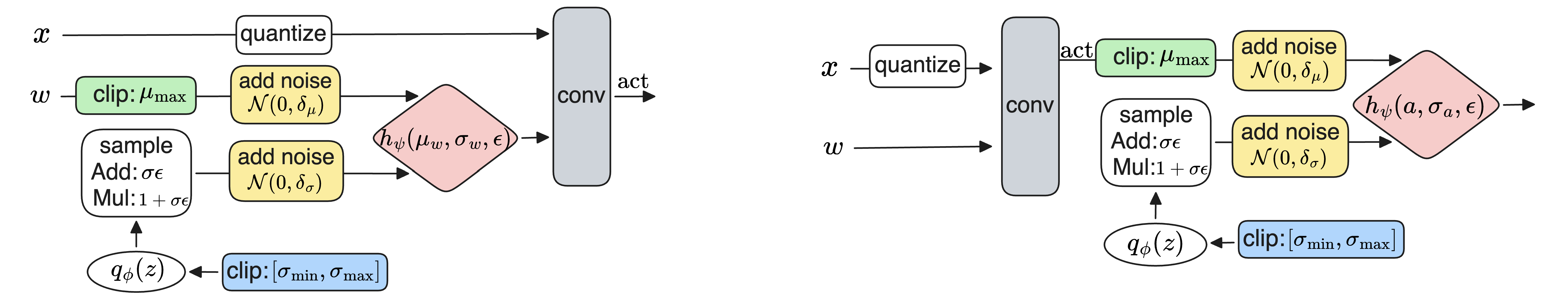}
	\caption{
		Schematic implementation of the two Pyro operators used to simulate photonic hardware.
		Left: Weight space locality; Right: Activation space locality.
		Here $x$ and $w$ represent the input activations and weights respectively.
		All other parameters follow their definitions in Table \ref{tab:constraints}.
	}
	\label{fig:method:operator_overview}
\end{figure*}

%% file: sections/03b_operator.tex
\begin{table*}[t]
    \centering
    \caption{
        Stable constraint ranges. 
        Each hardware constraint is varied independently in ResNet-18 on CIFAR-10 (Figure~\ref{fig:mean-constraint-comparison}).
		Columns W-Add, W-Mul, A-Add, and A-Mul report the ranges for which the OOD performance loss remains below \(5\%\).
    }
    \label{tab:ablation}

    \footnotesize
    \renewcommand{\arraystretch}{1.08}
    \setlength{\tabcolsep}{3pt}

    \begin{tabularx}{\textwidth}{
        @{}
        >{\raggedright\arraybackslash}p{0.22\textwidth}
        >{\centering\arraybackslash}p{0.20\textwidth}
        *{4}{>{\centering\arraybackslash}X}
        @{}
    }

		\toprule
		\textbf{Constraint}
		& \textbf{Experimental range}
		& \textbf{W-Add}
		& \textbf{W-Mul}
		& \textbf{A-Add}
		& \textbf{A-Mul}
		\\
		\midrule

        Input quantization \(b\)
        & \(\{16,8,7,6,5,4,3,2\}\)
        & \(b\ge4\)
        & \(b\ge4\)
        & \(b\ge4\)
        & \(b\ge4\)
        \\
        \addlinespace[1.5pt]

        Signed mean bound \(\mu_{\max}\)
        & \(10^{-4}<|\mu_{\max}|<10^{4}\)
        & \(|\mu_{\max}|>0.05\)
        & \(|\mu_{\max}|>0.003\)
        & \(|\mu_{\max}|>2100\)
        & \(|\mu_{\max}|>9400\)
        \\
        \addlinespace[1.5pt]

        Std lower bound \(\sigma_{\min}\)
        & \(10^{-3}<\sigma_{\min}<10^{2}\)
        & \(\sigma_{\min}<2.0\)
        & \(\sigma_{\min}<0.45\)
        & \(\sigma_{\min}<0.9\)
        & \(\sigma_{\min}<0.45\)
        \\
        \addlinespace[1.5pt]

        Std upper bound \(\sigma_{\max}\)
        & \(10^{-12}<\sigma_{\max}<10\)
        & \(\sigma_{\max}>0.03\)
        & \(\sigma_{\max}>0.008\)
        & \(\sigma_{\max}>0.11\)
        & \(\sigma_{\max}>0.008\)
        \\
        \addlinespace[1.5pt]

        Mean programming noise \(\delta_{\mu}\)
        & \(10^{-3}<\delta_{\mu}<10^{4}\)
        & \(\delta_{\mu}<2.1\)
        & \(\delta_{\mu}<0.9\)
        & \(\delta_{\mu}<80\)
        & \(\delta_{\mu}<220\)
        \\
        \addlinespace[1.5pt]

        Std programming noise \(\delta_{\sigma}\)
        & \(10^{-3}<\delta_{\sigma}<10^{4}\)
        & \(\delta_{\sigma}<2.0\)
        & \(\delta_{\sigma}<0.32\)
        & \(\delta_{\sigma}<180\)
        & \(\delta_{\sigma}<0.42\)
        \\

        \bottomrule
    \end{tabularx}
\end{table*}

\section{Photonic Hardware as a Constrained Stochastic Operator}

To describe different photonic Bayesian accelerators (cf. Section~\ref{sec:related:photonic_hardware}) within a common framework, we model the photonic processor as a stochastic neural operator rather than an ideal sampler.
In software-based SVI, a stochastic component \(z\) (weight or activation) is sampled from a learned variational distribution \(q_\phi(z)\).
In photonic hardware, however, the learned distribution must be mapped to physical controls, such as optical power, bandwidth, modulator settings, or transmission values, and is therefore constrained by the underlying hardware.
The resulting samples are drawn from a physically realizable distribution
\[
\tilde z \sim q^{\mathrm{phys}}_{\psi}(z \mid \phi),
\]
where \(\psi\) summarizes hardware properties such as dynamic range, quantization, saturation, programming error, noise, and sample-independence constraints.
Consequently, photonic hardware implements a constrained physical variational family rather than the ideal variational family assumed during training.

This abstraction separates the design of a photonic BNN into three orthogonal dimensions:
\begin{enumerate}
    \item \emph{Location of stochasticity}: whether stochasticity is applied to weights or activations.
    \item \emph{Modality of stochasticity}: whether stochasticity is applied additively or multiplicatively.
    \item \emph{Hardware constraints}: which means, variances, and samples can be represented physically.
\end{enumerate}
The considered constraints include finite interface precision, limited representable mean and variance ranges, programming error, non-ideal distribution shapes, and possible correlations between samples.

\subsection{Stochastic Operator Formulations}
\label{sec:operator_formulations}

For a generic linear or convolutional layer, two stochastic operator formulations arise depending on whether stochasticity is applied to the weights or the activations.

For weight-space stochasticity, the layer evaluates
\[
    a = f(x,\tilde w),
    \qquad
    \tilde w = h_\psi(\mu_w,\sigma_w,\epsilon).
\]
Here, \(\mu_w\) and \(\sigma_w\) are the learned variational parameters and have the same shape as the corresponding weight tensor.
The stochastic sample \(\epsilon\) is provided either by the software surrogate or by the physical entropy source.
In the ideal Gaussian surrogate, \(\epsilon \sim \mathcal{N}(0,1)\), whereas in hardware it follows the standardized physical noise distribution.
\todo[inline, color=yellow]{WX: should we move the explaination of $\epsilon$ to the background section? -> HB: Yeah, I think that would be good, since the related work already uses the location definition.}

For activation-space stochasticity, the deterministic layer output is computed first and stochasticity is then applied to the activations,
\[
    a = f(x,w),
    \qquad
    \tilde{a} = h_\psi(a,\sigma_a,\epsilon).
\]
Here, \(a\) serves as the mean of the stochastic activation distribution, while \(\sigma_a\) denotes the corresponding standard deviation.
Depending on the operator design, \(\sigma_a\) may be learned globally, per layer, per channel, per activation, or fixed by the hardware model.
Unlike weight-space stochasticity, the mean is therefore determined by the input and deterministic layer computation rather than learned directly.
This operator-level abstraction is independent of the physical injection point; photonic hardware may introduce stochasticity before, during, or after the optical linear operation.

The function \(h_\psi\) denotes the hardware-constrained sampling process. 
Its simplest realizations are additive,
\[
h_\psi(\mu,\sigma,\epsilon)=
C_\psi(\mu+\sigma\epsilon),
\]
and multiplicative,
\[
h_\psi(\mu,\sigma,\epsilon)=
C_\psi\!\left(\mu(1+\sigma\epsilon)\right).
\]
The additive formulation uses an absolute noise scale, whereas the multiplicative formulation specifies a relative noise scale, for which the absolute standard deviation before hardware constraints is given by \(|\mu|\sigma\).

\begin{figure}[t]
	\centering
	\begin{subfigure}[b]{0.48\columnwidth}
		\centering
		\includegraphics[width=\linewidth]{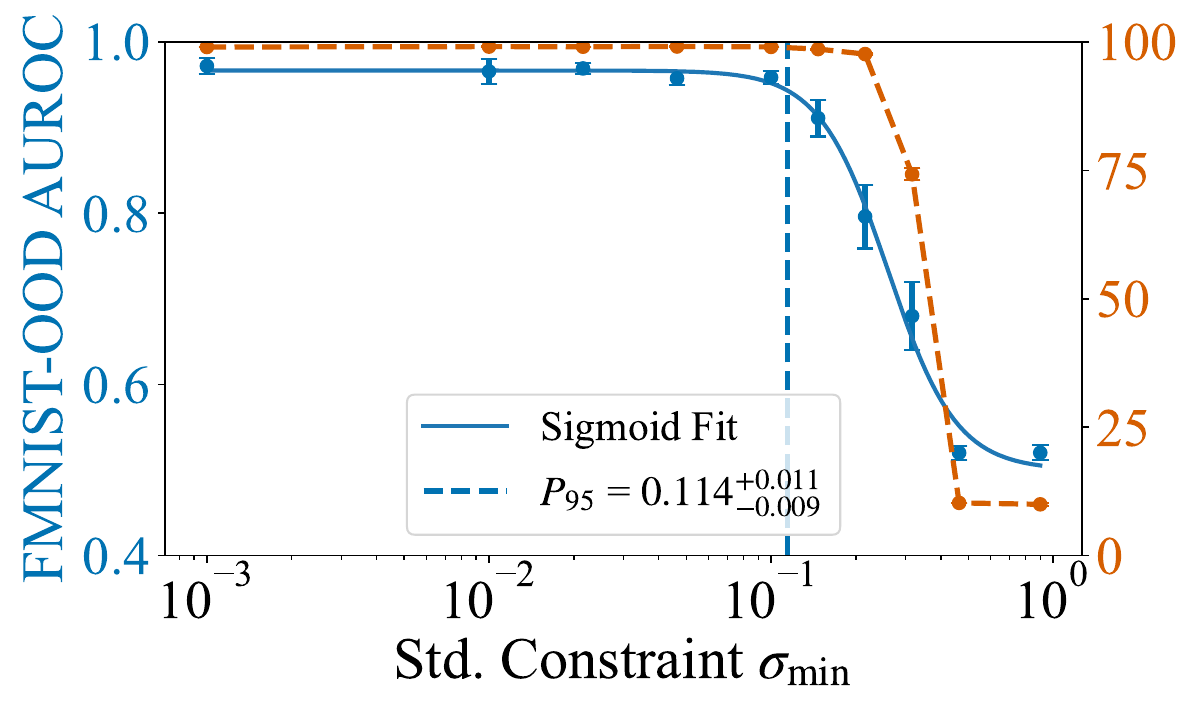}
		\caption{W-Add}
		\label{fig:a}
	\end{subfigure}
	\hfill
	\begin{subfigure}[b]{0.48\columnwidth}
		\centering
		\includegraphics[width=\linewidth]{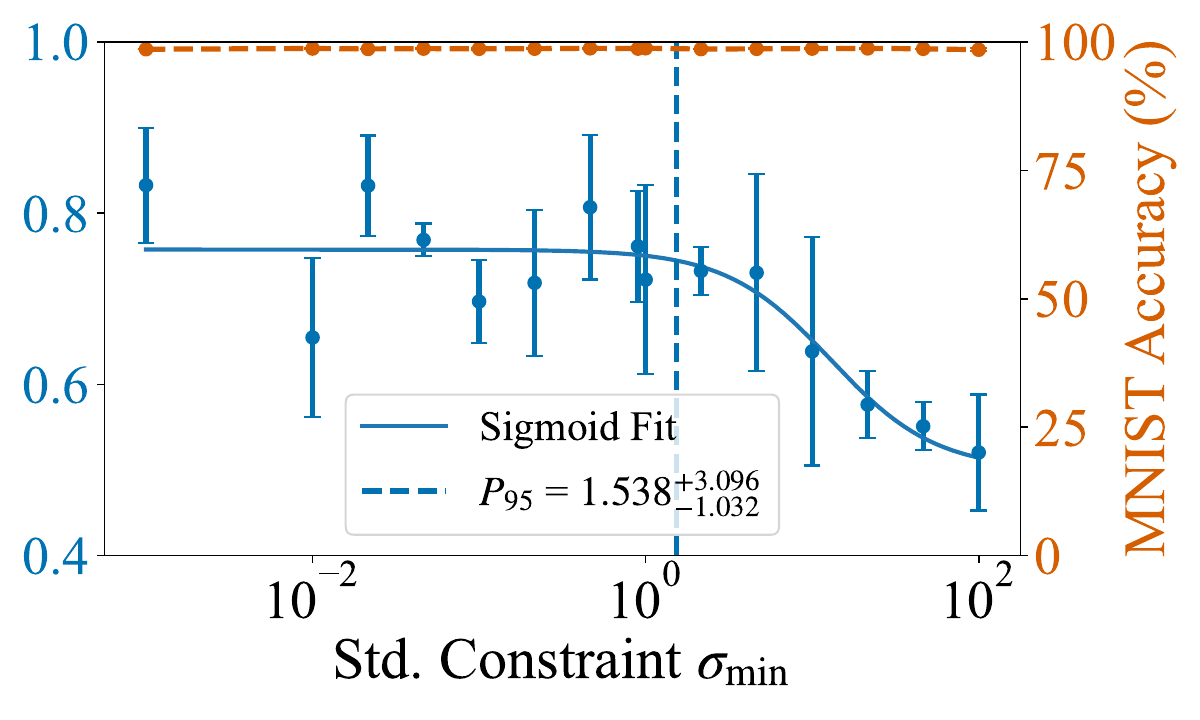}
		\caption{A-Add}
		\label{fig:b}
	\end{subfigure}
	
	\caption{Effects of lower bound of constrained Std. Results are reported for LeNet/Dirty-MNIST.
	}
	\label{fig:scale_lower}
\end{figure}

\subsection{Hardware Constraints and Scope}

The operator \(C_\psi\) collects the hardware constraints that act before,
during, or after sampling. 
These constraints are determined by which stochastic operator can be physically realized, and therefore which variational family is available to the BNN. 
Their relation to photonic hardware is summarized in Table~\ref{tab:constraints} and discussed in detail in Appendix~\ref{sec:sources_of_constraints}.

Relevant constraints include finite interface precision, limited representable mean and scale ranges, clipping, saturation, programming error, deviations from the Gaussian surrogate distribution, and sample correlations.
Distribution-shape mismatch and sample correlations are important potential hardware limitations, but previous photonic BNN implementations either found Gaussian surrogate models sufficient in practice (\WPRNG , \AXBar , \ATWI ) or explicitly avoided measurable sample correlations (\AVac , \AXBar , \ATWI ).
We therefore include these effects in the operator model, but do not investigate them as co-design parameters.

Instead, this work focuses on constraints that directly affect the operator parameterization and can be systematically ablated: the location and modality of stochasticity, input quantization, programming error on the representable mean and scale, and lower and upper bounds on the representable mean and scale.
The binary location and modality dimensions give rise to four stochastic operator variants: weight-additive (\textbf{W-Add}), weight-multiplicative (\textbf{W-Mul}), activation-additive (\textbf{A-Add}), and activation-multiplicative (\textbf{A-Mul}).
The remaining continuous constraints are evaluated independently for each variant.
\todo[inline, color=yellow]{Table 1 also shows the constraints from TWI paper.}
\todo[inline, color=yellow]{Should we rephrase symmetric/asymmetric as signed and unsigned or something else? Because what we want to stress here is whether the range contains negative-part. -> HB: Yes }

\begin{figure}[t!]
	\centering
	
	\begin{subfigure}[b]{0.48\linewidth}
		\centering
		\includegraphics[width=\linewidth]{
			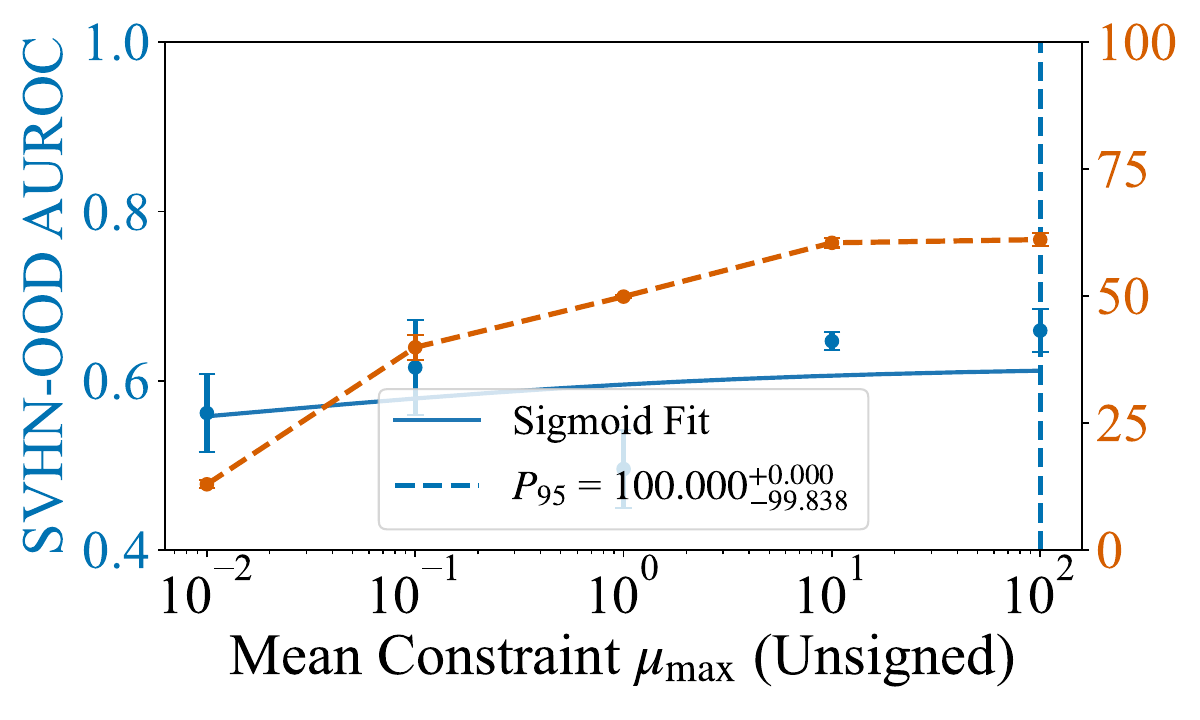
		}
		
		\vspace{0.5em}
		
		\includegraphics[width=\linewidth]{
			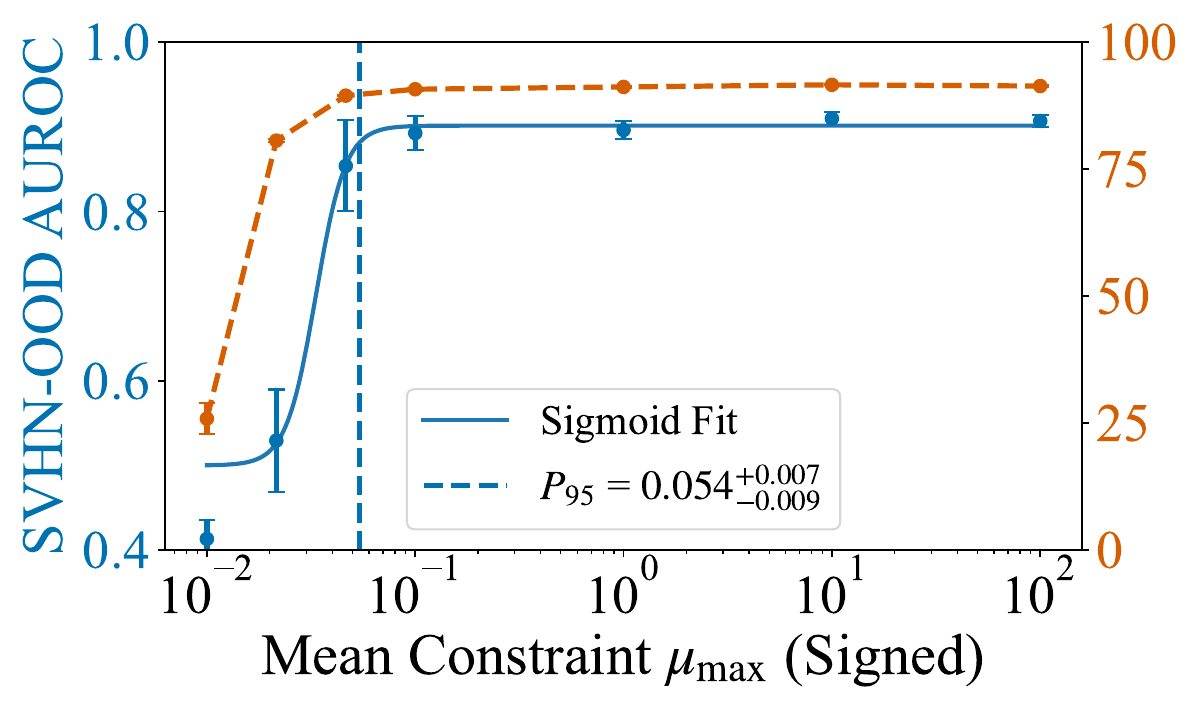
		}
		\caption{W-Add}
		\label{fig:mean-constraint-w-add}
	\end{subfigure}
	\hfill
	\begin{subfigure}[b]{0.48\linewidth}
		\centering
		\includegraphics[width=\linewidth]{
			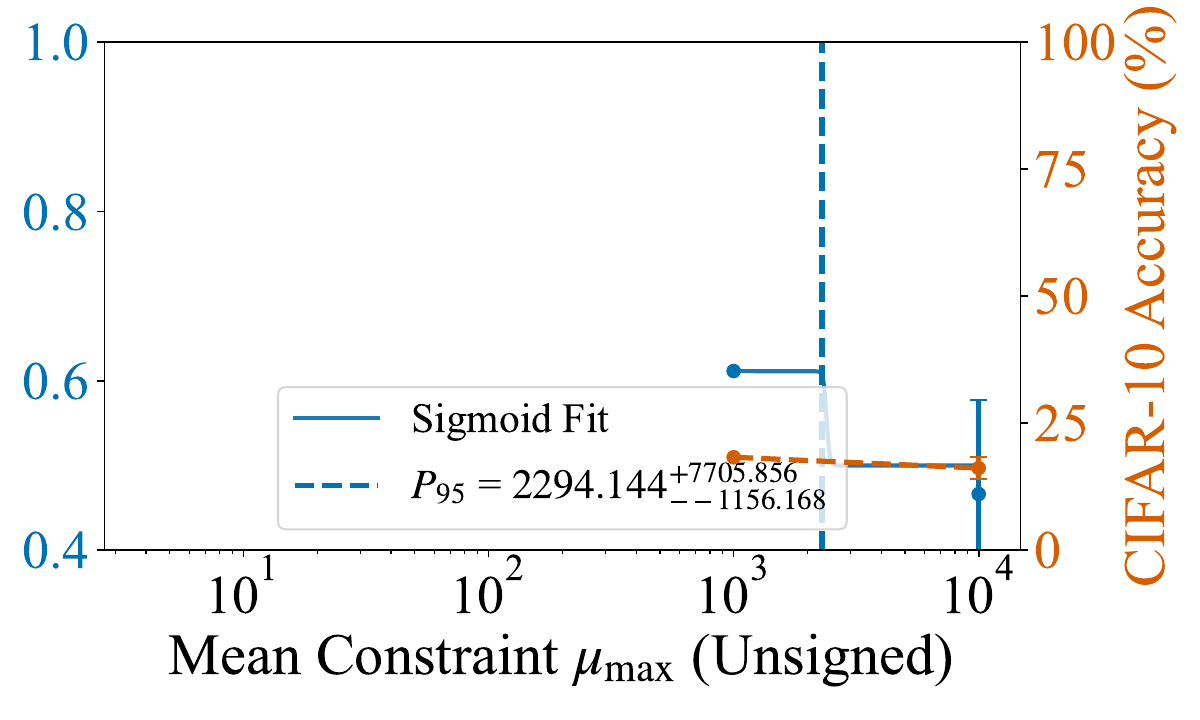
		}
		
		\vspace{0.5em}
		
		\includegraphics[width=\linewidth]{
			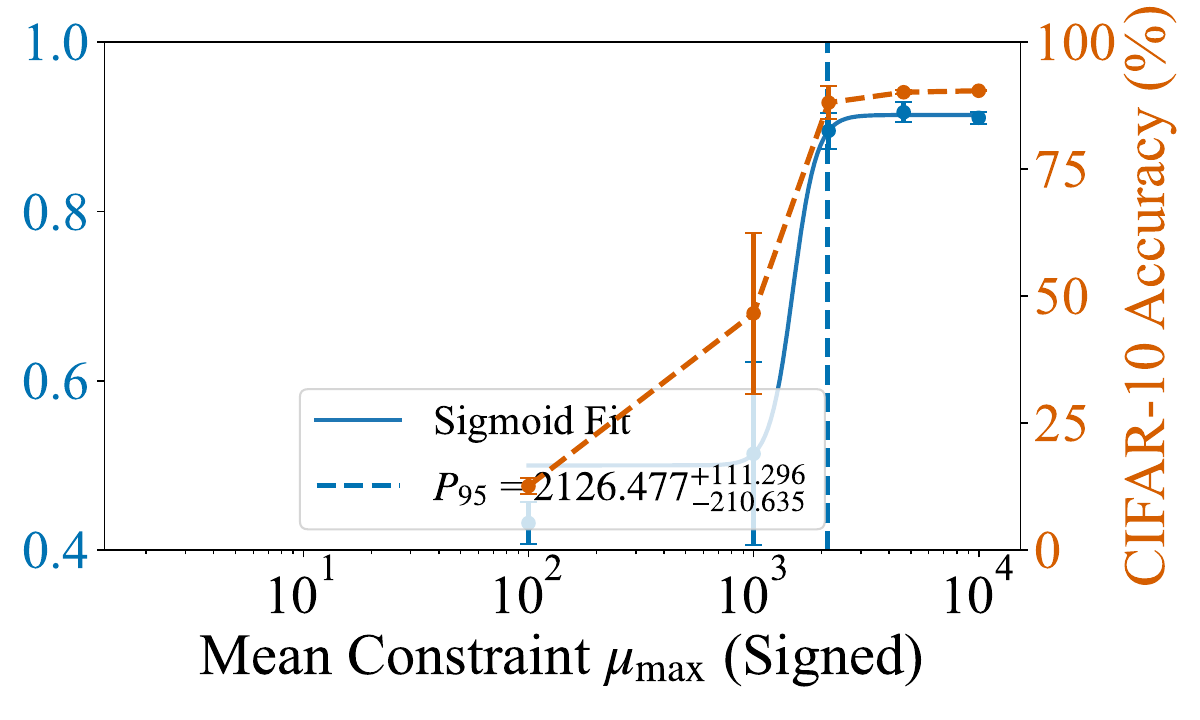
		}
		\caption{A-Add}
		\label{fig:mean-constraint-a-add}
	\end{subfigure}
	
	\caption{
		Effects of unsigned (top) and signed (bottom) mean constraints for
		weight-space and activation-space additive stochasticity.
		Results are reported for ResNet-18/CIFAR-10.
	}
	\label{fig:mean-constraint-comparison}
\end{figure}

The stochastic operator formulations in Section~\ref{sec:operator_formulations} are intentionally architecture-agnostic.
Whether stochasticity is realized in a photonic crossbar, a TWI architecture, or another photonic accelerator, the hardware is described by its location and modality of stochasticity together with its representable parameter ranges, precision, and sampling properties.
The central co-design question is therefore which constrained variational family a given hardware platform realizes and whether training can compensate for the resulting limitations.
We investigate this question by implementing the stochastic operators in the \texttt{Pyro} probabilistic programming library, explicitly incorporating the hardware constraints described above.
Figure~\ref{fig:method:operator_overview} illustrates the implementation for a convolutional layer.
For weight-space stochasticity, all constraints are applied to the weights before the convolution, whereas for activation-space stochasticity they are applied to the layer outputs after the deterministic computation.
Constraints on the standard deviation are enforced through the variational distributions defined by the \texttt{Pyro} guides rather than within the operator itself, requiring separate guides for the weight-space and activation-space formulations.

%% file: sections/05_experiments.tex
\begin{table*}[t]
	\centering
	\scriptsize
	
	\resizebox{\textwidth}{!}{%
		\begin{threeparttable}
			\caption{
				Summary of empirical hardware constraints and corresponding model- and hardware-side responses.
			}
			\label{tab:design_guidelines}
			\begin{tabularx}{\linewidth}{
				>{\raggedright\arraybackslash}p{0.11\linewidth}
				>{\raggedright\arraybackslash}p{0.35\linewidth}
				>{\raggedright\arraybackslash}p{0.22\linewidth}
				X
			}
			\toprule
			\textbf{Aspect}
			&
			\textbf{Empirical finding}
			&
			\textbf{Model-side response}
			&
			\textbf{Hardware-side response}
			\\
			\midrule

			\multicolumn{4}{@{}l}{\textbf{Hardware capabilities}}
			\\
			\addlinespace[1pt]

			Signedness
			&
			Unsigned representations severely degrade weight-space performance and collapse activation-space inference.
			&
			Use positive--negative decomposition or signed variational parameterization.
			&
			Support signed representations, e.g., differential or dual-rail encoding.
			\\
			\addlinespace[1pt]

			Stochasticity location
			&
			Weight space provides stronger UQ, whereas activation space tolerates substantially more mean-programming error.
			&
			Choose the stochasticity location according to the primary design objective.
			&
			Support the stochasticity location matching the targeted uncertainty-quality--noise-robustness trade-off.
			\\
			\addlinespace[1pt]

			Stochasticity modality
			&
			Additive configurations tolerate larger minimum std; multiplicative configurations tolerate smaller maximum std.
			&
			Match variational parameterization to the hardware's native stochastic process.
			&
			Avoid costly transformations of the native stochastic process.
			\\

			\midrule

			\multicolumn{4}{@{}l}{\textbf{Representational constraints}}
			\\
			\addlinespace[1pt]

			Input quantization
			&
			Stable down to approximately 4 effective input bits.
			&
			Apply quantization-aware training.
			&
			Provide at least four effective input bits.
			\\
			\addlinespace[1pt]

			Mean bounds
			&
			Activation space requires a substantially larger mean range.
			&
			Constrain or clip represented means during training.
			&
			Increase dynamic range or provide programmable gain.
			\\
			\addlinespace[1pt]

			Std bounds
			&
			Additive models better tolerate a nonzero minimum std, whereas multiplicative models tolerate a smaller maximum std.
			&
			Use a constrained variational parameterization.
			&
			Extend the available std range or select a compatible stochasticity modality.
			\\

			\midrule

			\multicolumn{4}{@{}l}{\textbf{Fidelity constraints}}
			\\
			\addlinespace[1pt]

			Mean programming error
			&
			Weight-space implementations degrade earlier.
			&
			Train with the expected programming-error model.
			&
			Improve mean-programming SNR or use activation-space stochasticity.
			\\
			\addlinespace[1pt]

			Std programming error
			&
			A-Add is very tolerant, A-Mul is not tolerant, W-Add is more tolerant than W-Mul
			&
			Train with the expected programming-error model.
			&
			Improve std-programming SNR or favour additive implementations.
			\\

			\bottomrule
			\end{tabularx}

		\end{threeparttable}
	}
\end{table*}

\section{Sensitivity Analysis}
\label{sec:ablation}

We aim to isolate the impact of individual hardware constraints on BNN performance. We  adopt a two-stage methodology. 
First, we identify a well-tuned baseline for each group through hyperparameter optimization. 
We then perform ablation studies, varying only a single hardware constraint while keeping other settings fixed.

Since both the location and type of stochasticity influence the optimization dynamics of a BNN, each of the four configurations is tuned independently. 
Hyperparameter optimization is performed with Optuna~\cite{optuna}. For each configuration and dataset, we run between 29 and 37 trials: 100-epoch training runs from random initialization for LeNet on Dirty-MNIST, and 25-epoch training runs initialized from task-specific pretrained weights for ResNet-18 on CIFAR-10/CINIC-10.
We optimize the learning rate, the maximum factor used for KL annealing, the prior standard deviation, and the initial standard deviation of the variational distribution. 
All models are trained using the RAdamScheduleFree optimizer~\cite{schedulefree} and TrivialAugment~\cite{trivialaugment} for image augmentation. 
Since predictive accuracy and OOD AUROC are competing objectives, the final hyperparameter configuration is selected manually from the Pareto-optimal set to achieve a balanced trade-off between both metrics.

Using these optimized baselines, each hardware constraint is varied independently over a broad range. 
Every measurement is repeated three times to estimate variability. 
We fit a logistic curve to the AUROC values and identify the constraint level causing a 5\% drop from its fitted maximum.
We use OOD AUROC rather than accuracy because it is more sensitive to hardware-induced performance degradation. 

Uncertainties in the logistic curve fit were estimated using Monte Carlo uncertainty propagation. 
To this end, bootstrap sampling was used to fit the curve 500 times to varied data, and the resulting distribution of fitted parameters was then used to quantify parameter uncertainty coverage with equivalence to one-sigma standard deviation coverage.

\subsection{Analysis of Individual Constraints}

\begin{figure*}[t]
	\centering
	
	\begin{subfigure}[b]{0.49\textwidth}
		\centering
		\includegraphics[width=\linewidth,trim={0 0 0 1.4em},clip]{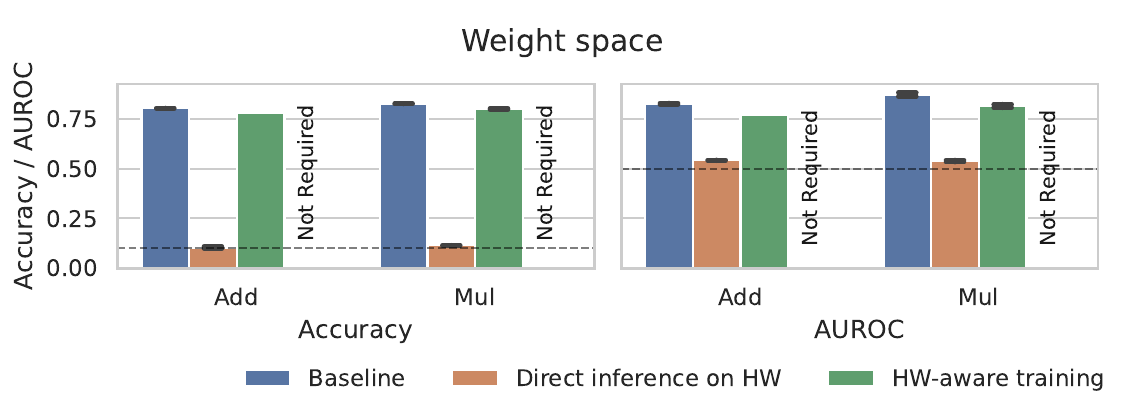}
	\end{subfigure}
	\hfill
	\begin{subfigure}[b]{0.49\textwidth}
		\centering
		\includegraphics[width=\linewidth,trim={0 0 0 1.4em},clip]{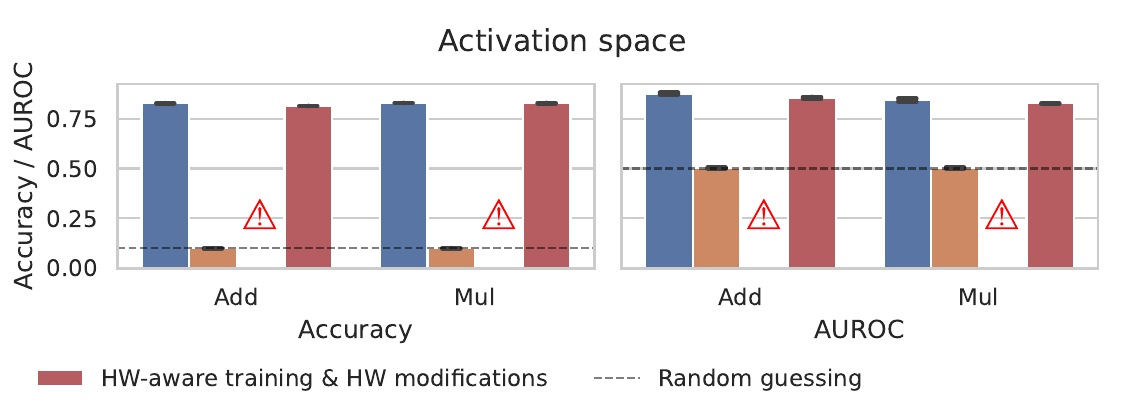}
	\end{subfigure}
	
	\caption{
		Co-design study for an example accelerator based on the \ATWI{} constraints.
		For weight-space stochasticity, the constraints lie within the stable operating range of Table~\ref{tab:ablation}, requiring only hardware-aware training.
		For activation-space stochasticity, hardware-aware training does not converge until the relevant hardware constraints are relaxed according to Table~\ref{tab:ablation}.
	}
	\label{fig:5-full-CINIC}
\end{figure*}

Table~\ref{tab:ablation} summarizes the evaluated ranges and the resulting 5\%-degradation thresholds under HW-aware training.
Results without hardware awareness are provided in the supplementary material and consistently result in worse robustness.
This is also substantiated by the results in the following Section~\ref{sec:fully_integrated}.
The ablation reveals clear differences between weight- and activation-space stochasticity. Overall, weight-space implementations provide more reliable uncertainty quantification, consistently achieving higher AUROC across the evaluated constraints. 
In contrast, activation-space implementations are more robust to strong hardware noise, maintaining classification accuracy when weight-space models begin to degrade. 
We attribute this partly to the number of stochastic components: LeNet on Dirty-MNIST contains ~61.7K stochastic weights versus 6.5K stochastic activations, while ResNet-18 on CIFAR-10 contains 11.7M versus 0.6M, respectively.
Activation-space stochasticity is therefore introduced at substantially fewer points in the network, limiting noise accumulation and improving predictive robustness. However, fewer stochastic variables also yield weaker and less stable uncertainty estimates, even when the scale parameter is only weakly constrained (Figure~\ref{fig:scale_lower}).

A second key observation concerns constraints on the representable mean. Signed constraints on the representable mean consistently outperform unsigned constraints. 
In weight space, restricting the mean to positive values reduces classification accuracy by at least 30\% and AUROC by more than 20\%. In activation space, the effect is even more pronounced, causing complete model collapse (Figure~\ref{fig:mean-constraint-comparison}). 

Finally, activation-space stochasticity requires a substantially larger representable mean range than weight-space stochasticity. 
As shown in Figure~\ref{fig:mean-constraint-w-add} and~\ref{fig:mean-constraint-a-add}, the required signed range differs by approximately five orders of magnitude ($10^{3}$ versus $10^{-2}$). 
This reflects the accumulation of activations across successive layers, while weights remain bounded by the learned parameter scale.

\subsection{Hardware Design Guidelines} %
\label{sec:guidelines}

Table~\ref{tab:design_guidelines} translates the sensitivity analysis into model--hardware co-design guidelines. 
Signed representation is a hard requirement: one-sided means severely degrade weight-space models and cause activation-space models to collapse. 
In contrast, input quantization can be compensated through quantization-aware training down to approximately four bits in the evaluated setup. 
Stochasticity location and modality are largely fixed by the hardware architecture and introduce explicit trade-offs. 
Weight-space stochasticity provides stronger UQ, whereas activation-space stochasticity tolerates substantially more mean-programming error. 
Additive configurations better tolerate low bounds and programming error on the standard deviation, particularly for A-Add, while multiplicative configurations tolerate tighter upper scale bounds.

Mean and standard deviation bounds can be compensated through training as long as they fall within the tolerated ranges (see Table~\ref{tab:ablation} for the example of ResNet-18/CIFAR-10); beyond these ranges, the hardware can no longer represent the required variational distributions. 
This is particularly restrictive for activation-space stochasticity, which requires a substantially larger mean range. 
Programming error similarly acts as a fidelity constraint: training under the expected noise model can maintain performance within the tolerated regime, whereas larger errors require improved programming SNR. 
Consequently, the variational parameterization should match the hardware's native stochastic process, hardware constraints should be included during training, and hardware intervention is required only when signedness, range, or programming fidelity falls outside the identified feasible region.

%% file: sections/05a_coupled.tex
\section{Co-Design Validation under Coupled Hardware Constraints}
\label{sec:fully_integrated}

The preceding analysis considered each hardware constraint independently.
We now evaluate whether the resulting design guidelines remain predictive when all measured \ATWI{} constraints (Table~\ref{tab:constraints}) are applied simultaneously.
We use ResNet-18 on CINIC-10 with SVHN for OOD detection, providing a more challenging validation setting than the CIFAR-10 experiments from which the thresholds in Table~\ref{tab:ablation} were derived.
We consider the native activation-space A-TWI mapping and a hypothetical weight-space mapping subject to the same measured A-TWI constraint values.

For the weight-space mapping, Table~\ref{tab:ablation} predicts that the \ATWI~constraint values remain within the trainable regime. 
Directly transferring an unconstrained BNN nevertheless causes substantial degradation, because its learned variational parameters are not adapted to the restricted mean and scale ranges, quantization, and programming error. 
When the same constraints are included during training, both predictive accuracy and OOD AUROC recover substantially, as shown in Figure~\ref{fig:5-full-CINIC}. 
This result demonstrates that individually tolerable constraints can also be compensated when applied jointly through retraining. 
Accordingly, the weight-space configuration does not require improvements of underlying hardware constraints, but does require hardware-aware training.

The activation-space mapping leads to a different outcome.
As for the weight-space mapping, direct transfer of an unconstrained BNN to the \ATWI{} constraint model causes severe performance degradation.
Unlike the weight-space case, however, the sensitivity analysis in Table~\ref{tab:ablation} predicts that the restricted representable mean range lies outside the trainable regime.
Accordingly, hardware-aware training remains unstable and fails to recover performance, confirming that the representable mean range acts as a hard representational constraint for this mapping.

We consequently modify only the mean representation, satisfying the minimum requirement identified in Table~\ref{tab:ablation}. 
With this guideline-derived hardware adjustment, training becomes stable and recovers substantial predictive and uncertainty-estimation performance, as shown in Figure~\ref{fig:5-full-CINIC}. 
The remaining \ATWI~constraints are left unchanged. 
This isolates the mean representation as the limiting hardware property and shows that a targeted modification is sufficient; redesigning all numerical and stochastic properties of the accelerator is unnecessary.

These experiments establish two complementary co-design regimes. 
When the required variational distributions remain representable, hardware-aware training compensates for coupled hardware constraints; otherwise, the individual-constraint analysis identifies the required hardware modification.
Our guidelines therefore provide an \emph{a priori} procedure for determining whether a photonic BNN mapping requires software adaptation, hardware intervention, or both.

%% file: sections/06_discussion.tex
\section{Discussion and Limitations}

Our results show that the scalability of photonic BNNs is determined not by any single hardware constraint, but by the variational distributions that remains representable under their combined effect. 
Weight-space stochasticity generally provides stronger UQ, whereas activation-space stochasticity offers greater tolerance to mean-programming noise. 
Likewise, additive stochasticity is comparatively robust to lower scale bounds and scale-programming noise, while multiplicative stochasticity better tolerates restrictive upper scale bounds. 
Signed representation remains a particularly important requirement, especially in activation space.

The integrated \ATWI~study further distinguishes constraints that can be compensated through hardware-aware training from those that require hardware intervention. 
The stable ranges in Table~\ref{tab:ablation} should, however, not be interpreted as universal device specifications. 
Their absolute values depend on the network parameterization, architecture, and workload, although the observed qualitative trade-offs transfer from CIFAR-10 to the more challenging CINIC-10 setting.

Several limitations remain. 
The individual ablations treat hardware constraints independently, whereas physical devices may couple the attainable mean and scale or exhibit simultaneous additive and multiplicative effects. 
We also assume sufficiently independent samples and a suitable Gaussian surrogate, leaving distribution mismatch and temporal or inter-channel correlations outside the present study. 
Finally, this work evaluates inference quality rather than measured hardware latency or energy efficiency. 
Nevertheless, the proposed framework provides a systematic basis for identifying which constraints can be trained through and which must be engineered around when co-designing future photonic Bayesian accelerators.

%% file: sections/07_appendix.tex
\section{Appendix}

\subsection{Additional information on physical sources of individual constraints}
\label{sec:sources_of_constraints}
\todo[inline]{BK: Add that there might be controllable noise + non-controllable base noise (base noise).}
\todo[inline]{To appendix, plus move citations into the constraint table.}

The following section adds more detailed information concerning the related work on photonic BNN accelerators:

\subsubsection{Location and modality of stochasticity}
Unlike standard, digital SVI based BNNs where stochasticity is typically injected at the weights, photonic devices present constraints where the point of stochasticity may not be freely chosen; in these systems, noise can also manifest on the activations, affecting the results of computations rather than just their parameters.
In particular \citet{PBM_vacuum,9931937,brckerhoffplckelmann2024probabilistic,brückerhoffplückelmann2025uncertaintyreasoningphotonicbayesian} utilize stochasticity on the activations instead of weights.
While, \citet{PBNNS} present an accelerator based on weight stochasticity.

Furthermore, the modality of this stochasticity---whether it acts additively or multiplicatively to a given mean value---is generally pre-determined by the hardware architecture.
Here \citet{PBM_vacuum,brckerhoffplckelmann2024probabilistic,brückerhoffplückelmann2025uncertaintyreasoningphotonicbayesian} implement multiplicative stochasticity, while \citet{PBNNS} implement it additively and \citet{9931937} observe combined additive and multiplicative effects.
While such constraints are less restrictive for weight stochasticity (where weights are determined during training), they are significant for activation stochasticity, as the mean is directly dependent on input data and only secondarily on network parameters.

\subsubsection{Representational constraints}
Beyond these conceptual constraints, physical hardware constraints also dictate co-design parameters.
The representable parameter range is limited in physical devices by non-linearities and saturation, as observed in the MZI and MRR devices of \citet{PBNNS,9931937}, the electro-optic modulators of \citet{brckerhoffplckelmann2024probabilistic,brückerhoffplückelmann2025uncertaintyreasoningphotonicbayesian}, or even the photonic process itself~\cite{PBM_vacuum}.
And although analog devices can theoretically represent values of infinite precision, this is limited in practice through limited programming precision \cite{PBNNS,brckerhoffplckelmann2024probabilistic,brückerhoffplückelmann2025uncertaintyreasoningphotonicbayesian}, 
as well as external noncontrollable factors such as thermal effects or manufacturing variances \cite{brückerhoffplückelmann2025uncertaintyreasoningphotonicbayesian}\todo{This last one is quite vague in our paper, drop?}.

\subsubsection{Distribution-shape constraint}
Additionally, the resulting stochastic distribution in photonic systems may deviate from the Gaussian distribution typically assumed in BNNs, e.g. exhibiting Bose-Einstein or folded distributions~\cite{brckerhoffplckelmann2024probabilistic}. 
Hardware limitations can also lead to unwanted cross-correlations between samples and need to be specifically addressed \cite{brckerhoffplckelmann2024probabilistic,brückerhoffplückelmann2025uncertaintyreasoningphotonicbayesian,9931937}. %

\subsubsection{System constraints}
Finally, the hardware is further constrained by the finite precision of the digital components surrounding the photonic core, including, but not limited to, digital to analog conversation and vice versa. \cite{PBNNS,brckerhoffplckelmann2024probabilistic,brückerhoffplückelmann2025uncertaintyreasoningphotonicbayesian}.

\subsection{Ablation studies on individual constraints for more models, datasets and settings}

To isolate the effects of individual hardware constraints on BNN performance, we conduct three ablation studies across two tasks,  with and without hardware-aware training. As described in Section~\ref{sec:ablation}, we first identify a well-tuned baseline for each experimental setting through hyperparameter optimization.

\begin{figure*}[p]
    \centering

    \begin{subfigure}[t]{0.235\textwidth}
        \centering
        \caption{W-Add}
        \includegraphics[width=\linewidth]
            {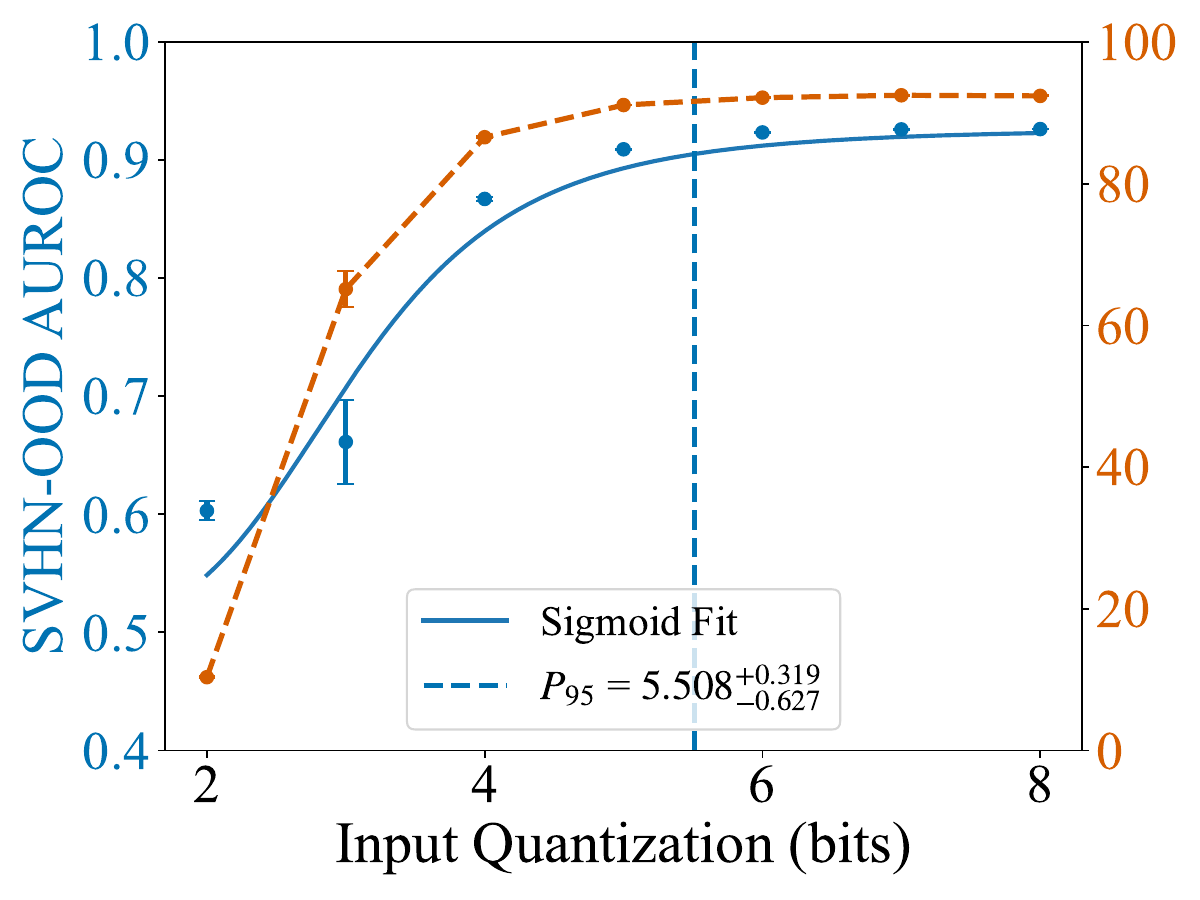}
        \includegraphics[width=\linewidth]
            {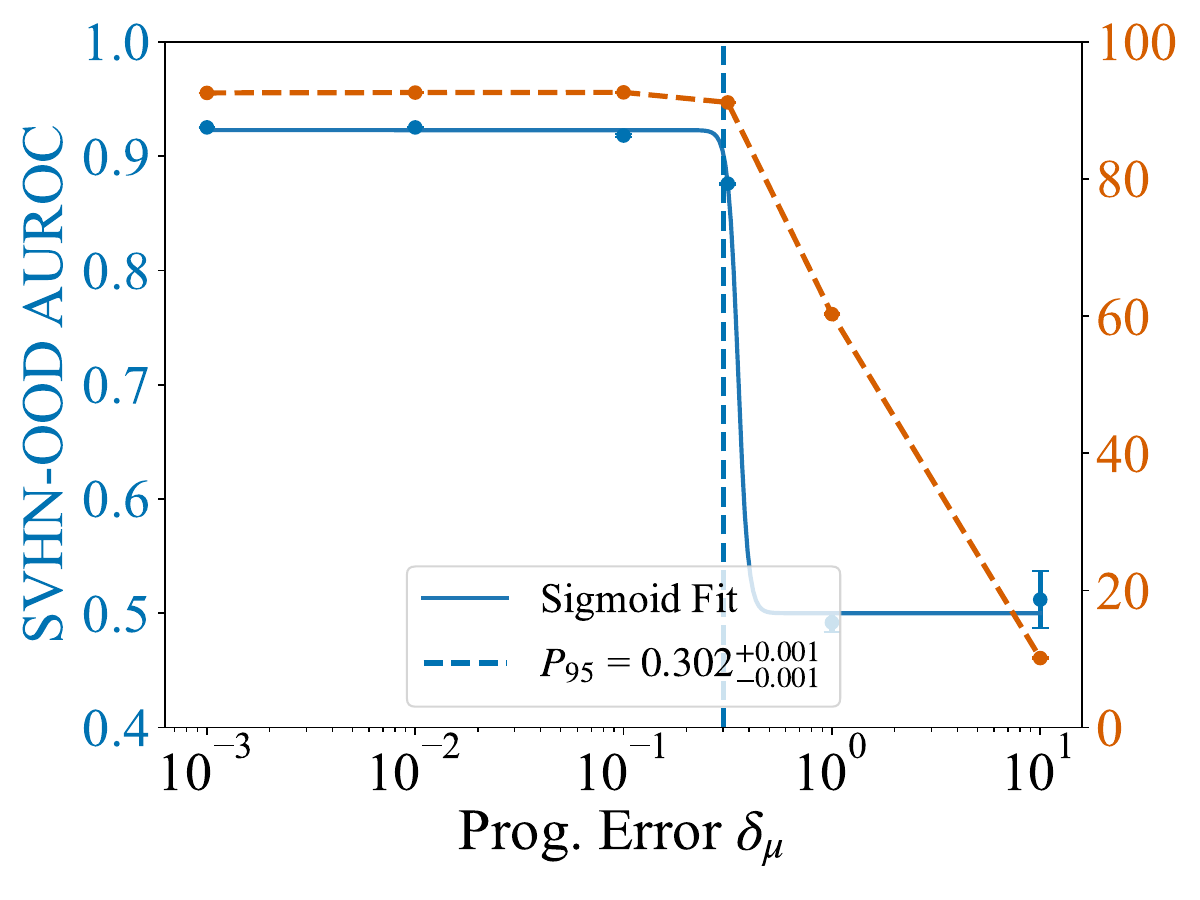}
        \includegraphics[width=\linewidth]
            {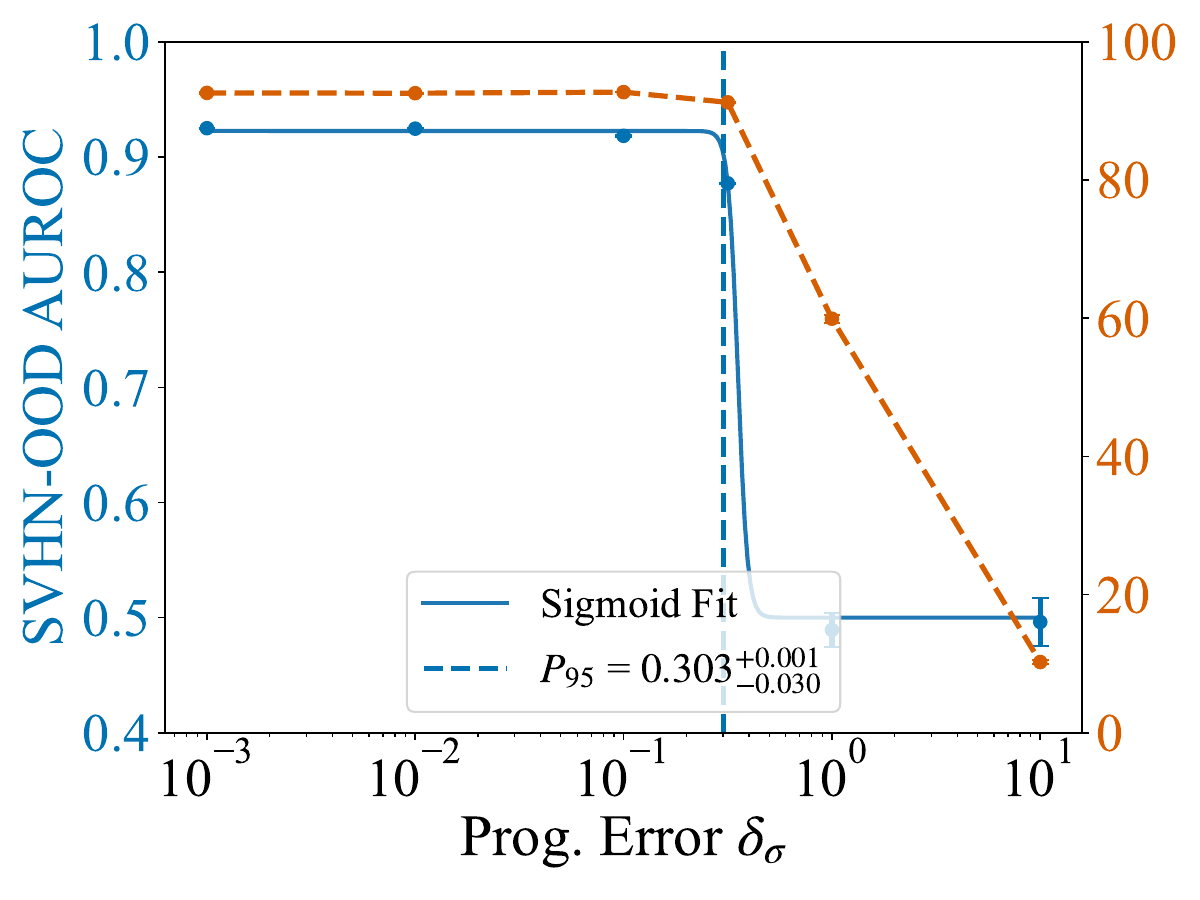}
        \includegraphics[width=\linewidth]
            {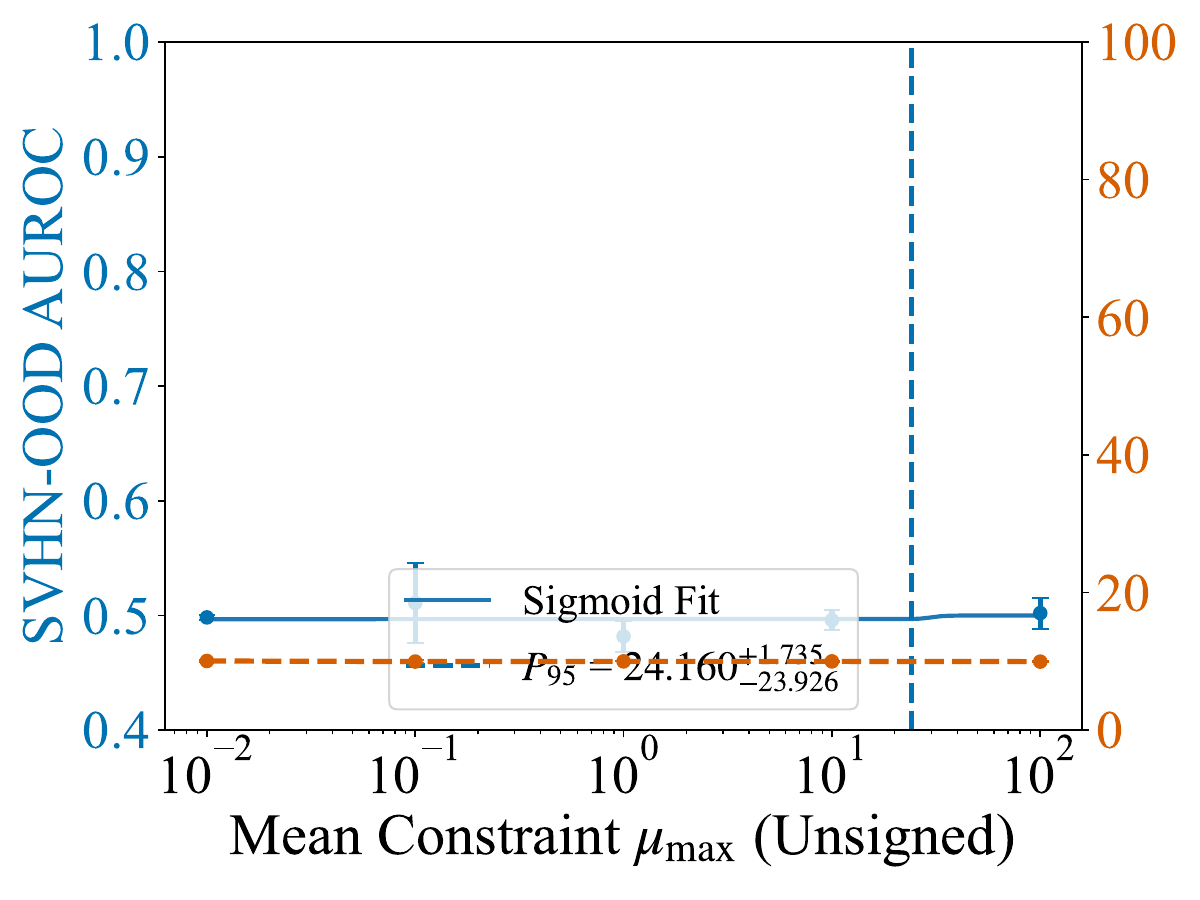}
        \includegraphics[width=\linewidth]
            {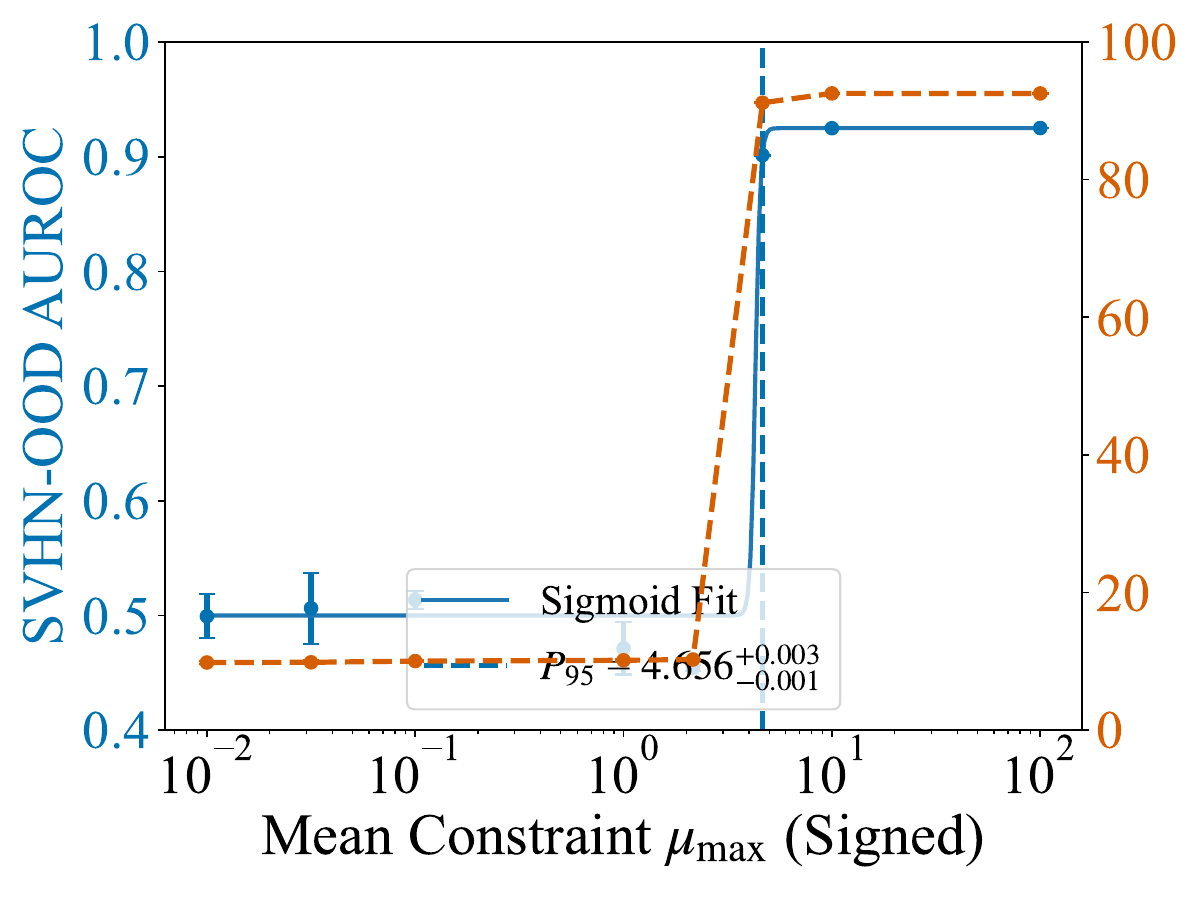}
        \includegraphics[width=\linewidth]
            {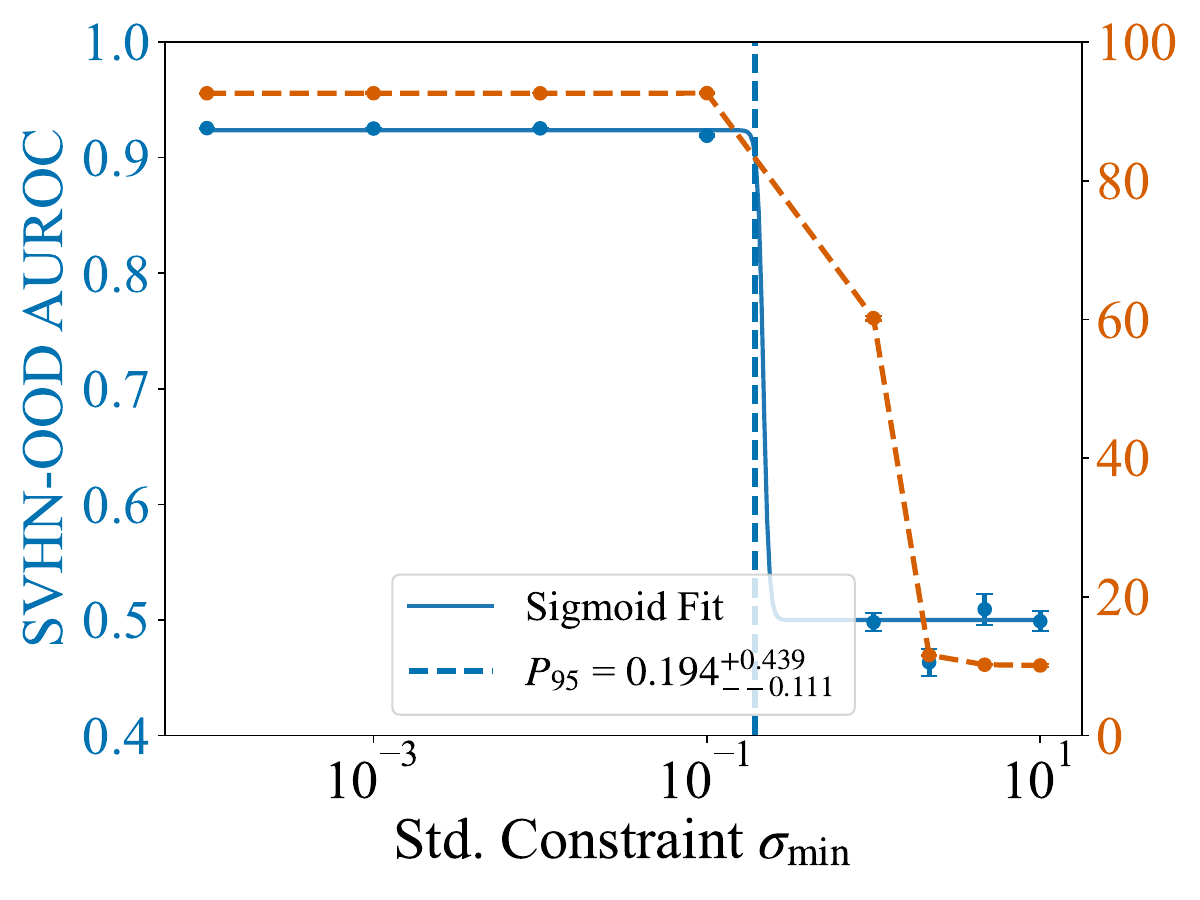}
        \includegraphics[width=\linewidth]
            {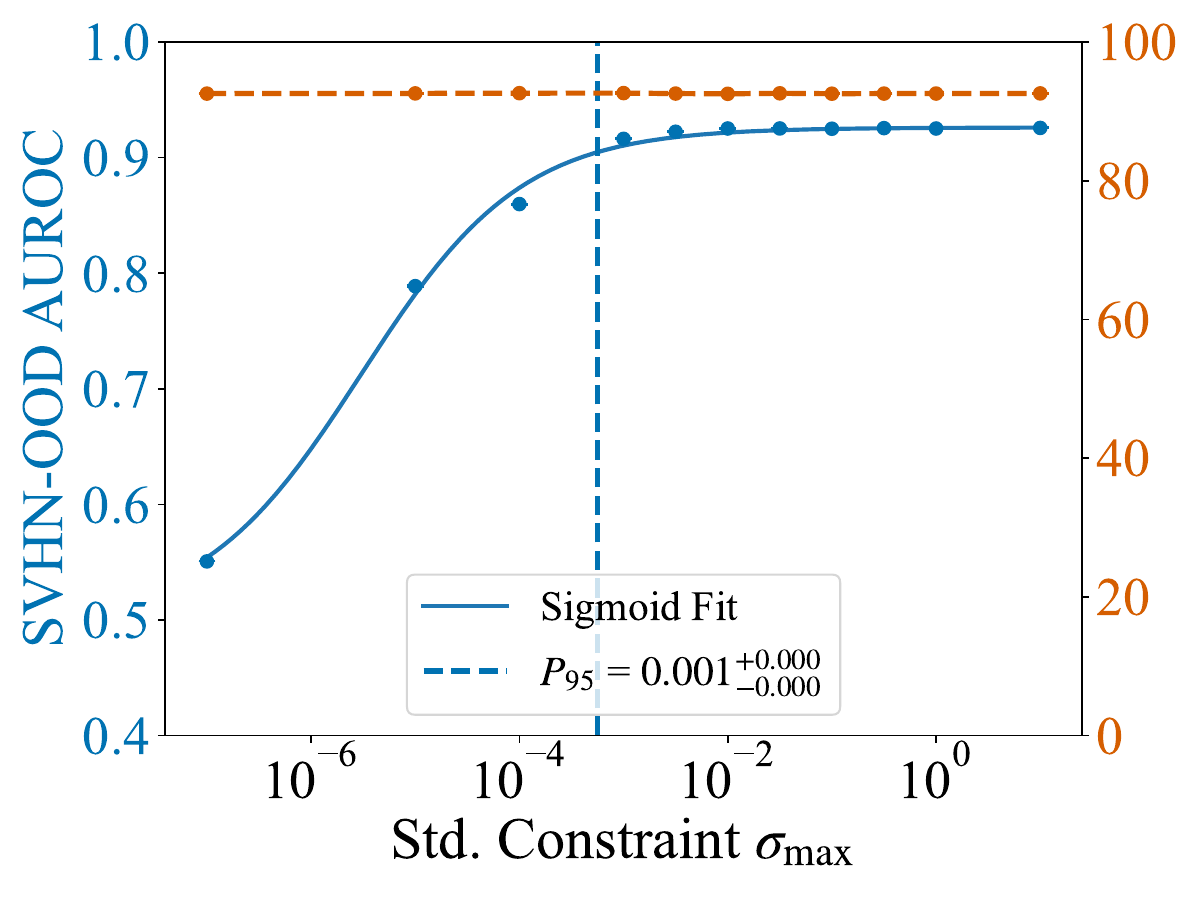}
    \end{subfigure}
    \hfill
    \begin{subfigure}[t]{0.235\textwidth}
        \centering
        \caption{W-Mul}
        \includegraphics[width=\linewidth]
            {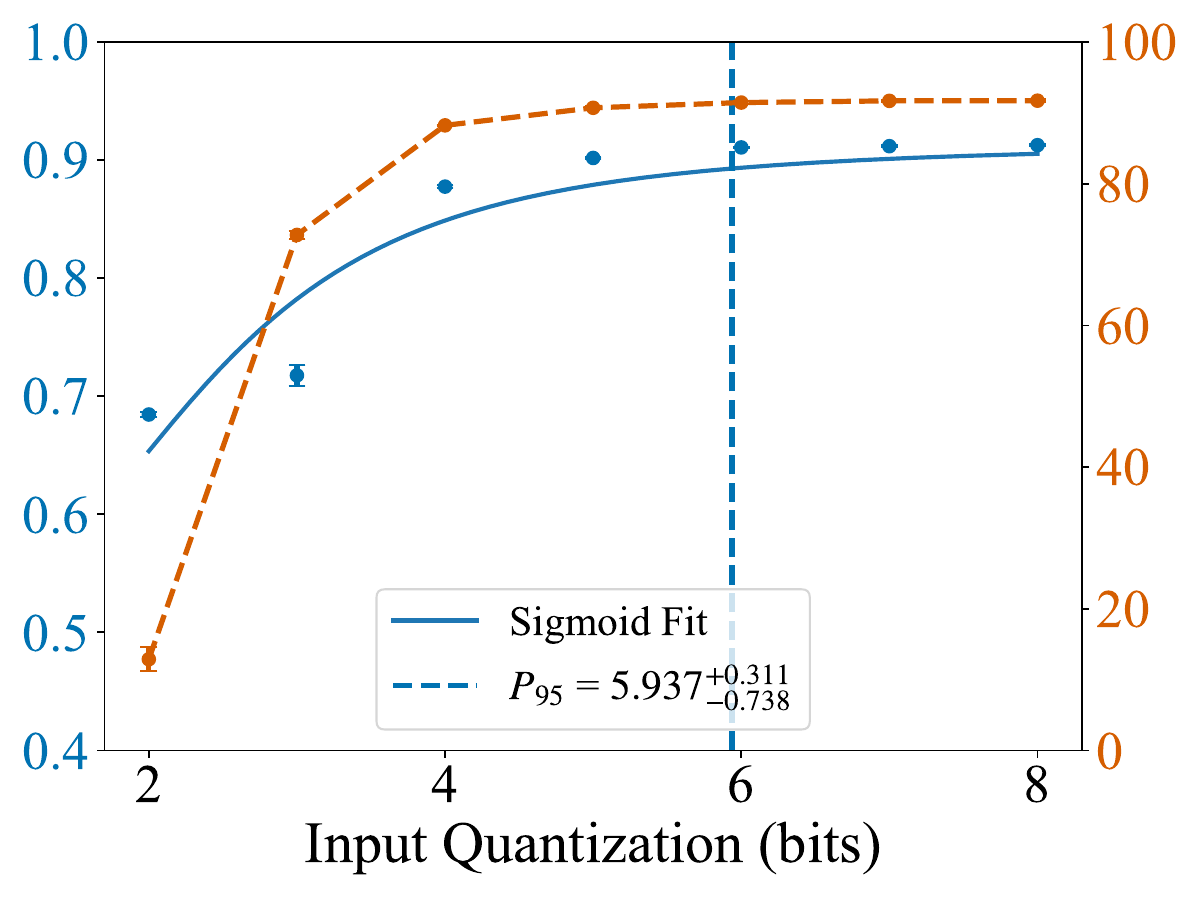}
        \includegraphics[width=\linewidth]
            {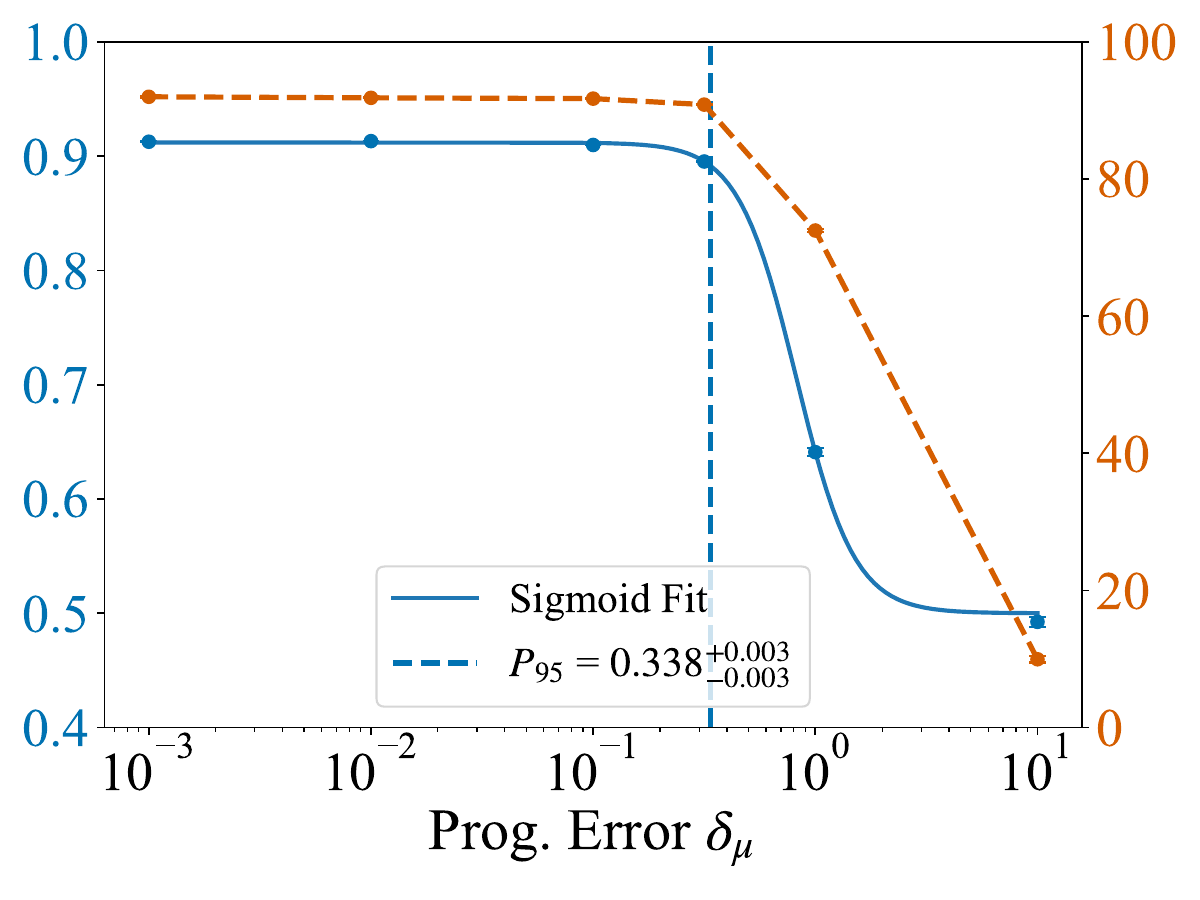}
        \includegraphics[width=\linewidth]
            {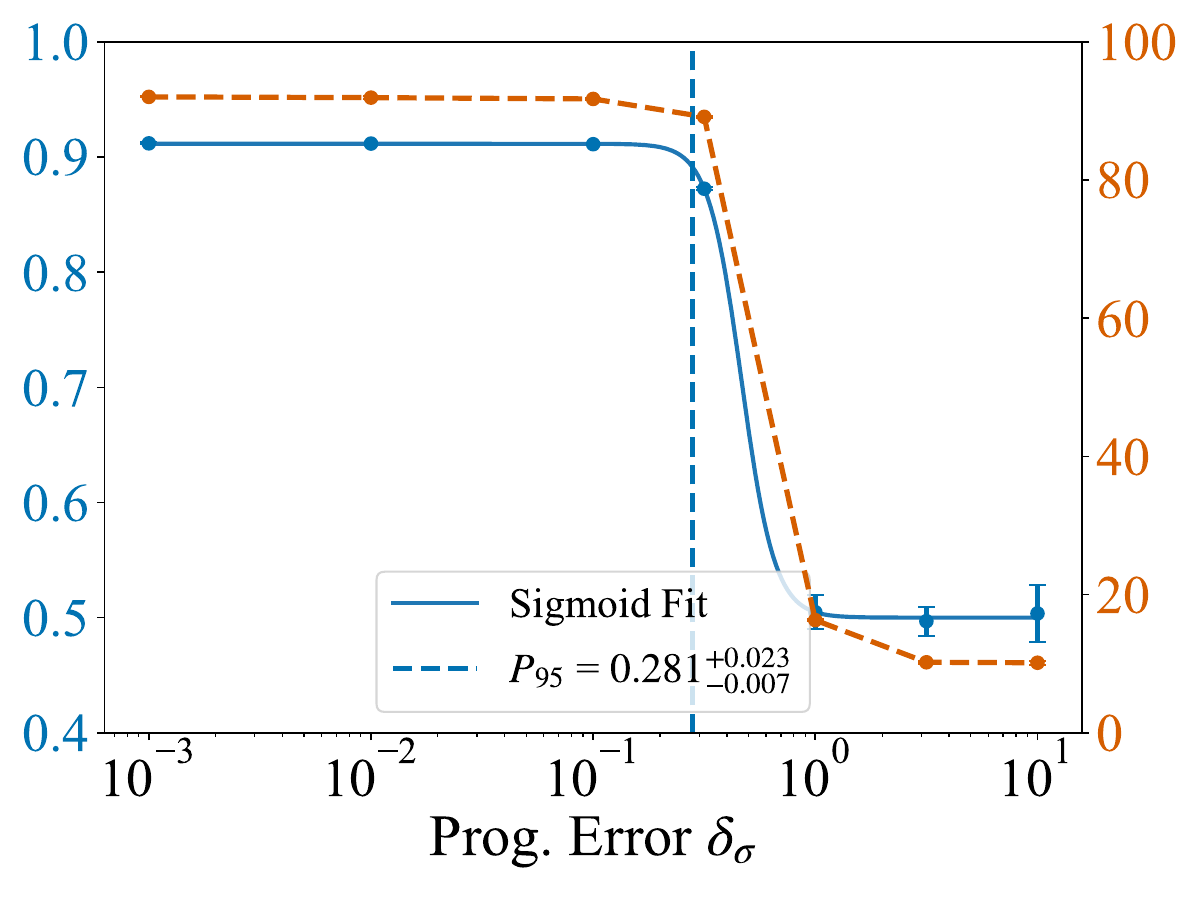}
        \includegraphics[width=\linewidth]
            {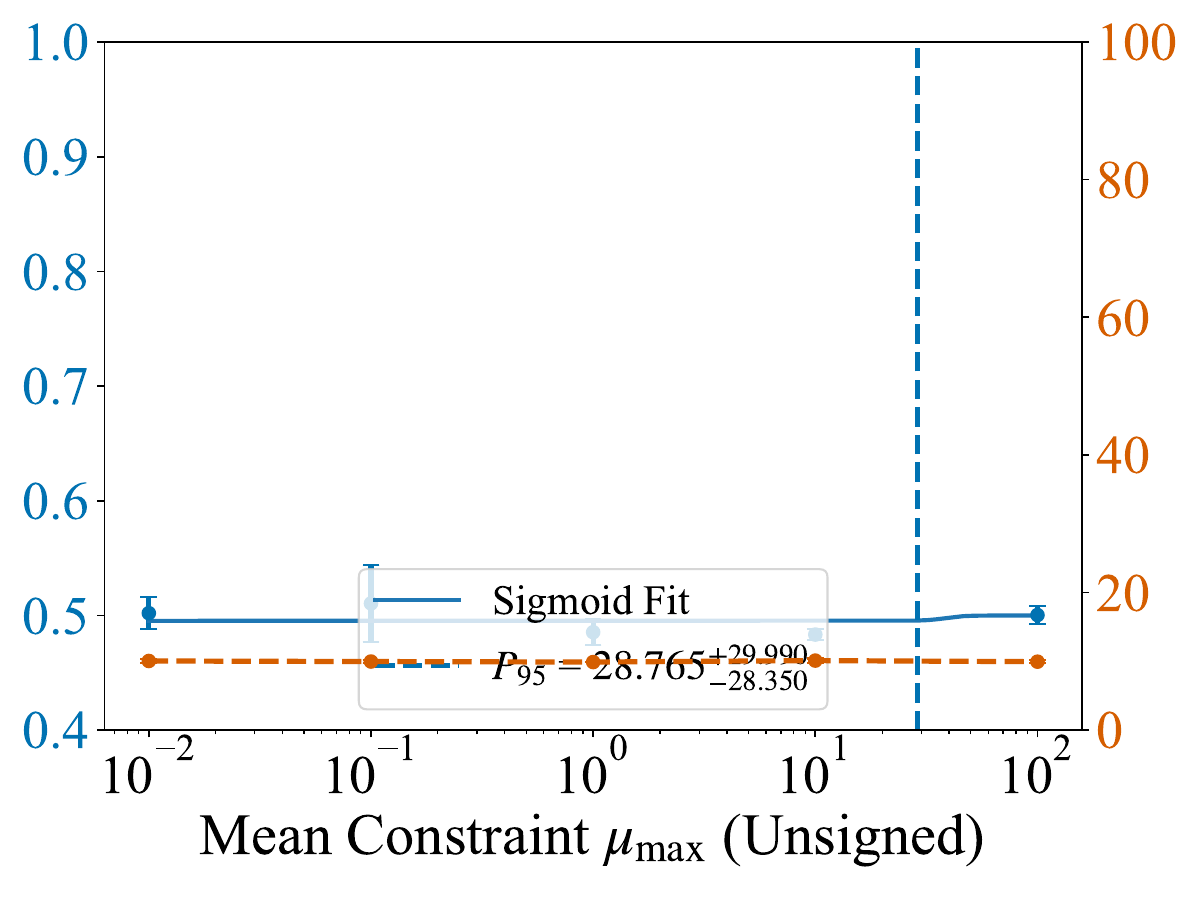}
        \includegraphics[width=\linewidth]
            {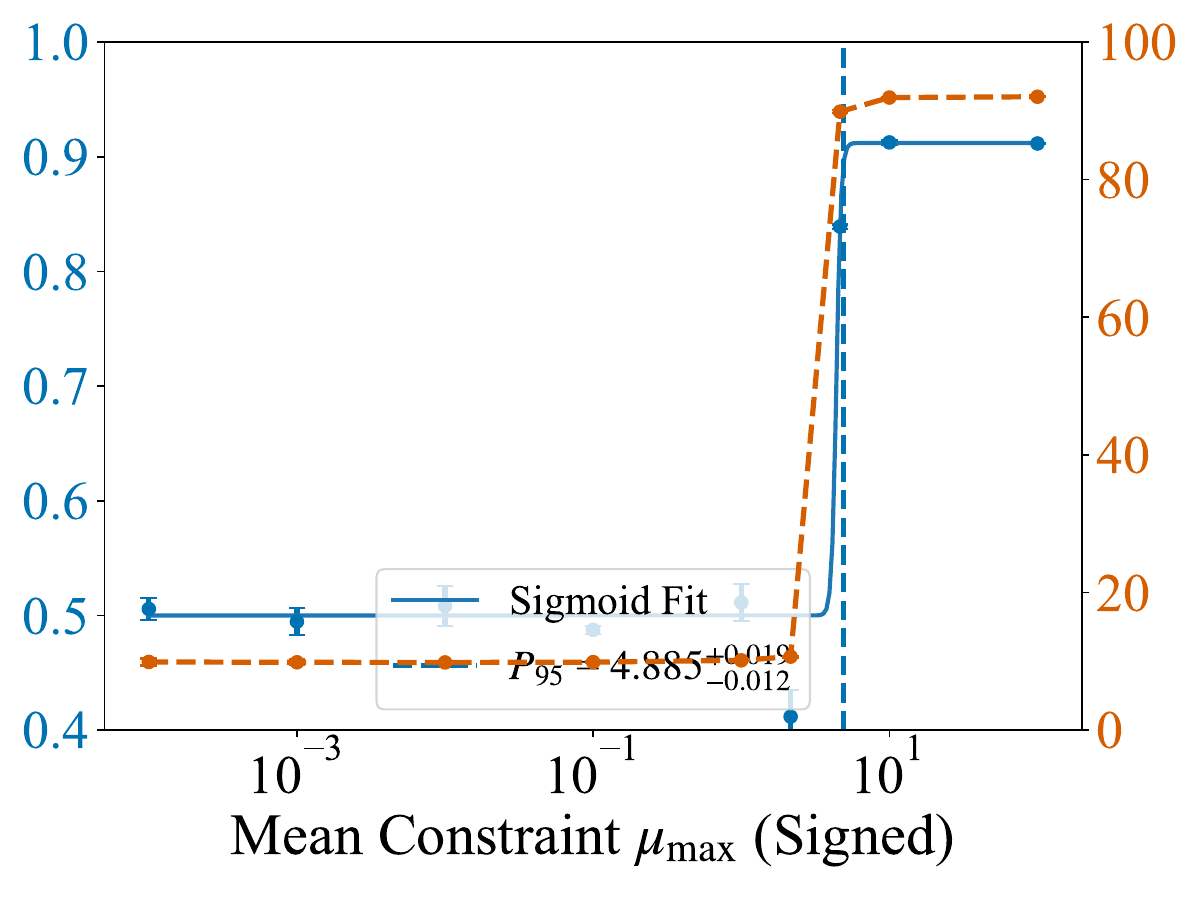}
        \includegraphics[width=\linewidth]
            {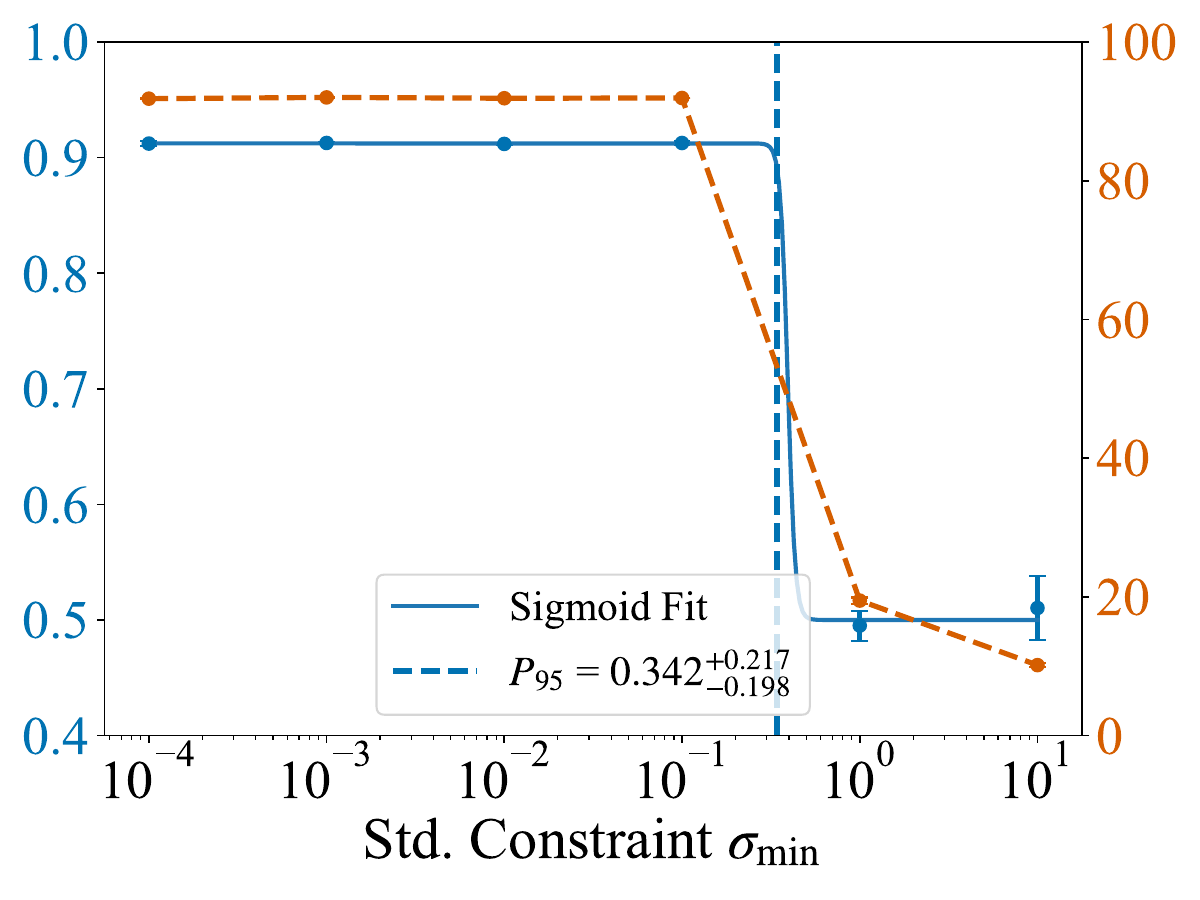}
        \includegraphics[width=\linewidth]
            {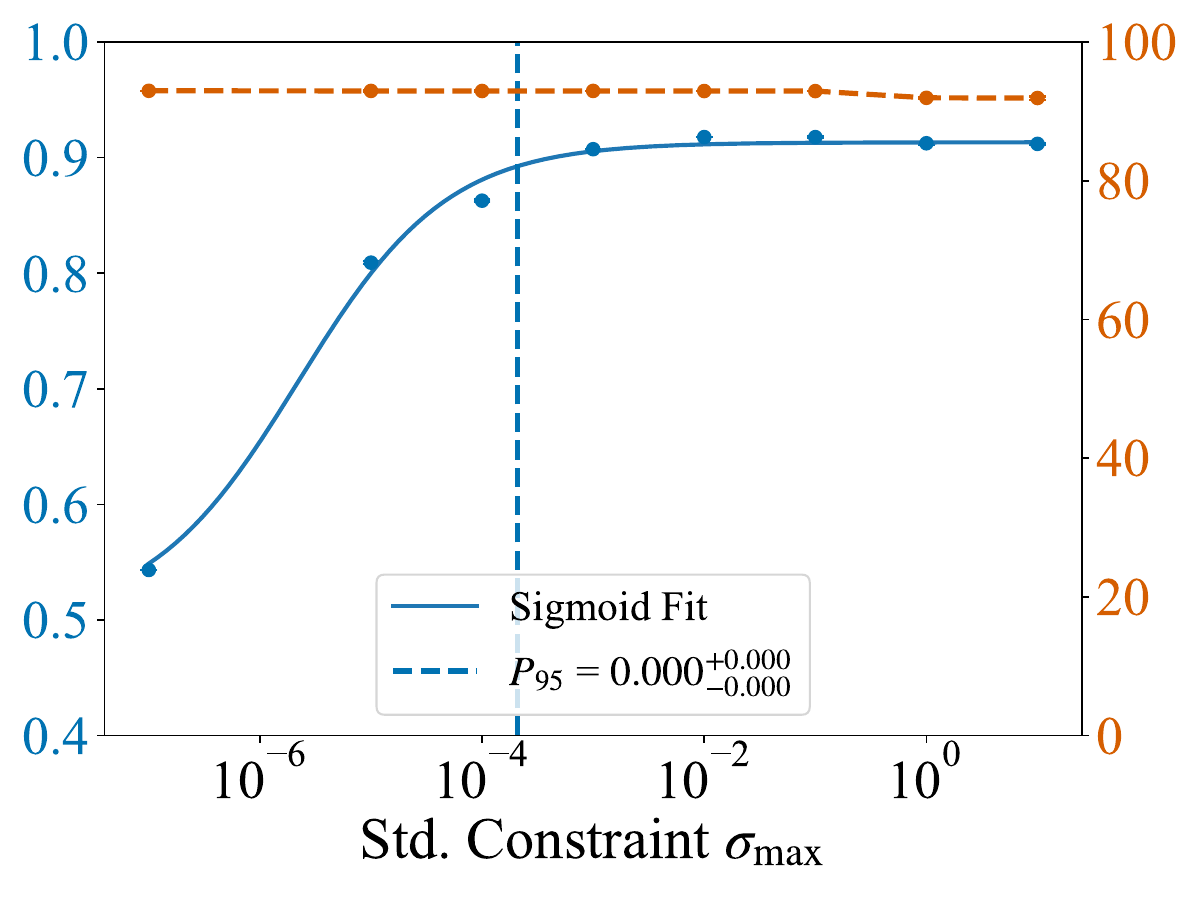}
    \end{subfigure}
    \hfill
    \begin{subfigure}[t]{0.235\textwidth}
        \centering
        \caption{A-Add}
        \includegraphics[width=\linewidth]
            {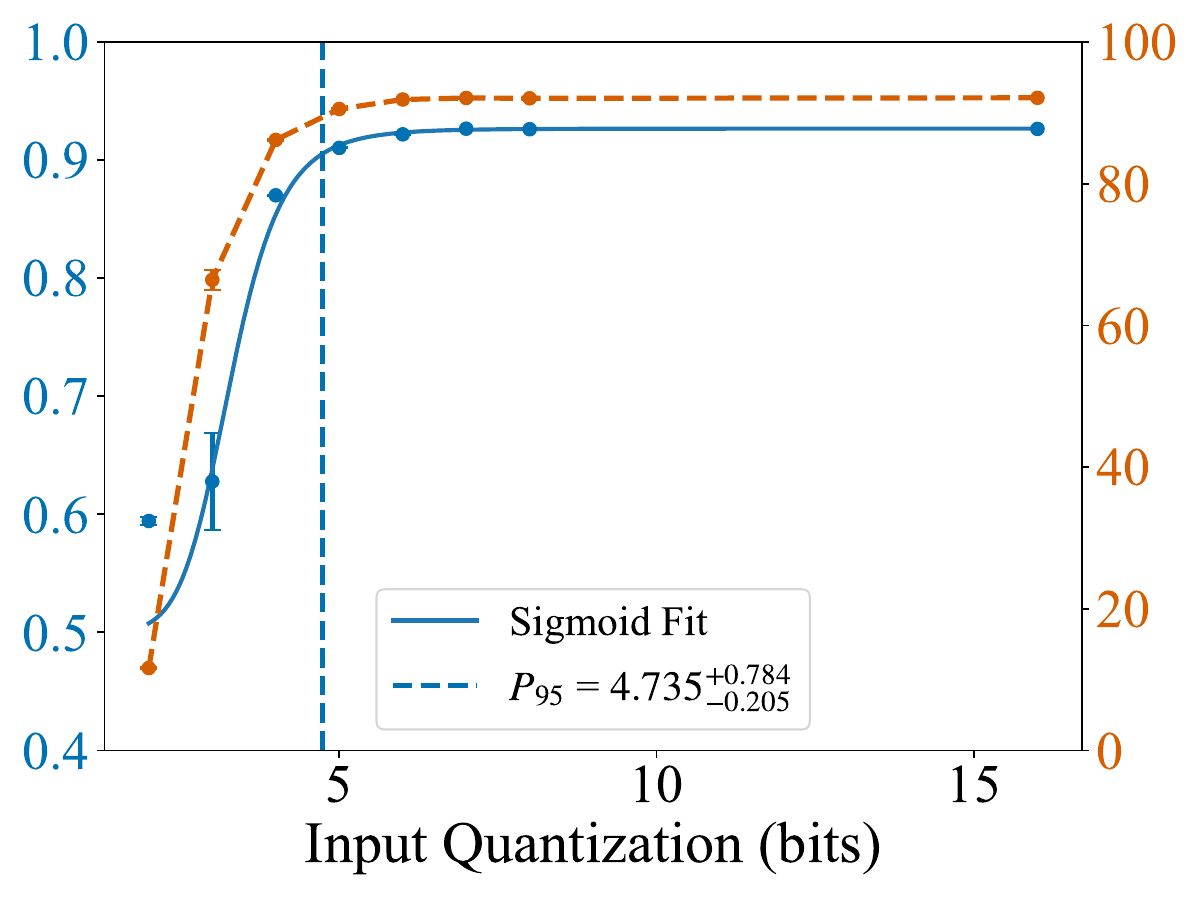}
        \includegraphics[width=\linewidth]
            {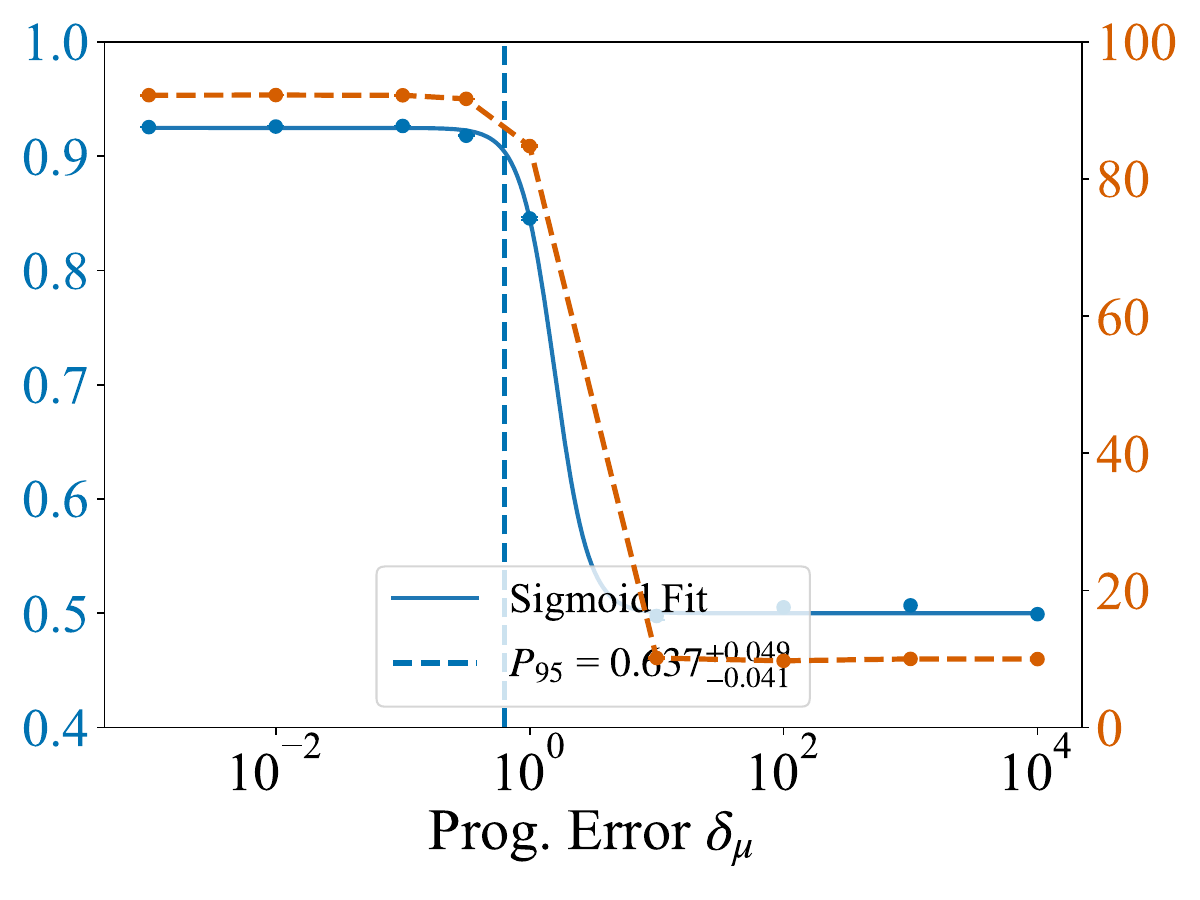}
        \includegraphics[width=\linewidth]
            {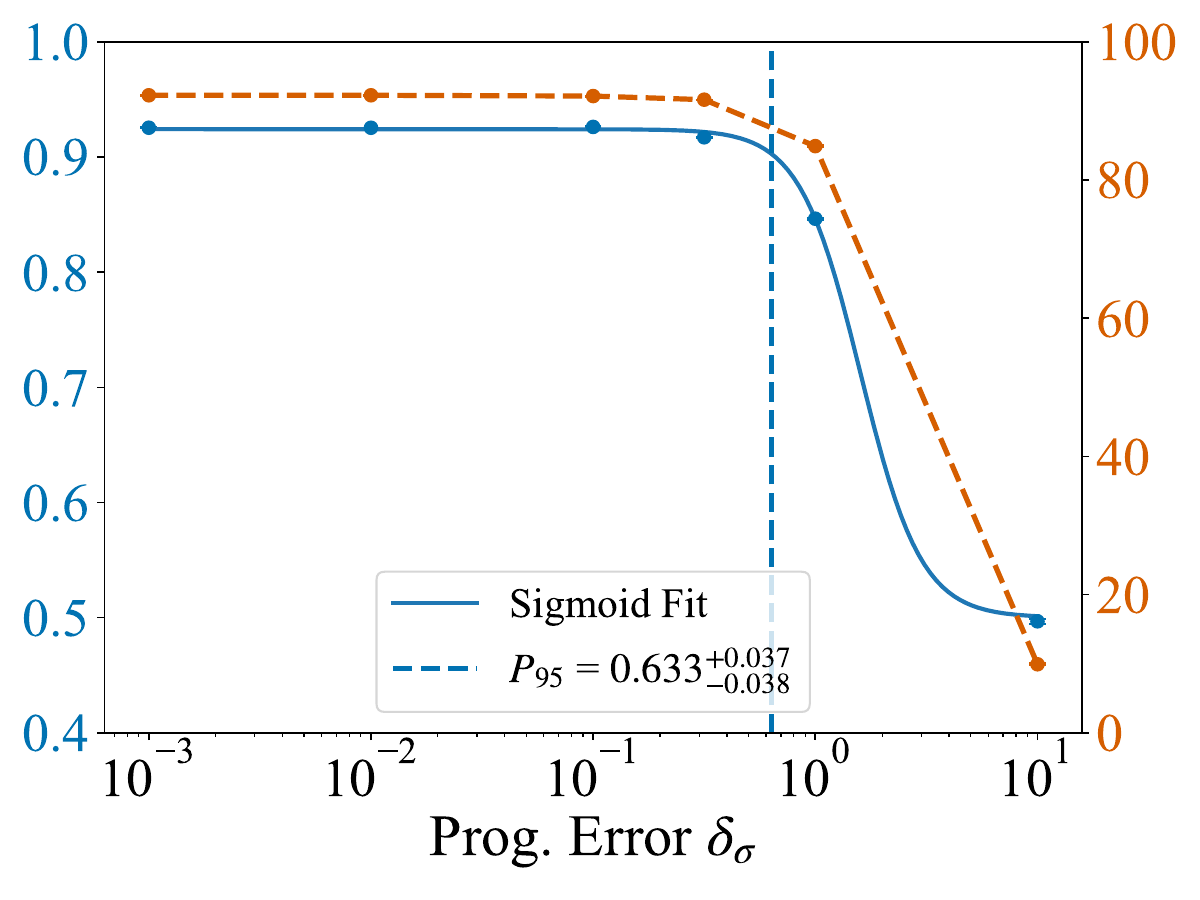}
        \includegraphics[width=\linewidth]
            {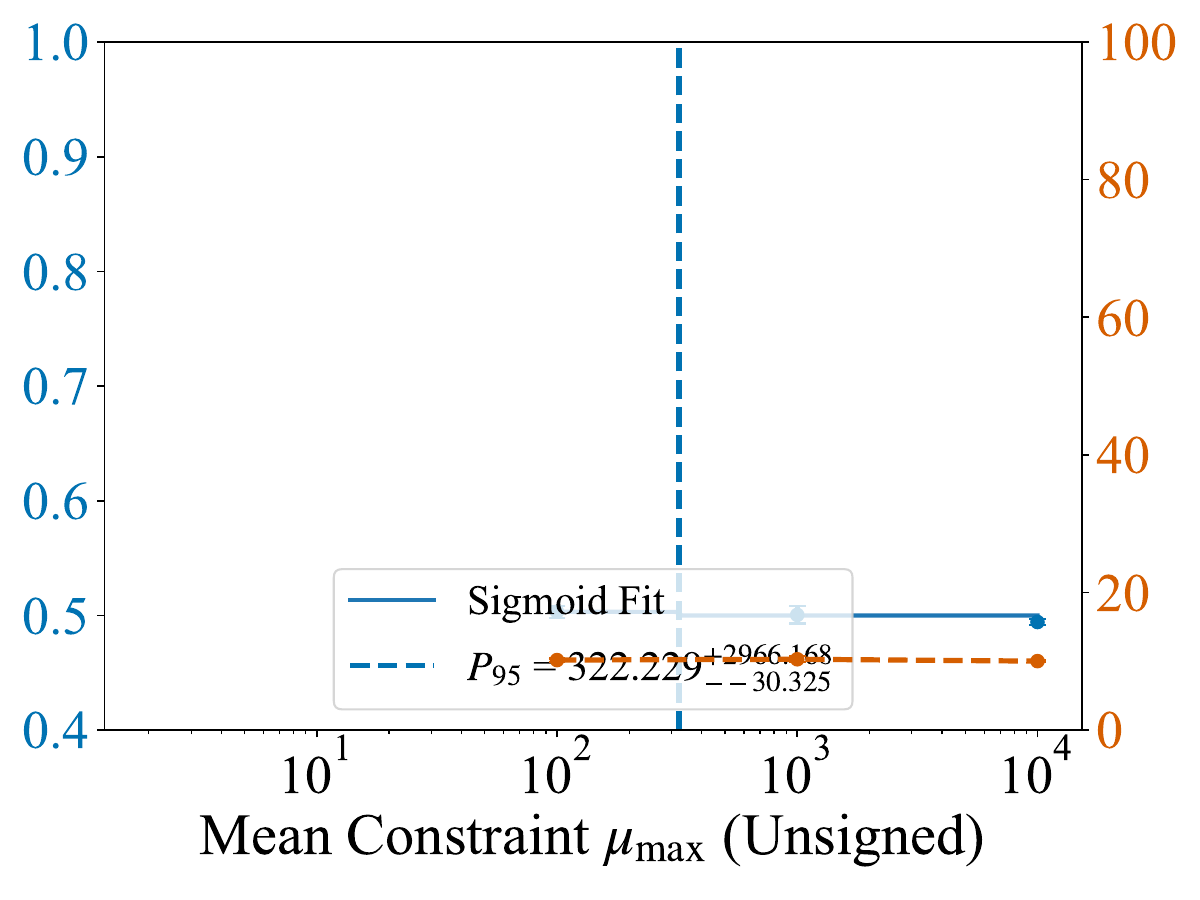}
        \includegraphics[width=\linewidth]
            {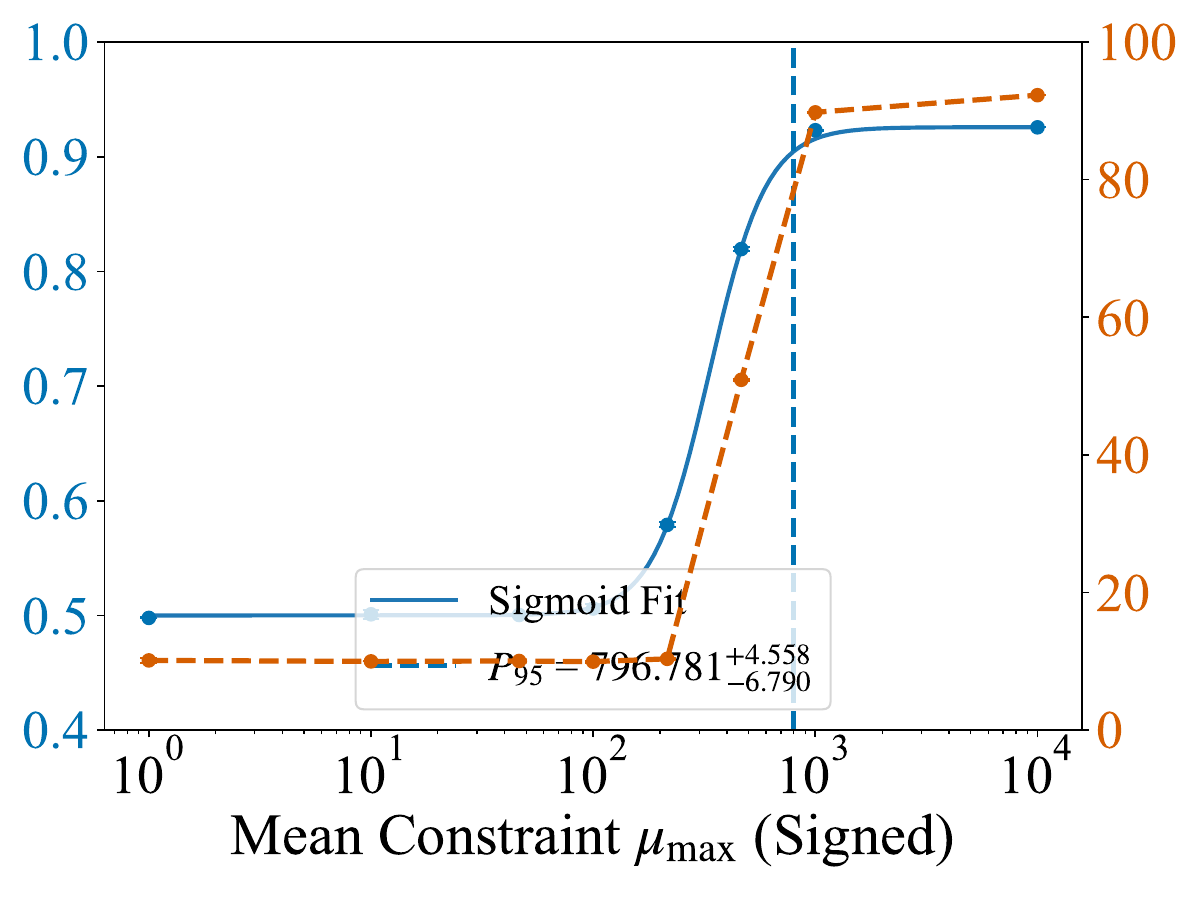}
        \includegraphics[width=\linewidth]
            {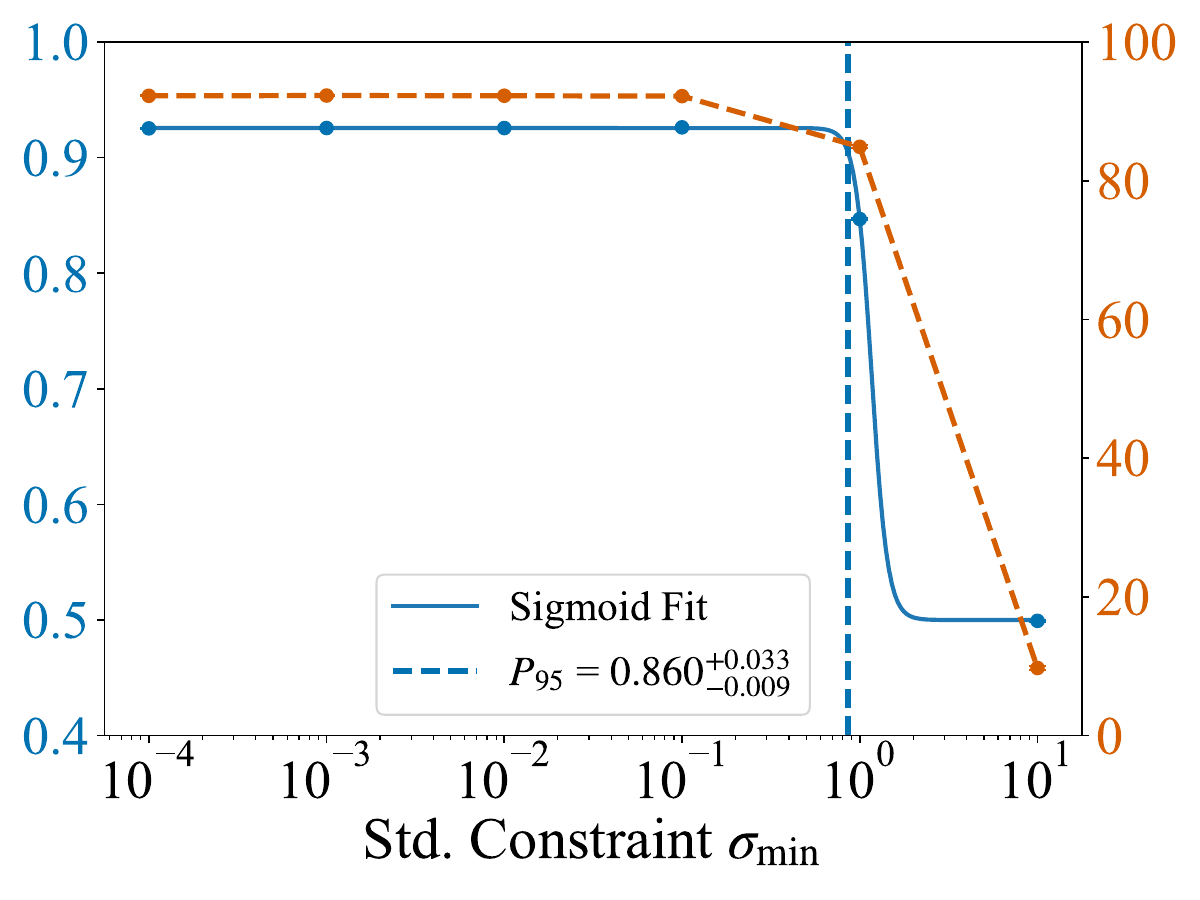}
        \includegraphics[width=\linewidth]
            {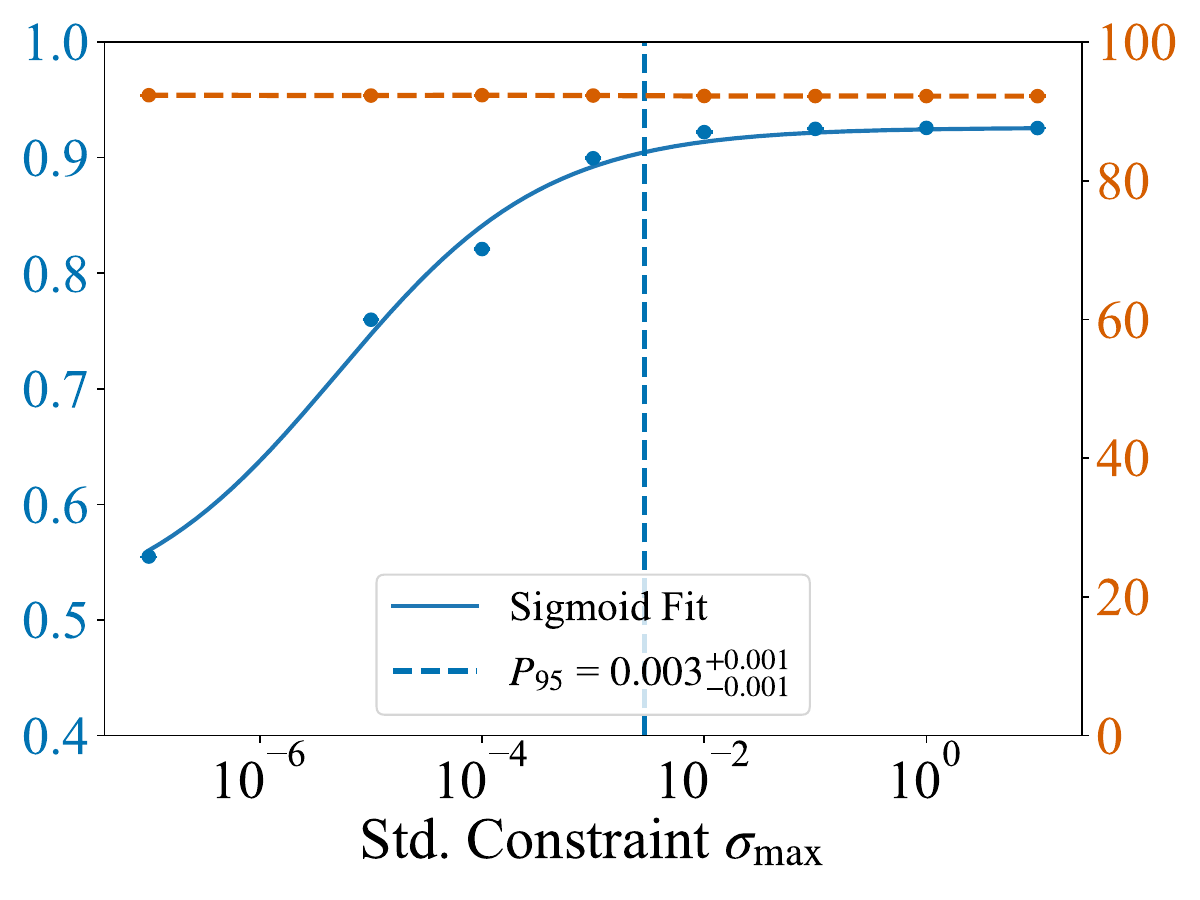}
    \end{subfigure}
    \hfill
    \begin{subfigure}[t]{0.235\textwidth}
        \centering
        \caption{A-Mul}
        \includegraphics[width=\linewidth]
            {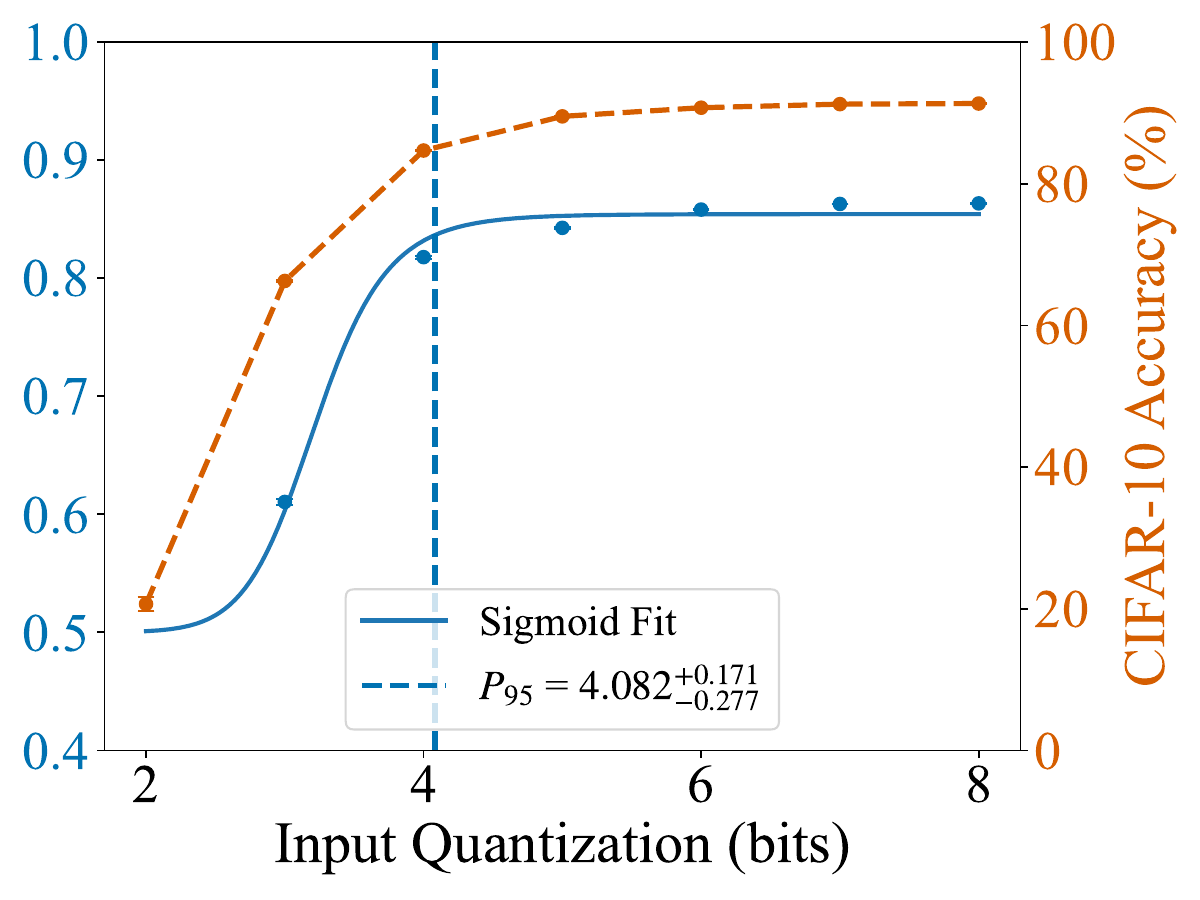}
        \includegraphics[width=\linewidth]
            {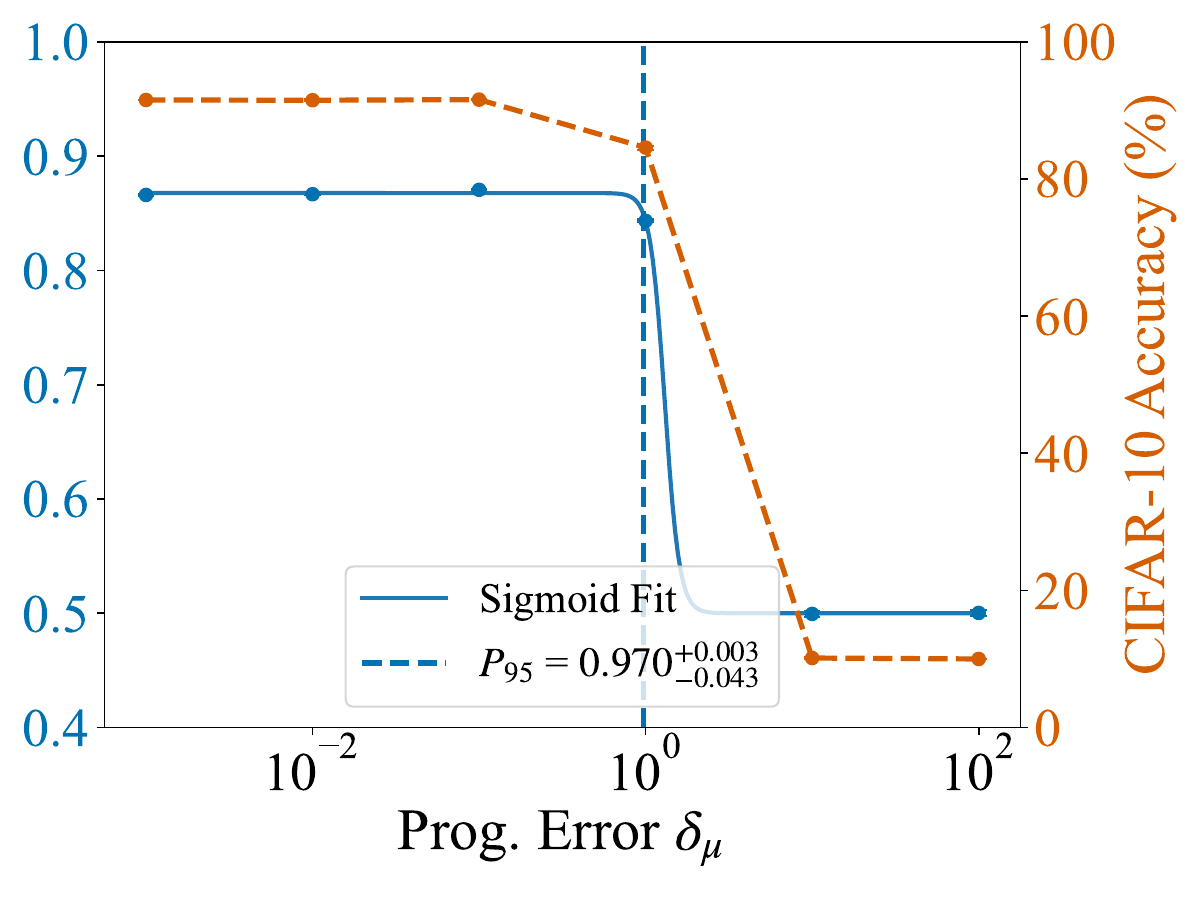}
        \includegraphics[width=\linewidth]
            {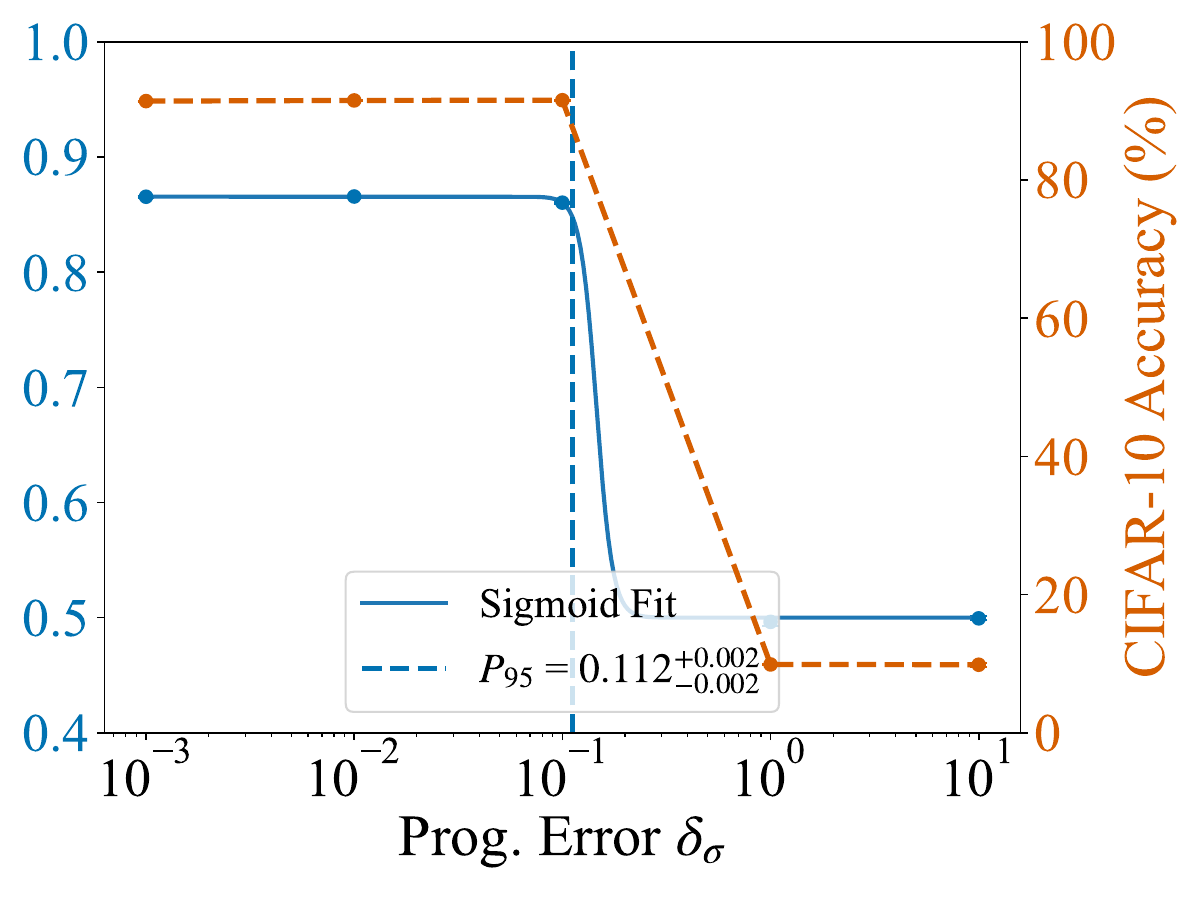}
        \includegraphics[width=\linewidth]
            {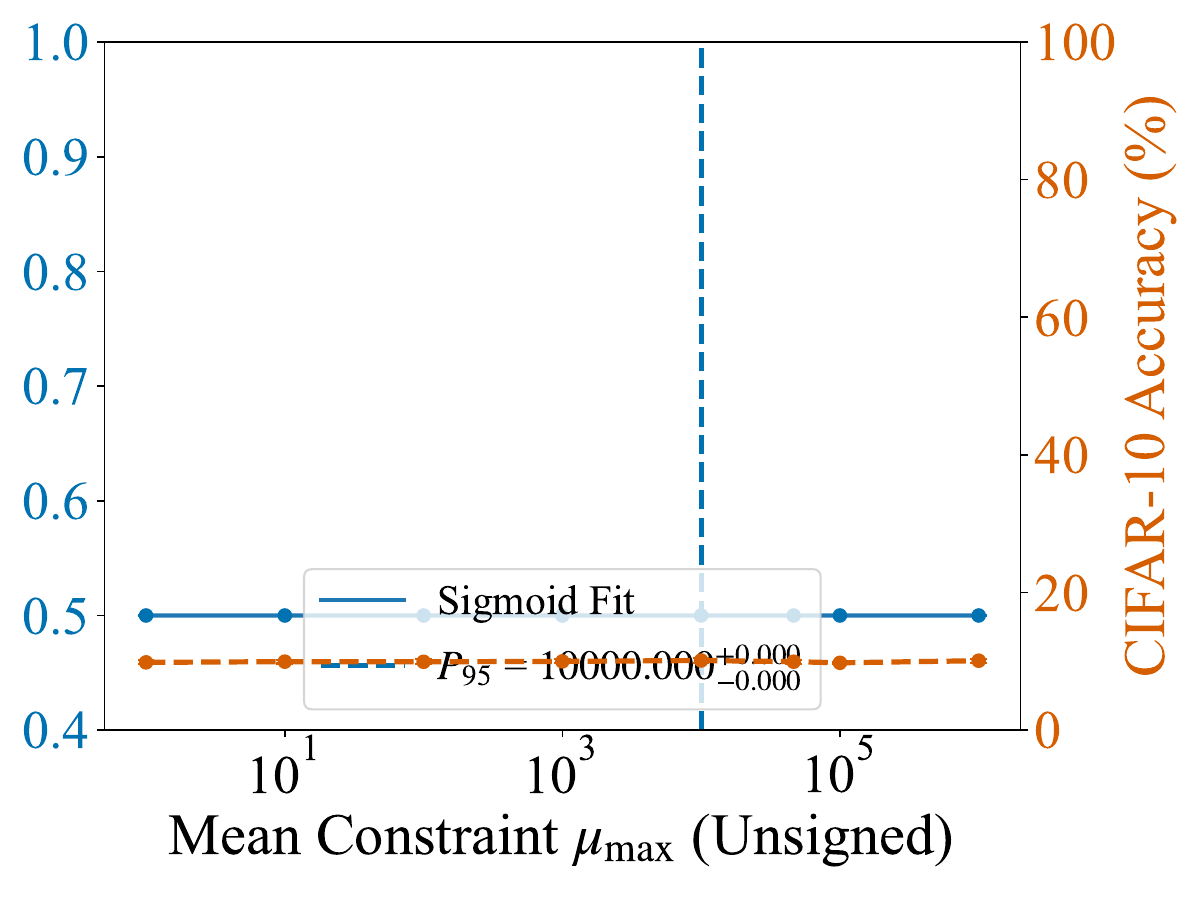}
        \includegraphics[width=\linewidth]
            {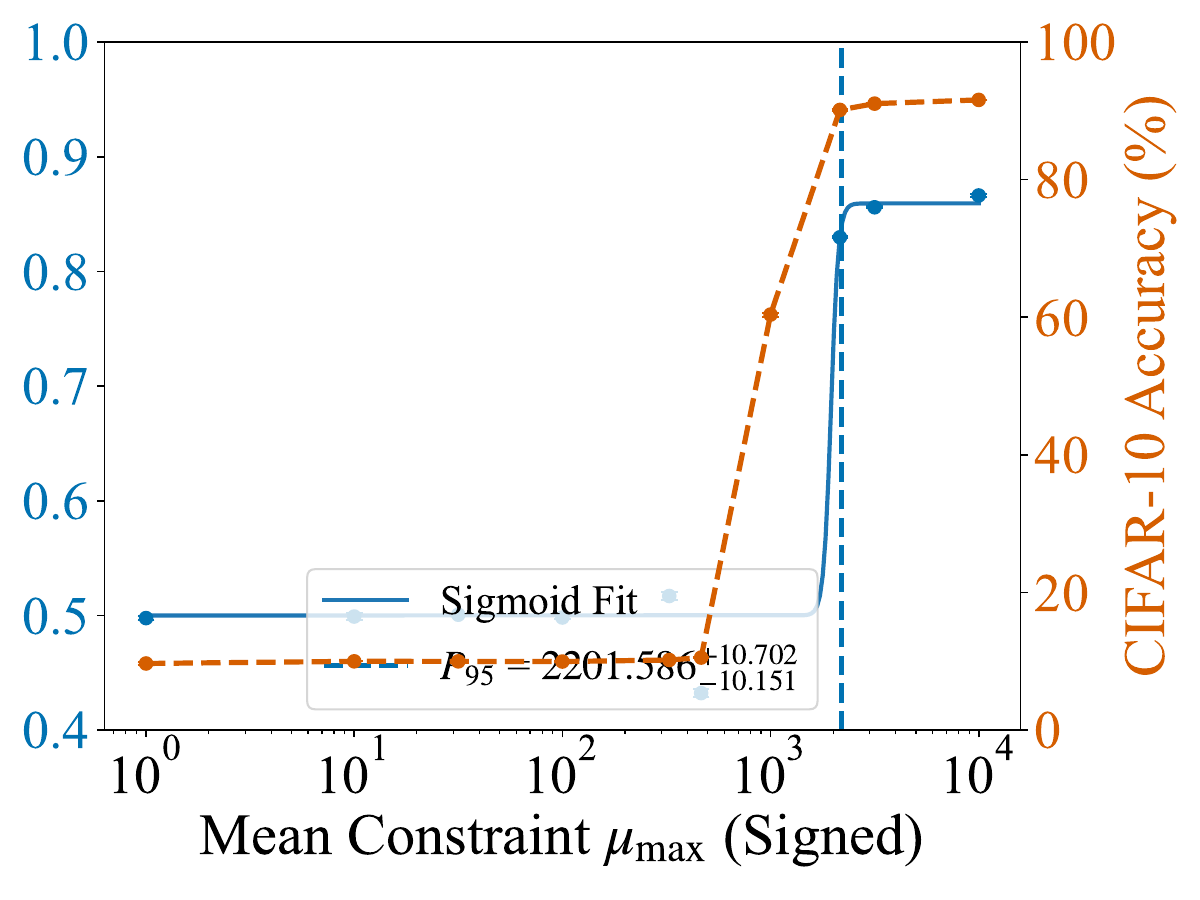}
        \includegraphics[width=\linewidth]
            {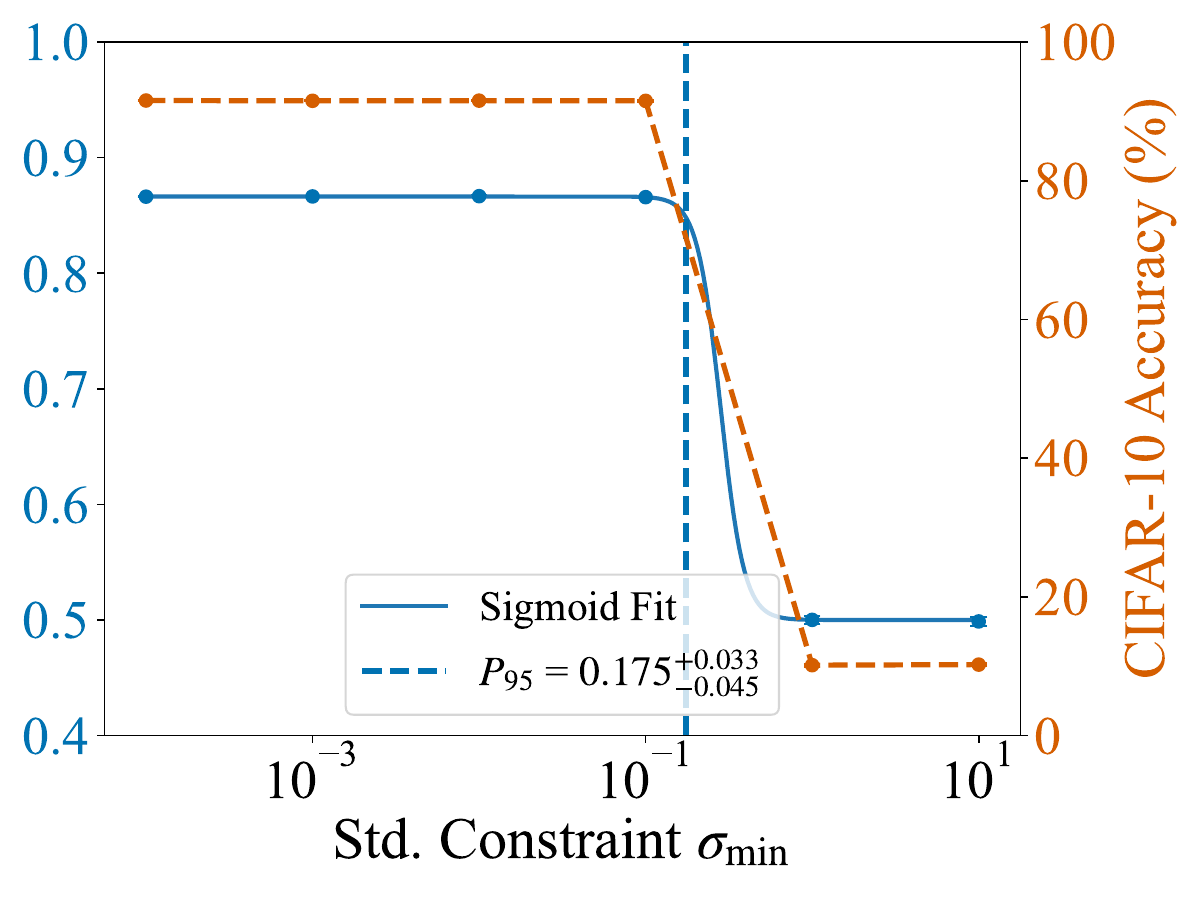}
        \includegraphics[width=\linewidth]
            {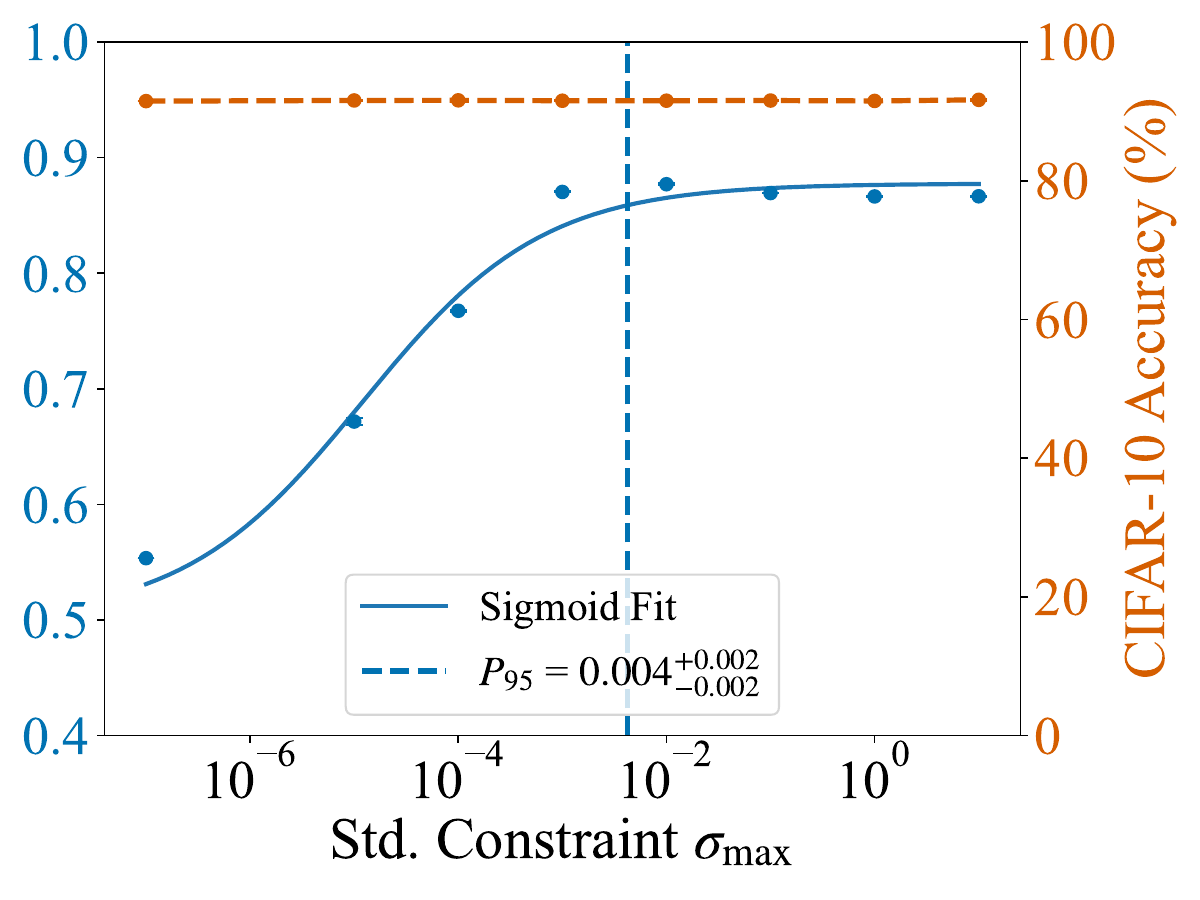}
    \end{subfigure}

    \caption{Ablation study of individual hardware constraints applied only
    during inference to the unconstrained baseline model. Results are reported
    for ResNet-18 on CIFAR-10.}
    \label{fig:AB-cifar-eval}
\end{figure*}

Figure~\ref{fig:AB-cifar} presents the results for ResNet-18 on CIFAR-10 with HW-aware training whereas Figure~\ref{fig:AB-cifar-eval} reports the results without HW-aware training. In the latter setting, we load checkpoints from the optimized unconstrained baseline and apply the constraints only during inference.  The results show that models without HW-aware training are consistently less robust and tolerate substantially tighter constraints.

\begin{figure*}[p]
    \centering

    \begin{subfigure}[t]{0.235\textwidth}
        \centering
        \caption{W-Add}
        \includegraphics[width=\linewidth]
            {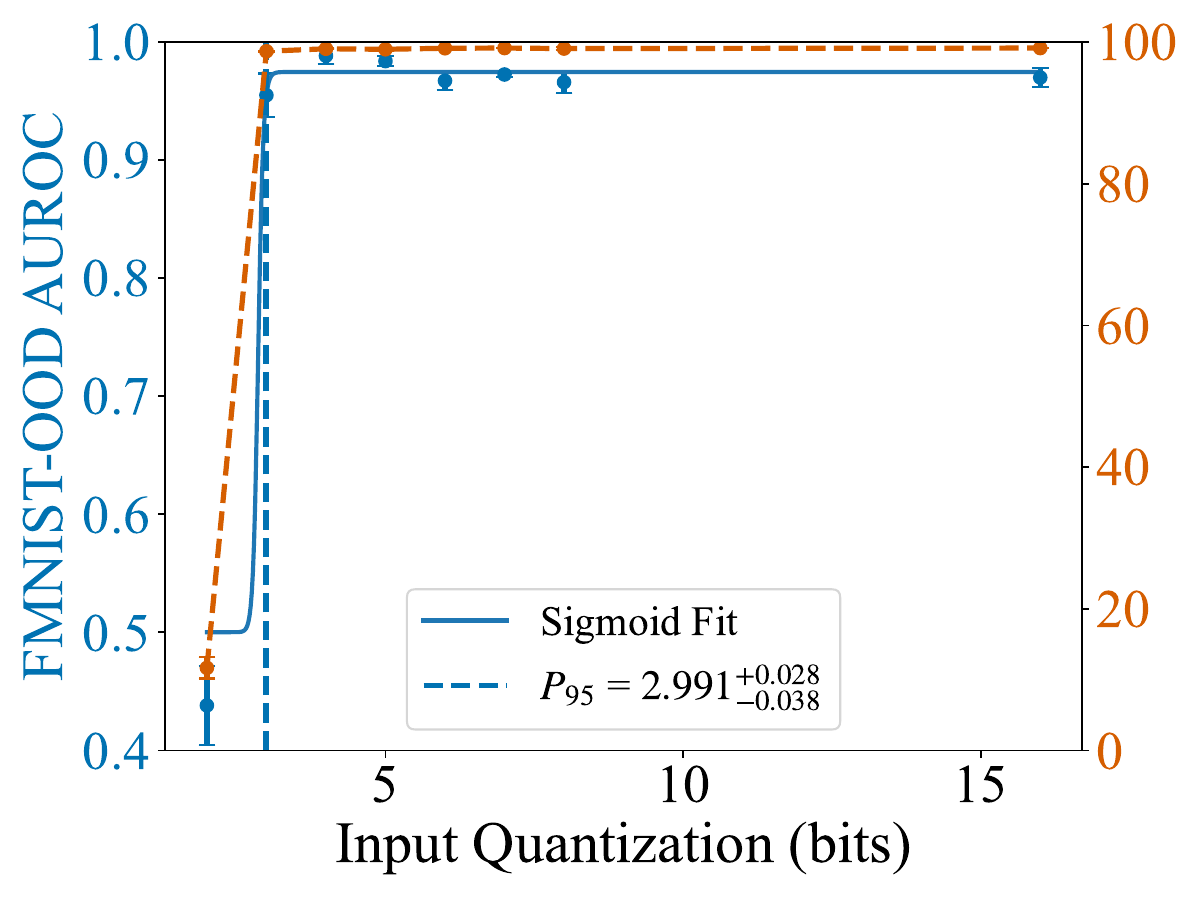}
        \includegraphics[width=\linewidth]
            {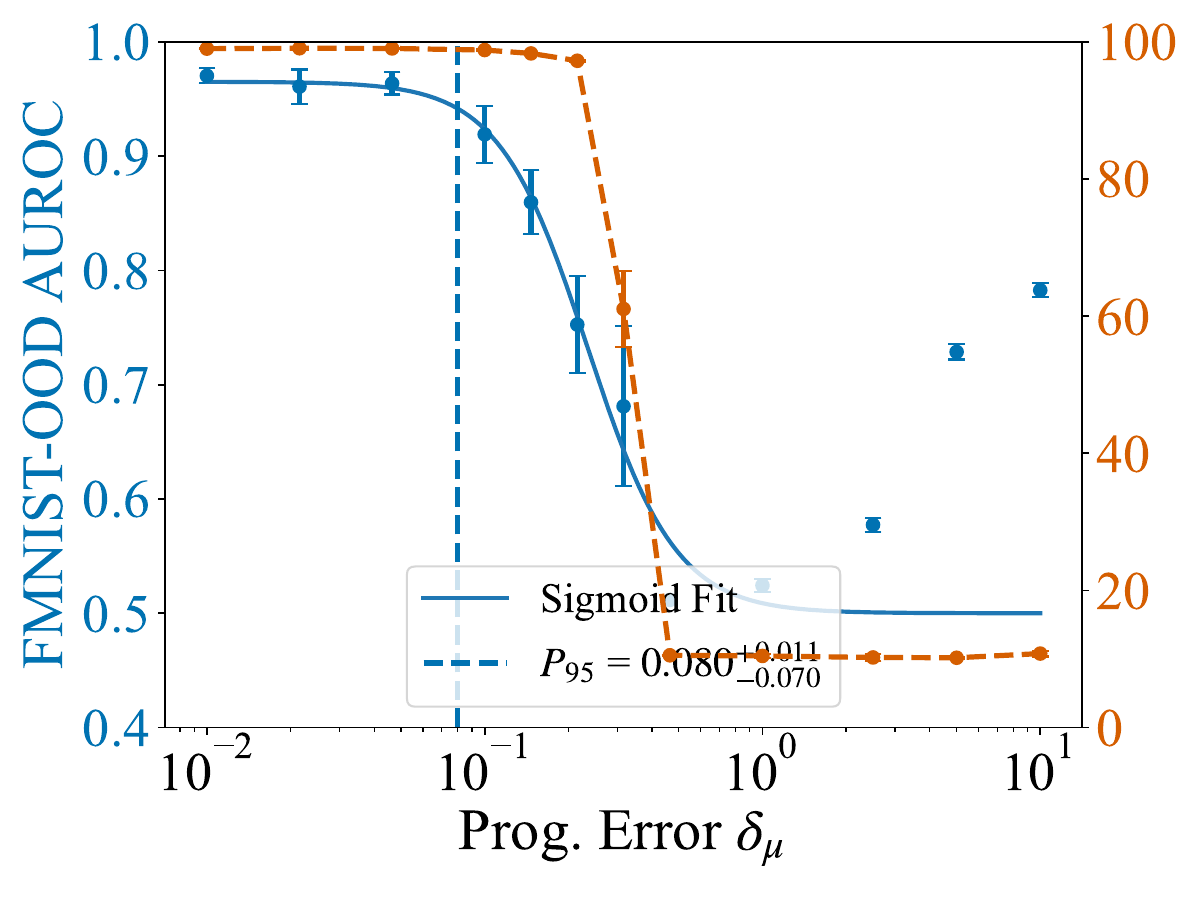}
        \includegraphics[width=\linewidth]
            {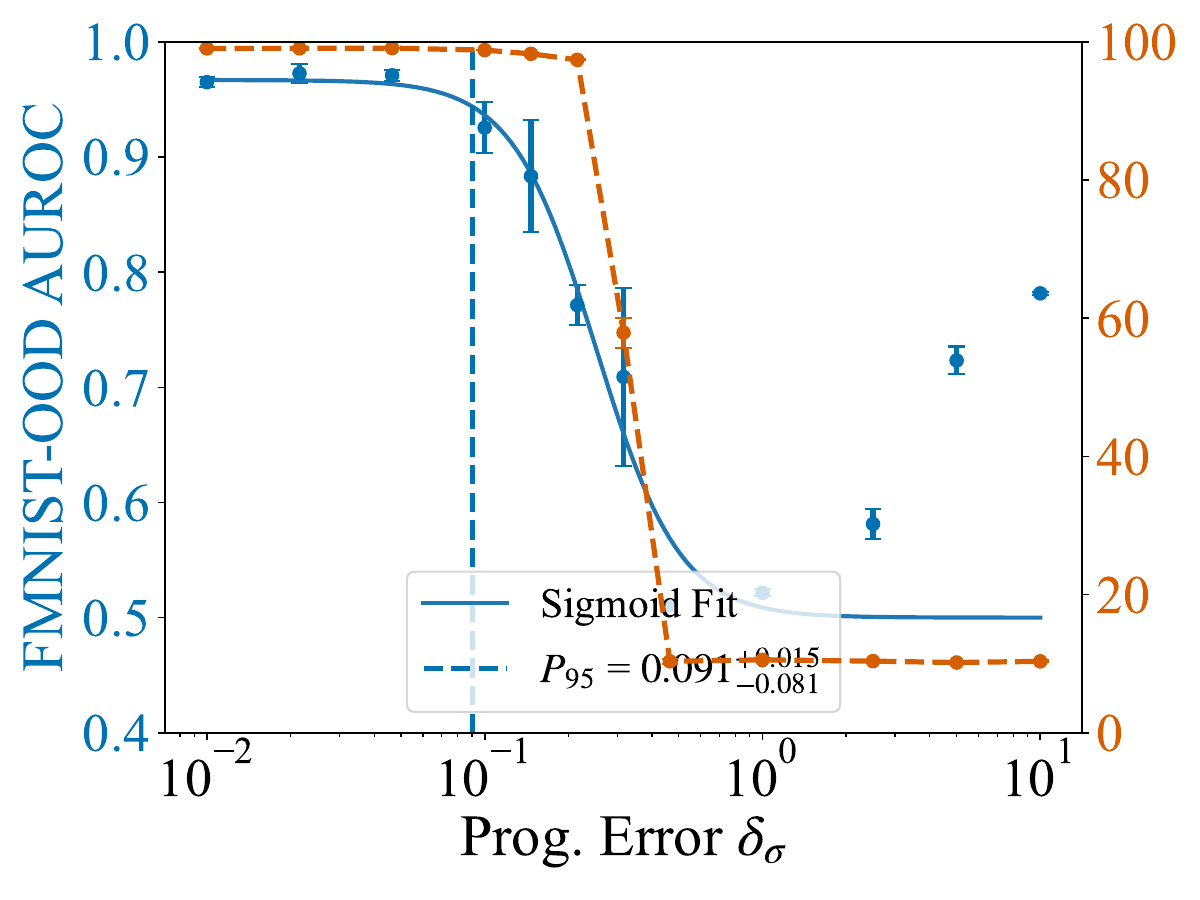}
        \includegraphics[width=\linewidth]
            {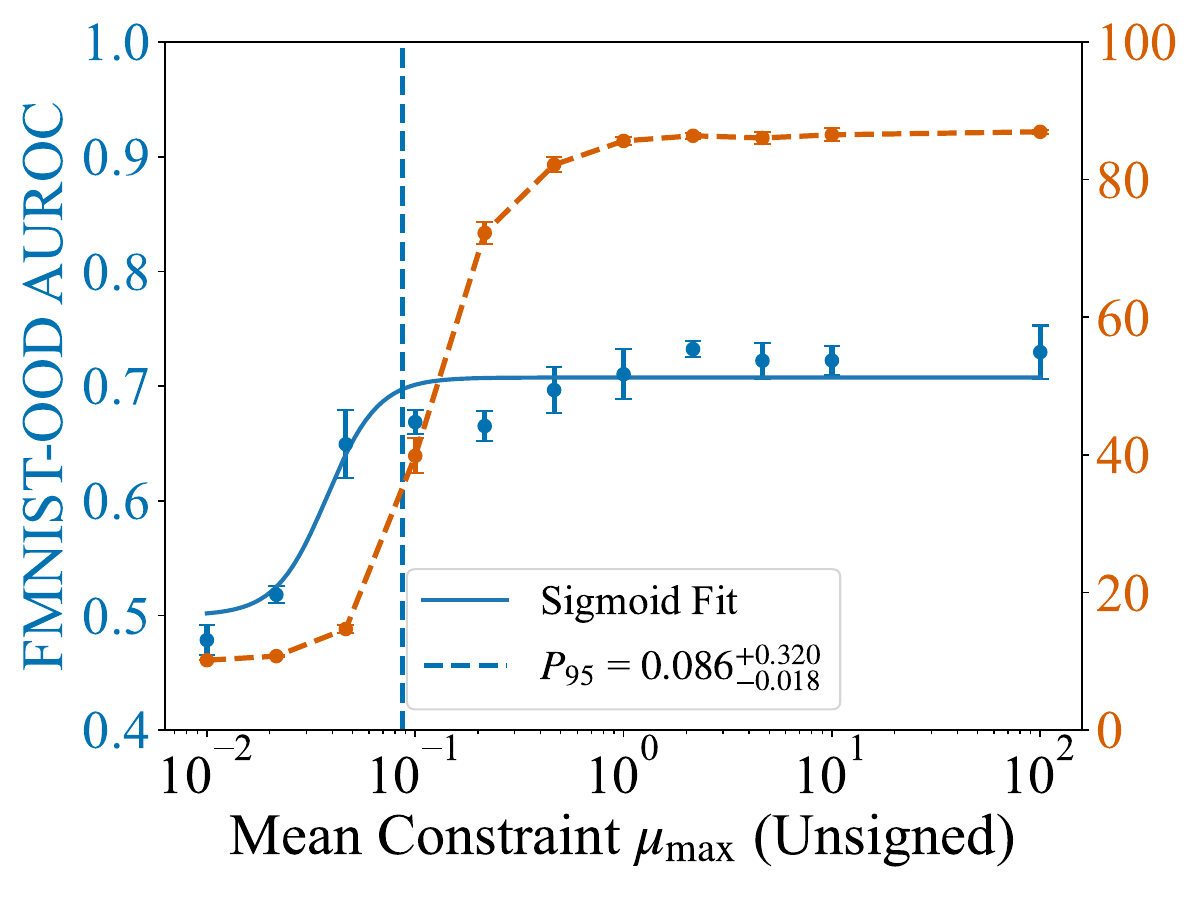}
        \includegraphics[width=\linewidth]
            {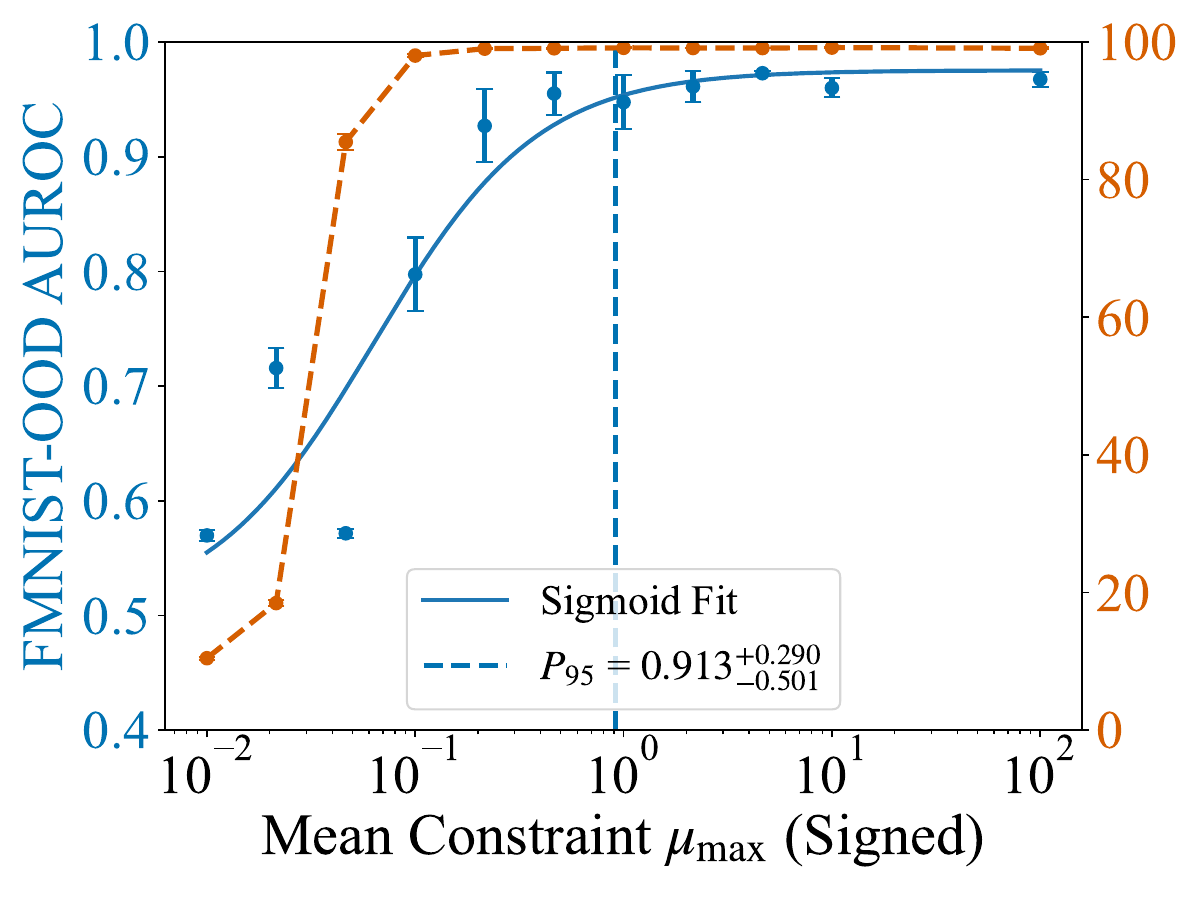}
        \includegraphics[width=\linewidth]
            {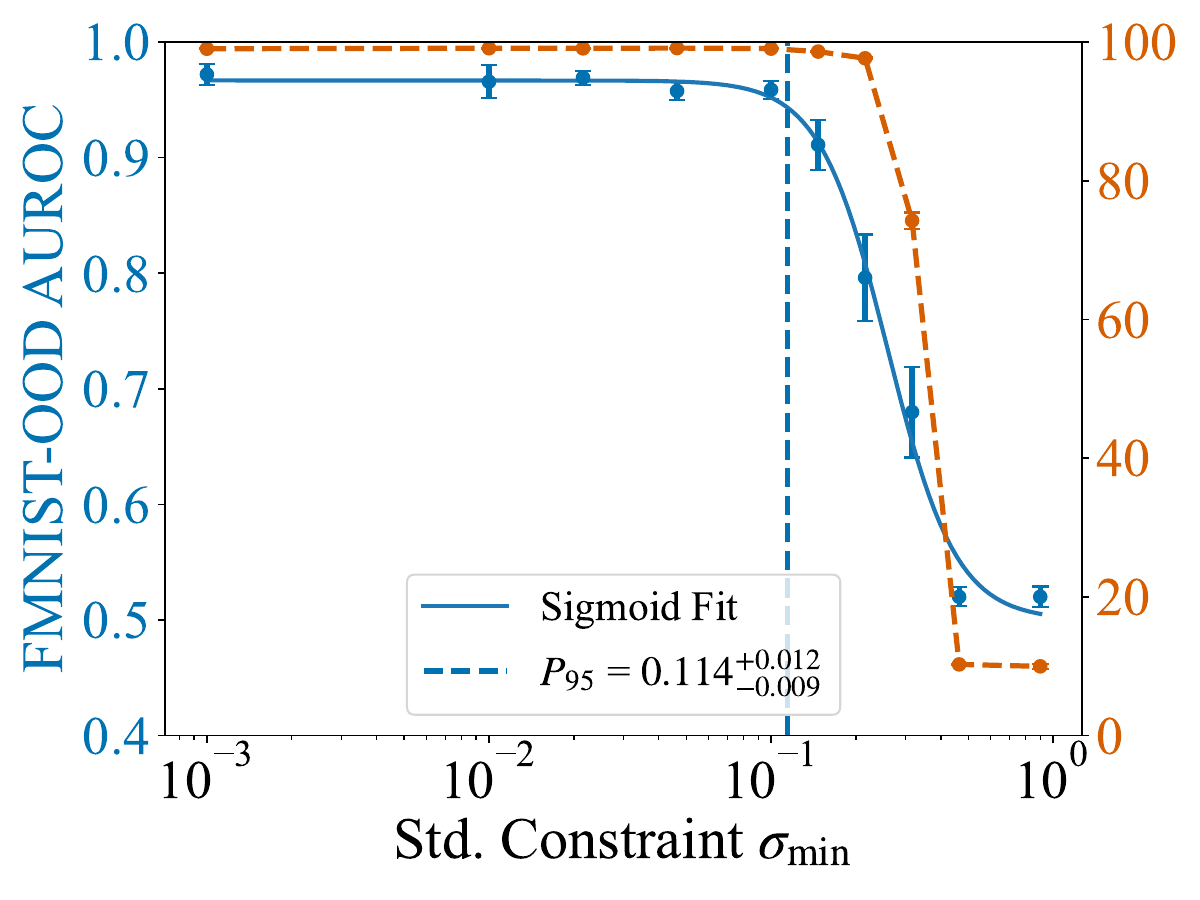}
        \includegraphics[width=\linewidth]
            {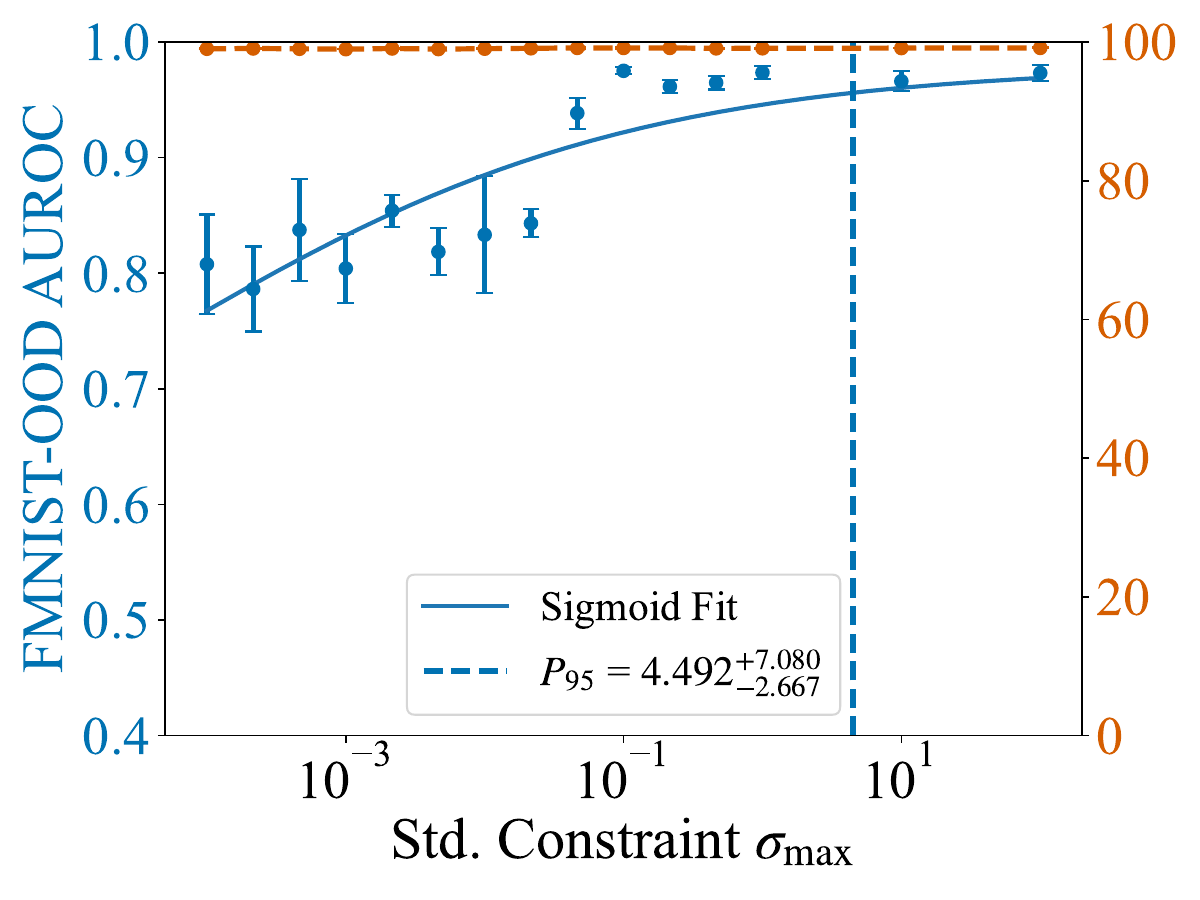}
    \end{subfigure}
    \hfill
    \begin{subfigure}[t]{0.235\textwidth}
        \centering
        \caption{W-Mul}
        \includegraphics[width=\linewidth]
            {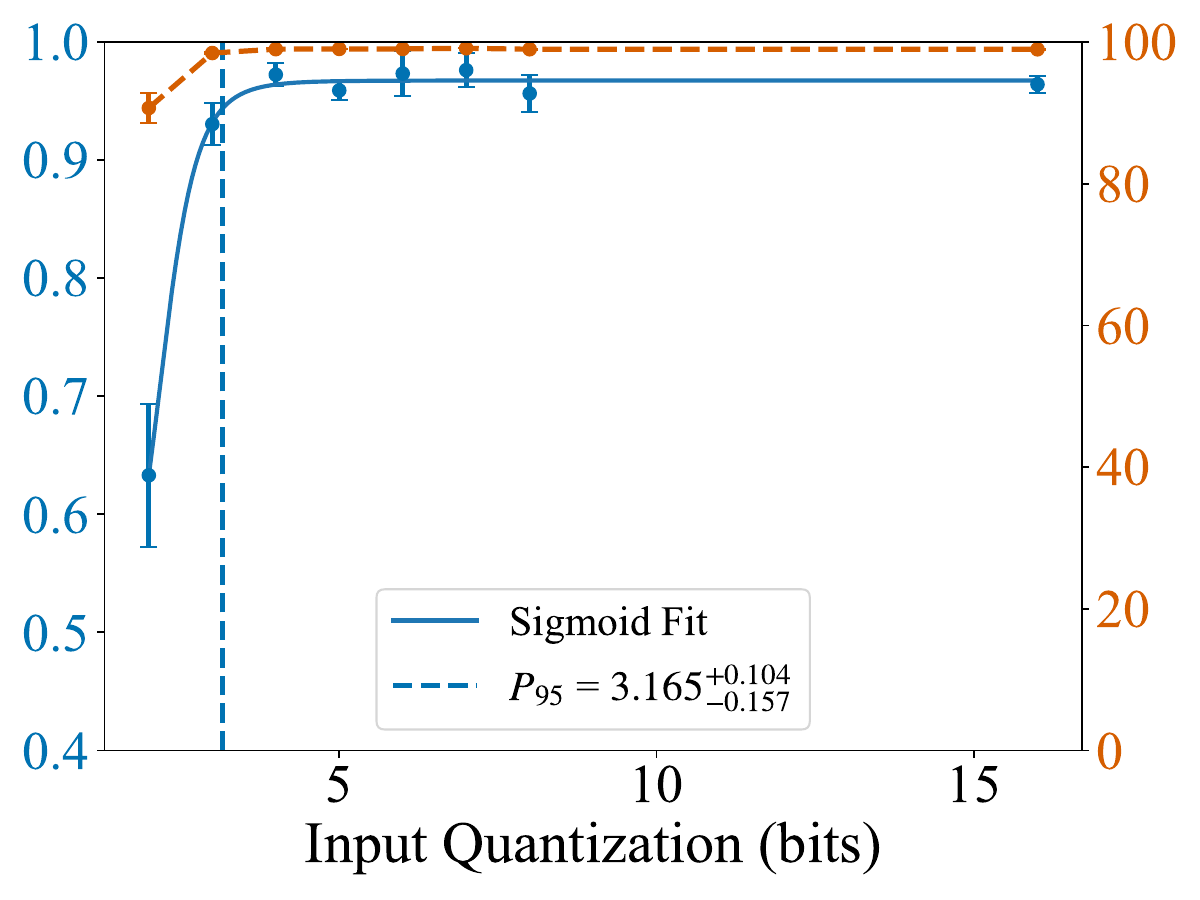}
        \includegraphics[width=\linewidth]
            {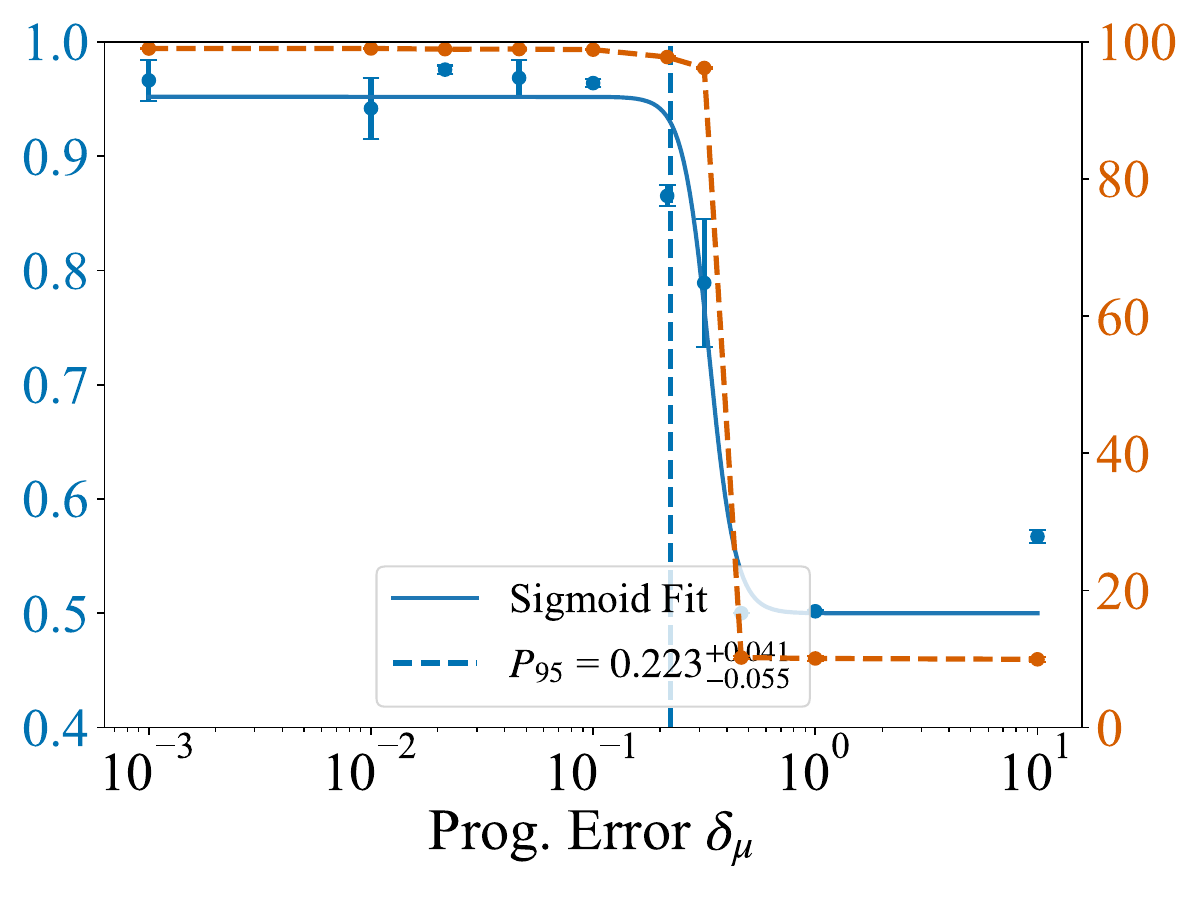}
        \includegraphics[width=\linewidth]
            {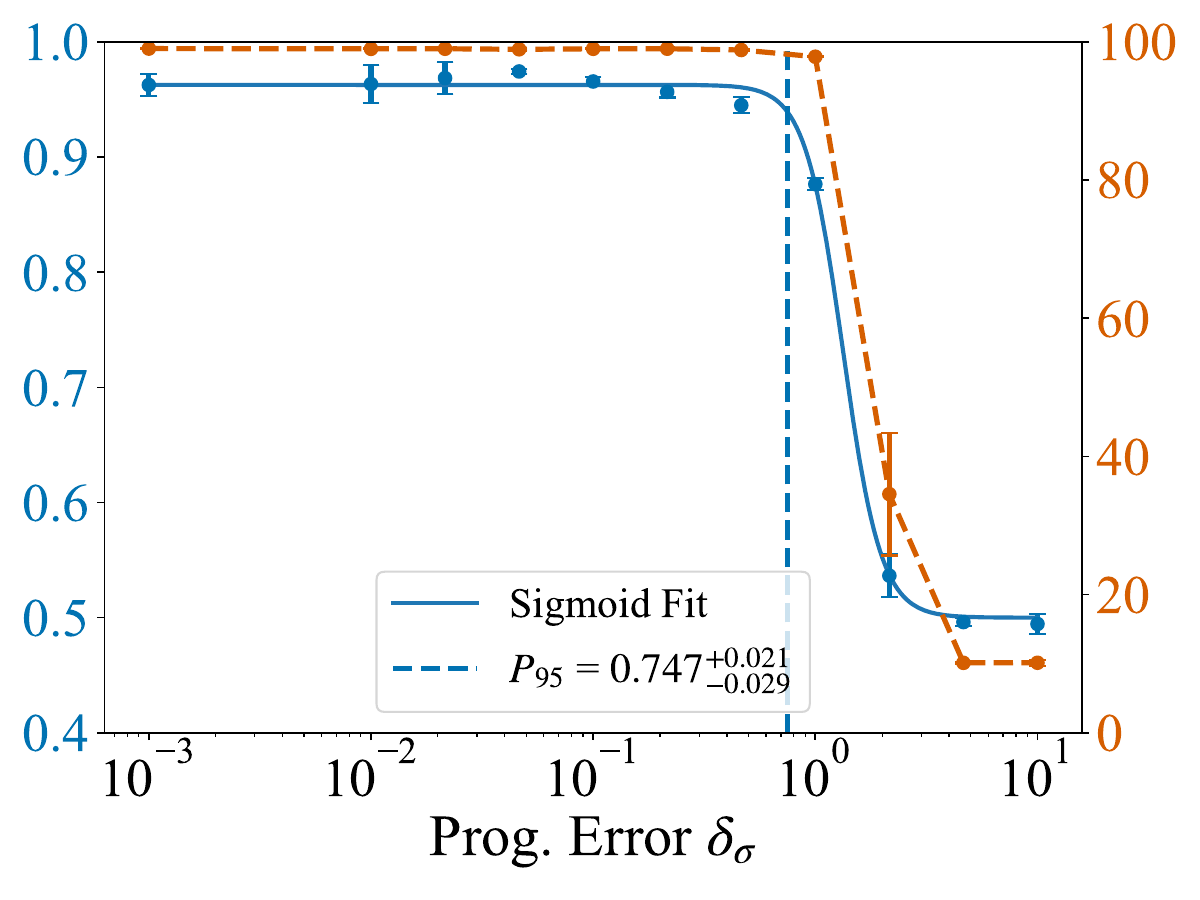}
        \includegraphics[width=\linewidth]
            {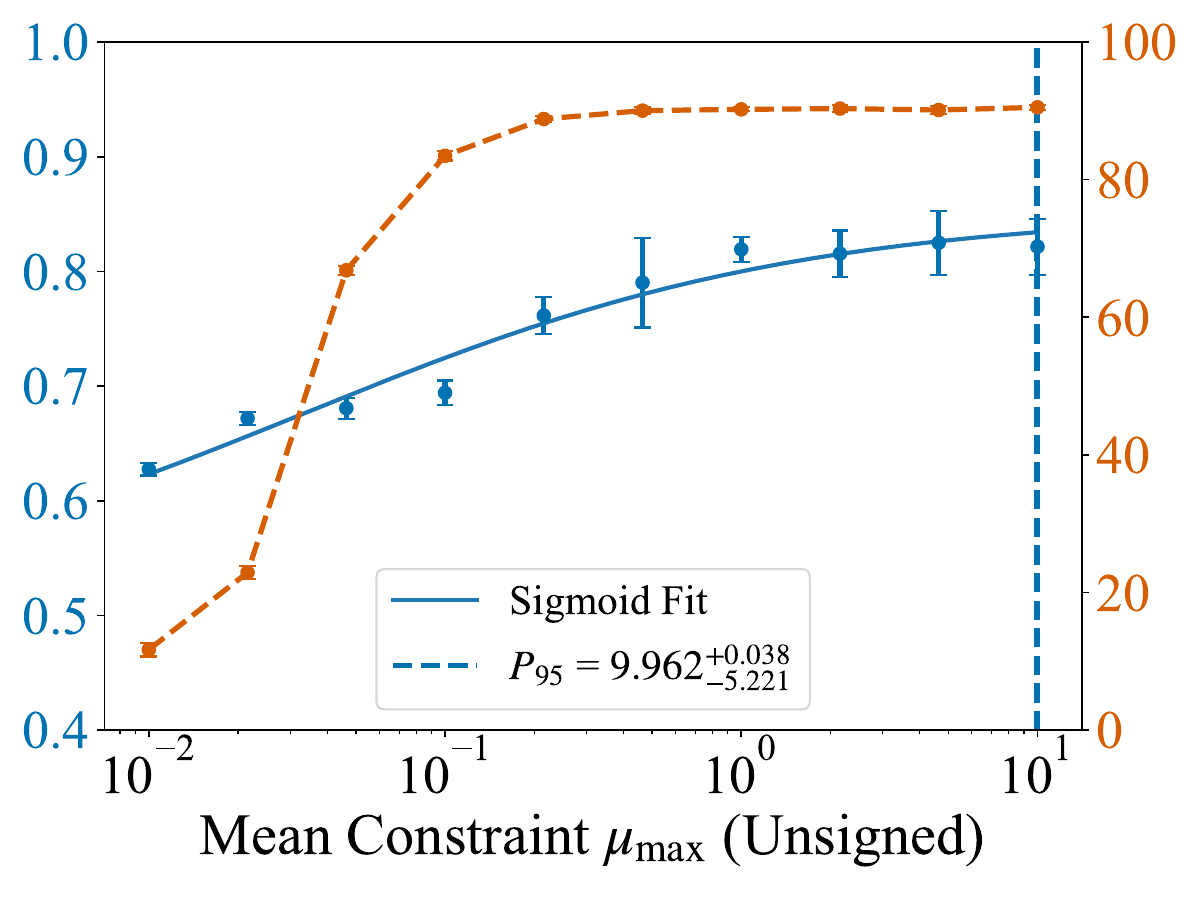}
        \includegraphics[width=\linewidth]
            {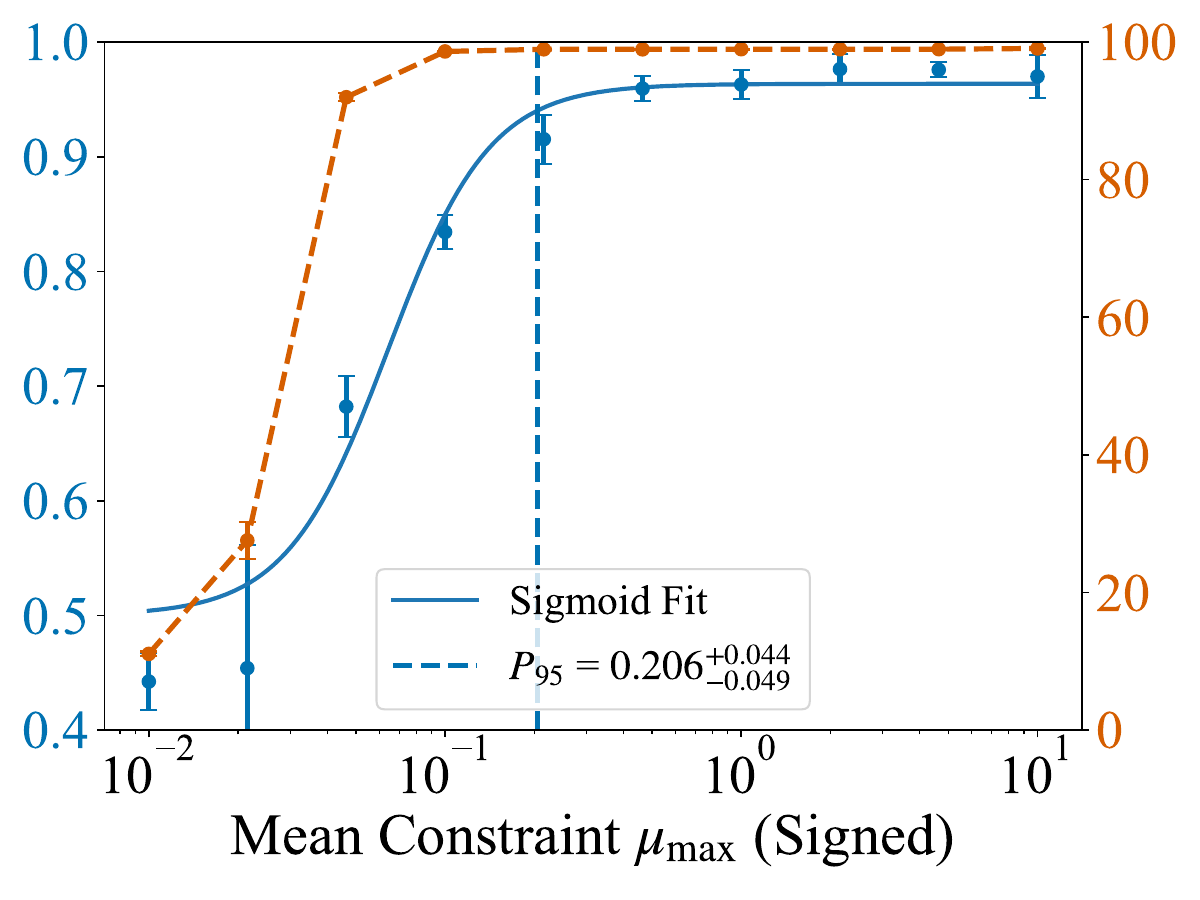}
        \includegraphics[width=\linewidth]
            {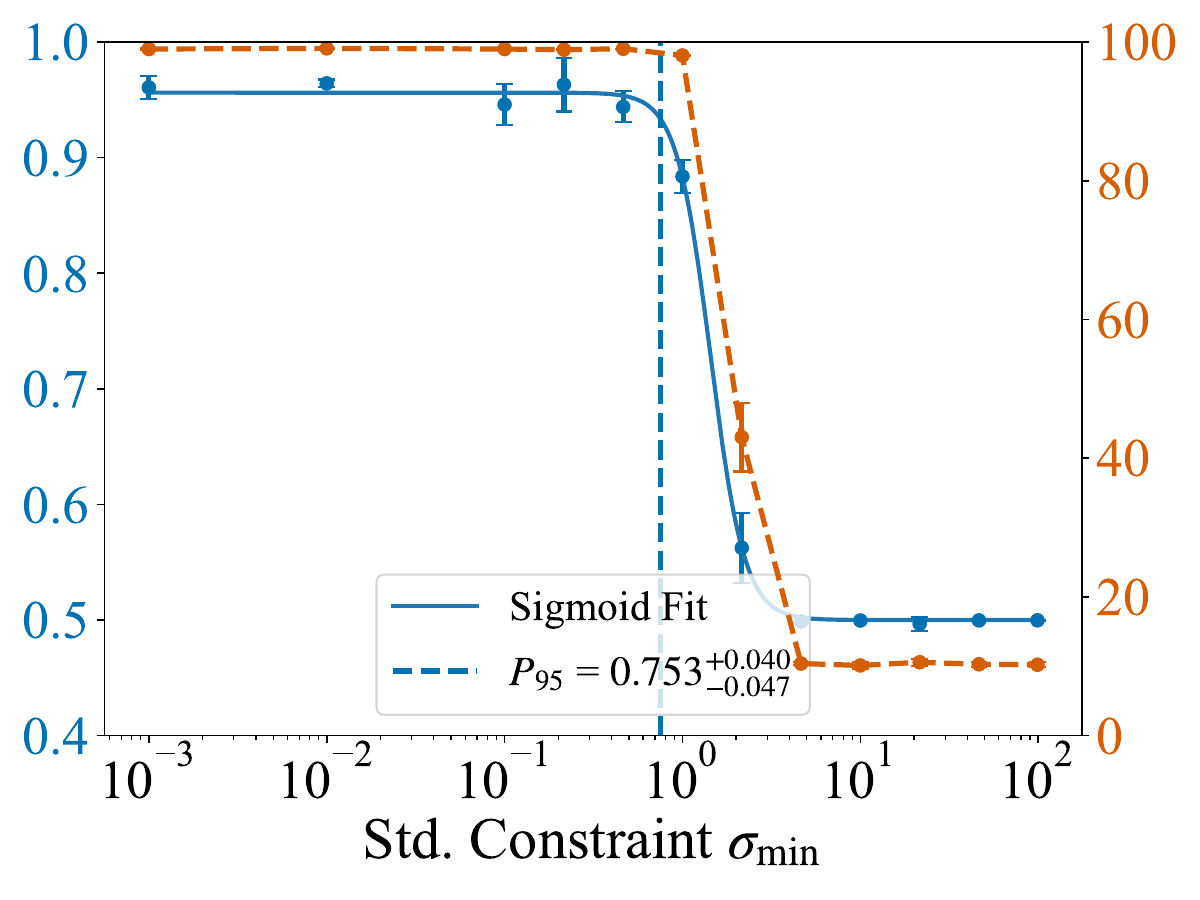}
        \includegraphics[width=\linewidth]
            {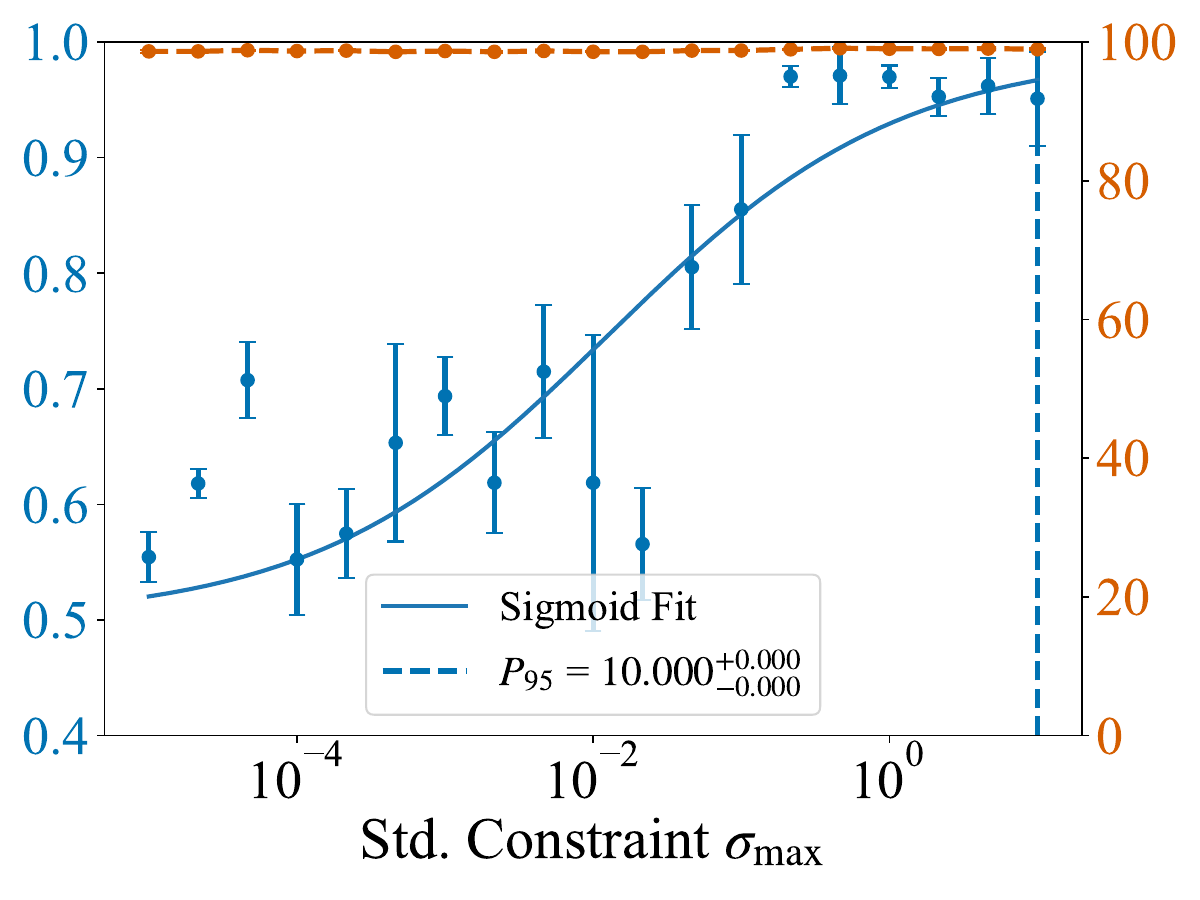}
    \end{subfigure}
    \hfill
    \begin{subfigure}[t]{0.235\textwidth}
        \centering
        \caption{A-Add}
        \includegraphics[width=\linewidth]
            {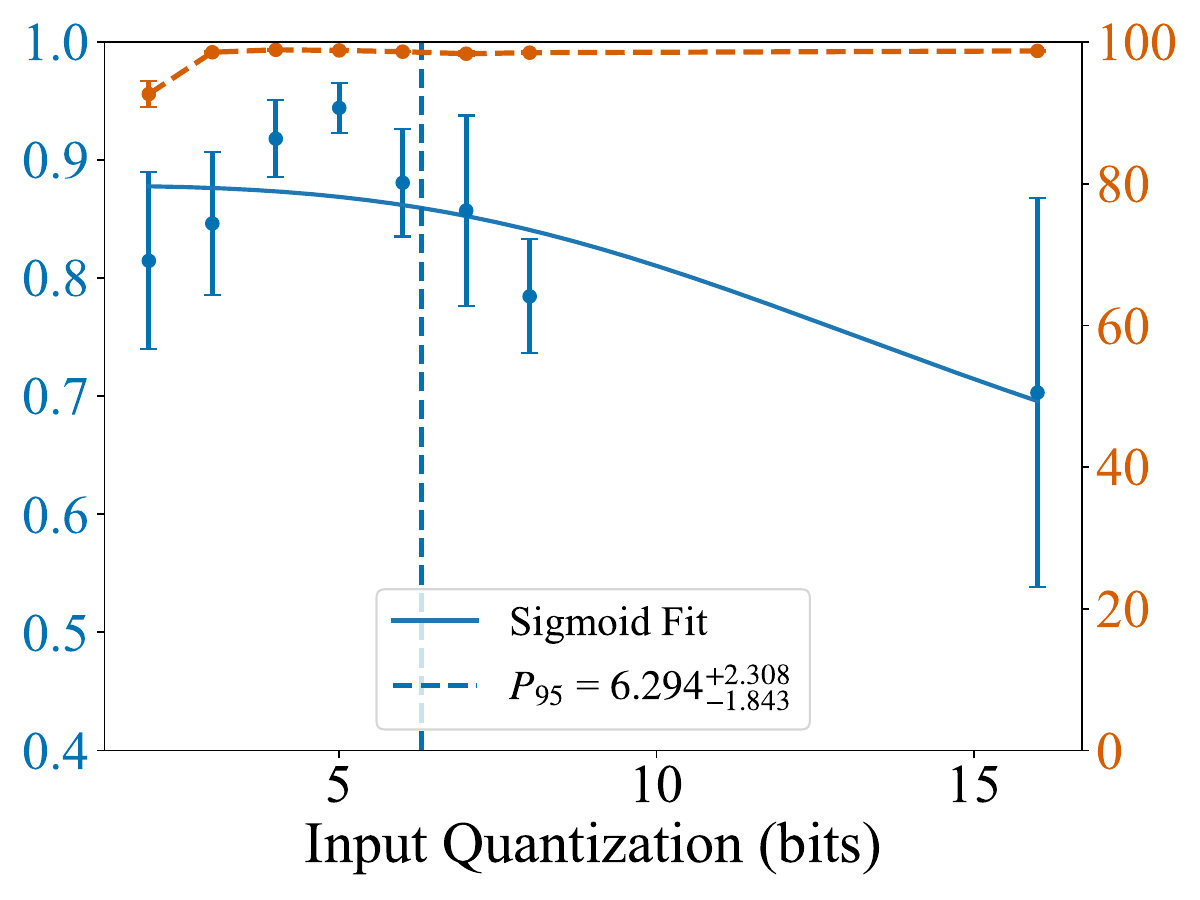}
        \includegraphics[width=\linewidth]
            {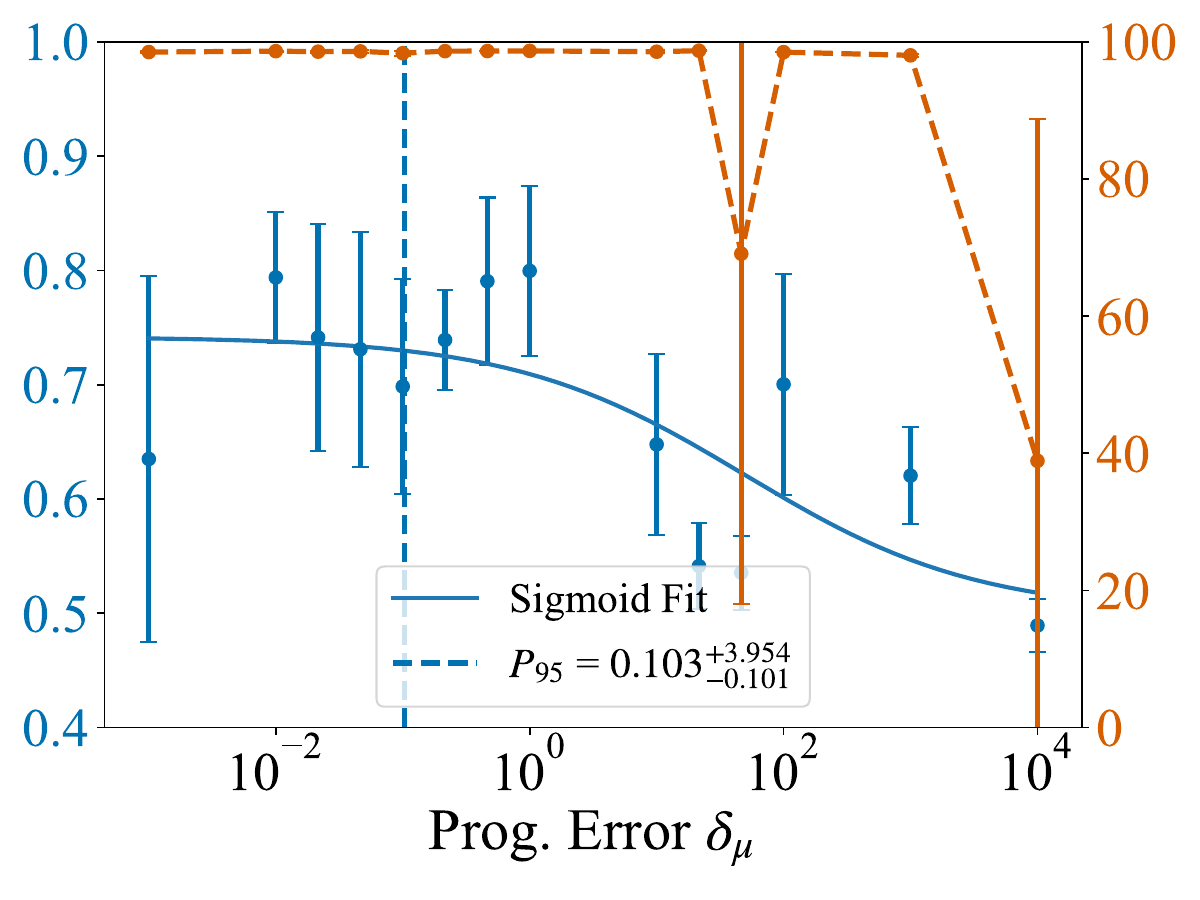}
        \includegraphics[width=\linewidth]
            {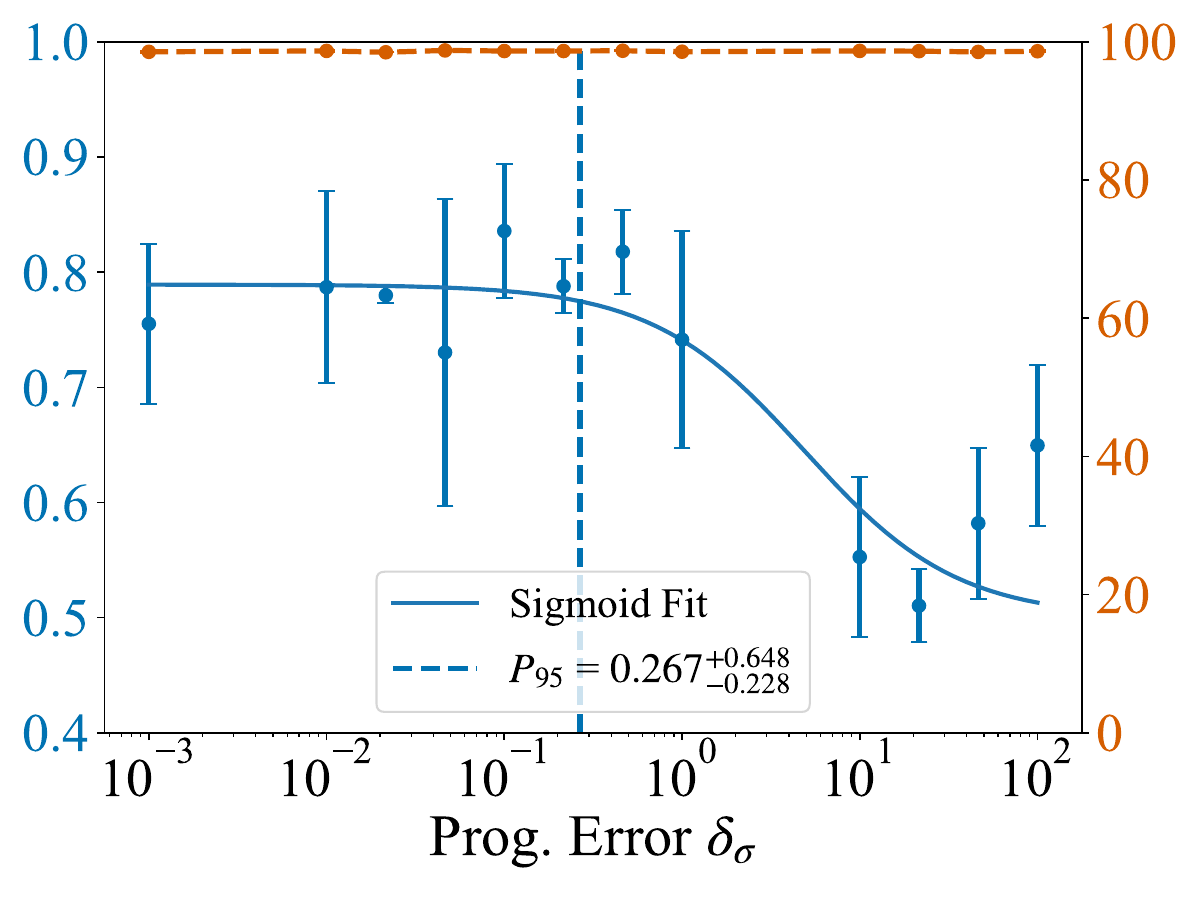}
        \includegraphics[width=\linewidth]
            {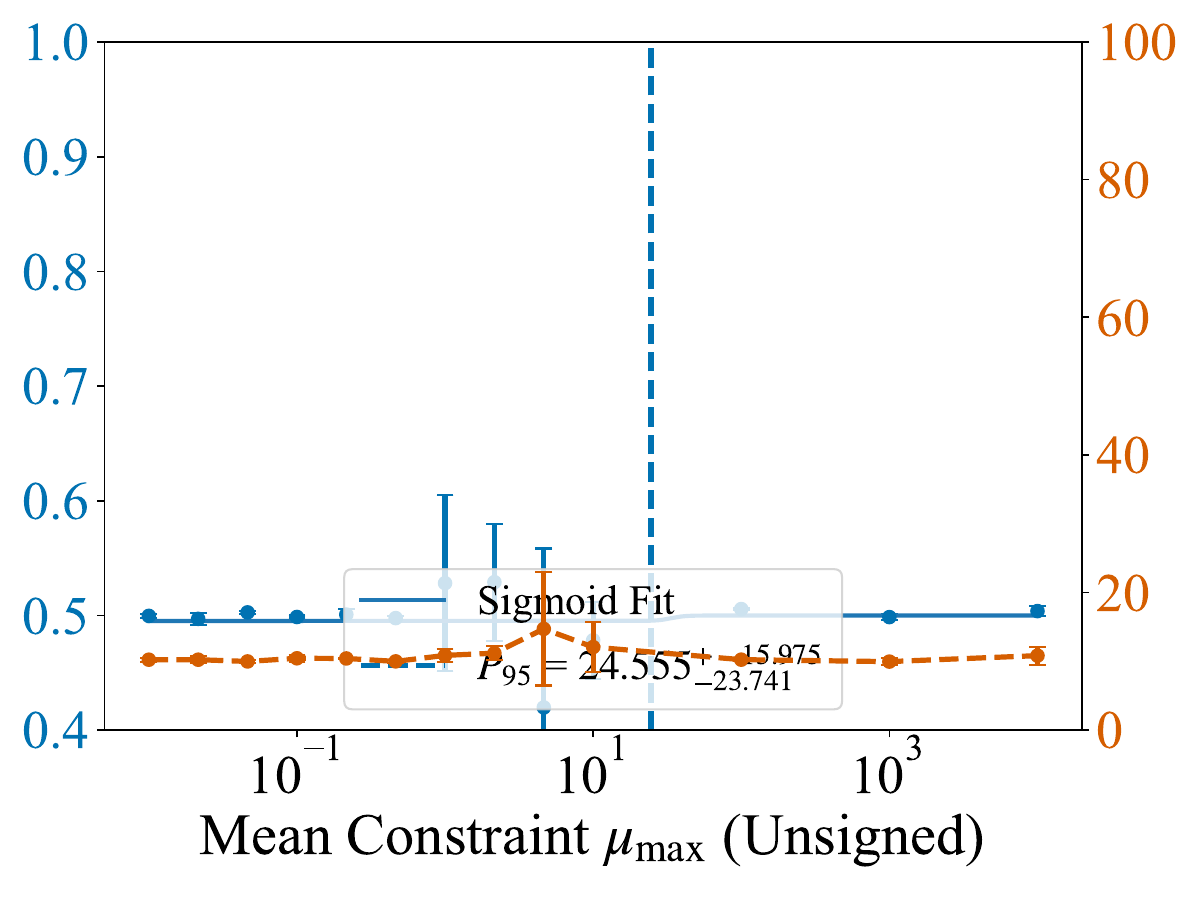}
        \includegraphics[width=\linewidth]
            {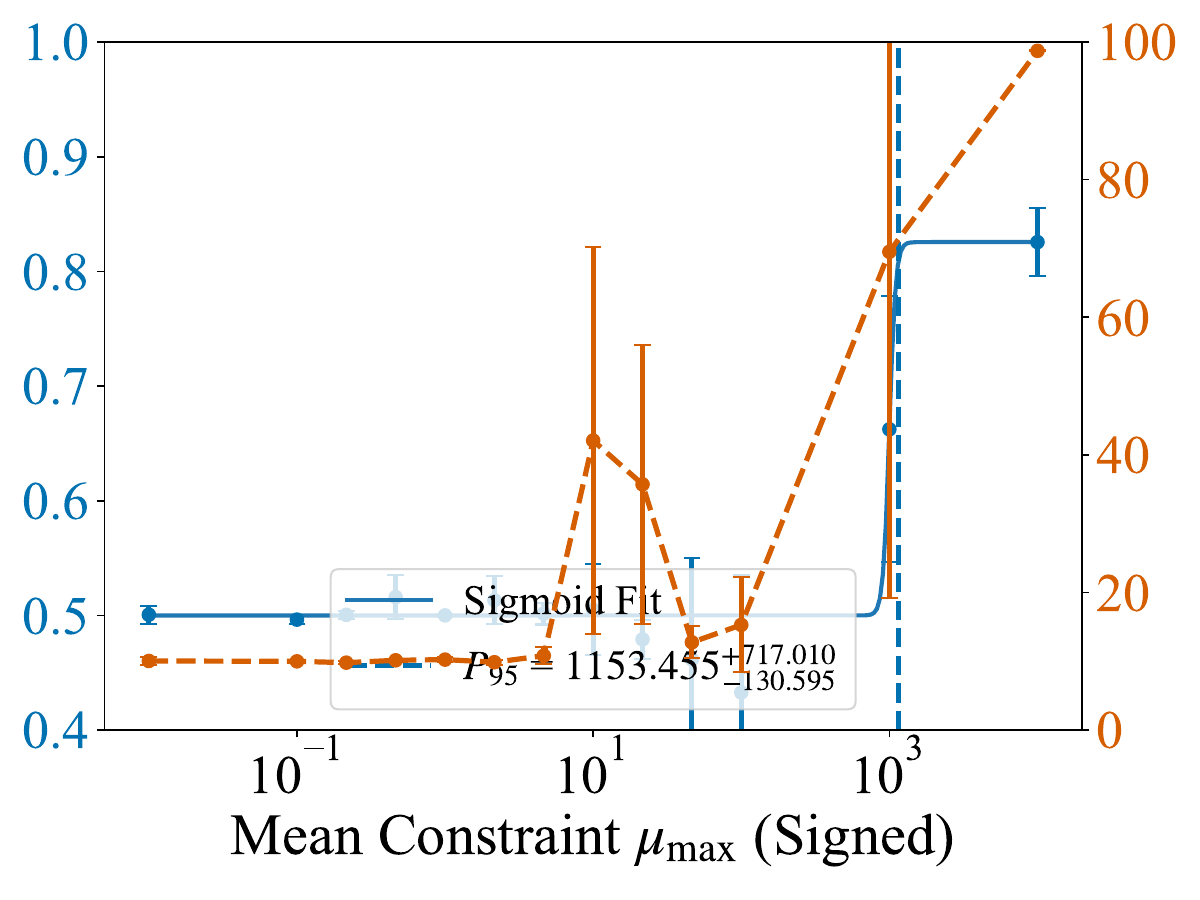}
        \includegraphics[width=\linewidth]
            {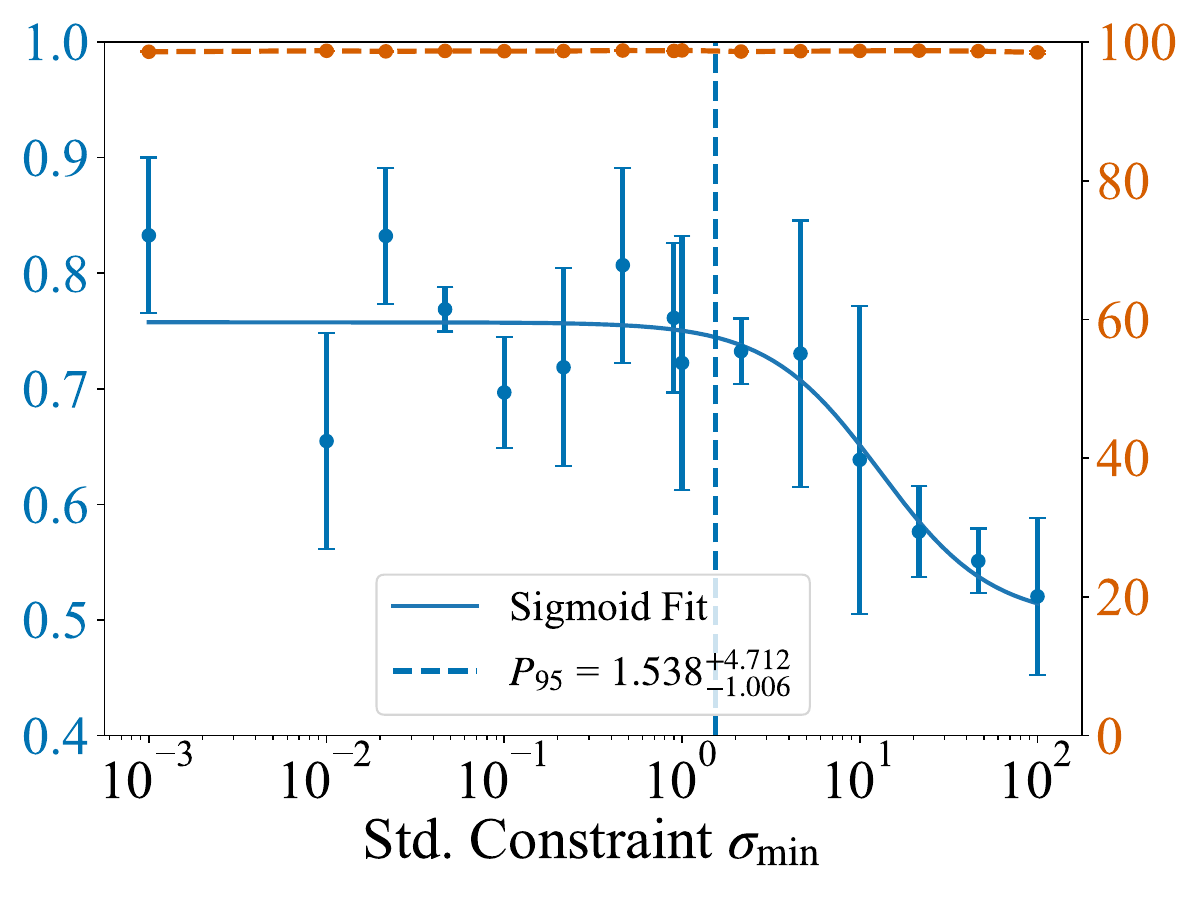}
        \includegraphics[width=\linewidth]
            {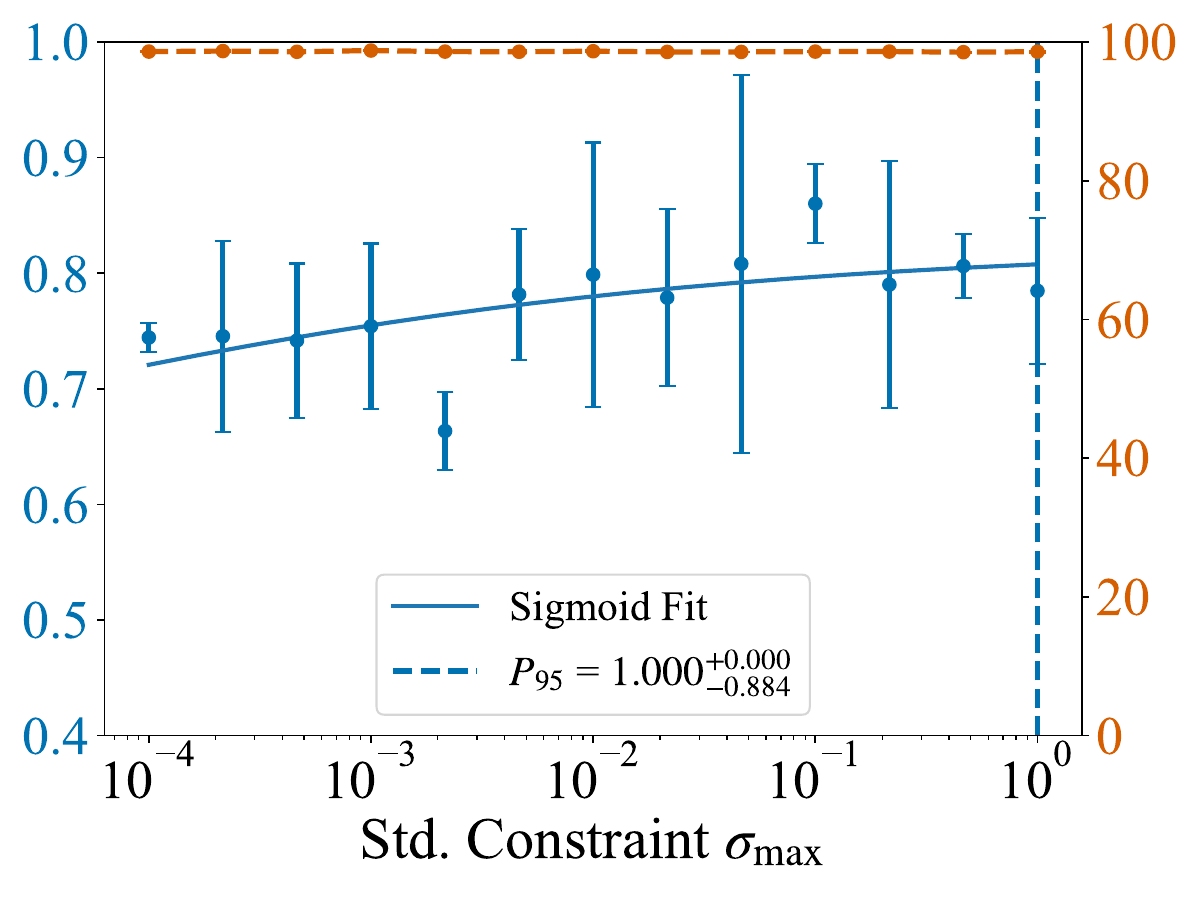}
    \end{subfigure}
    \hfill
    \begin{subfigure}[t]{0.235\textwidth}
        \centering
        \caption{A-Mul}
        \includegraphics[width=\linewidth]
            {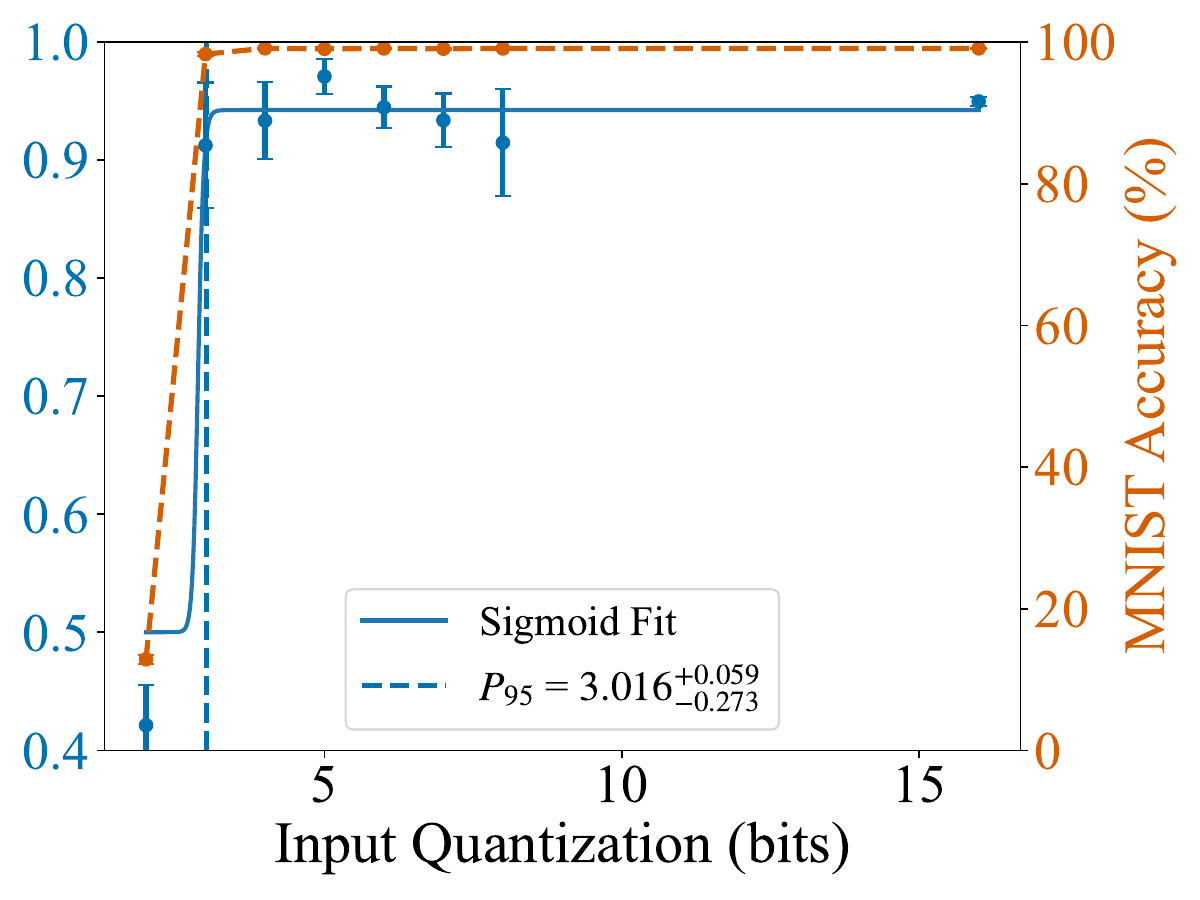}
        \includegraphics[width=\linewidth]
            {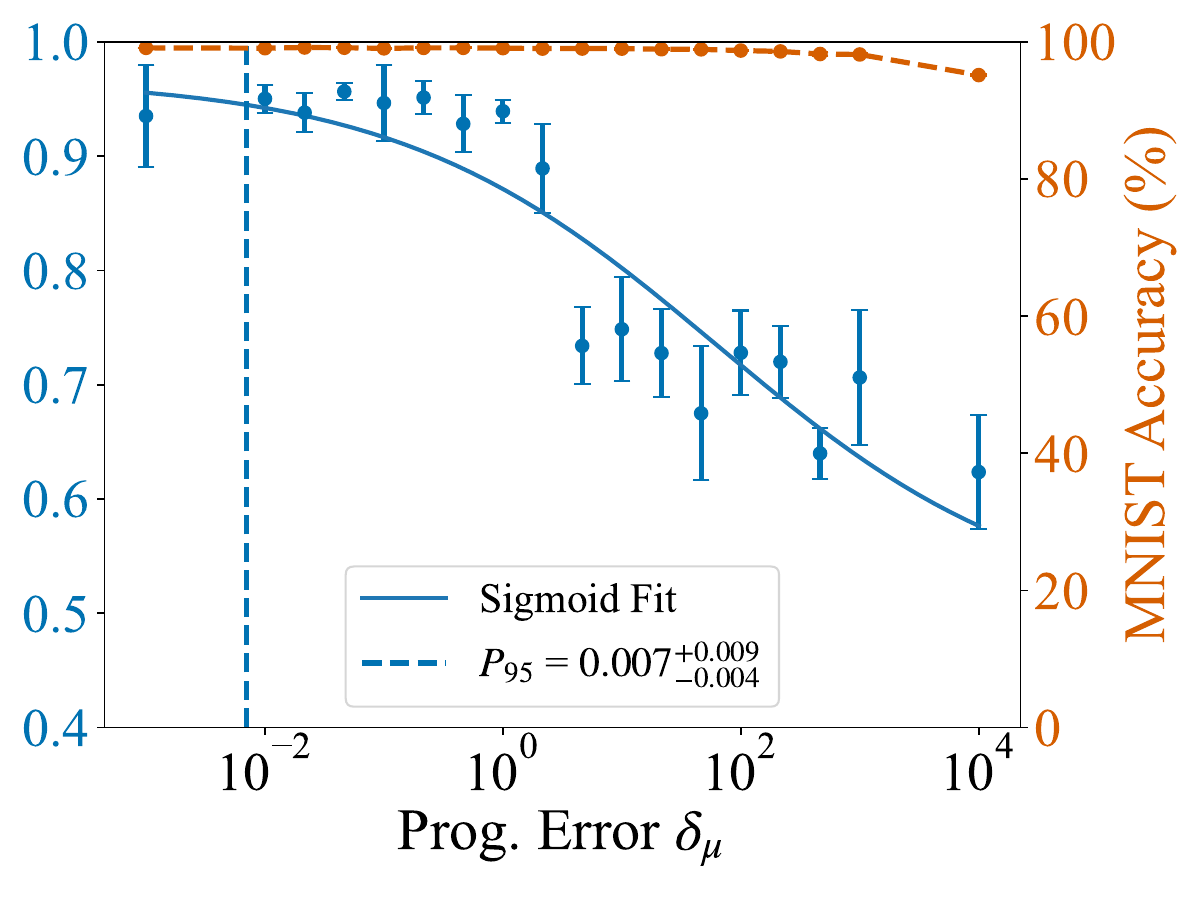}
        \includegraphics[width=\linewidth]
            {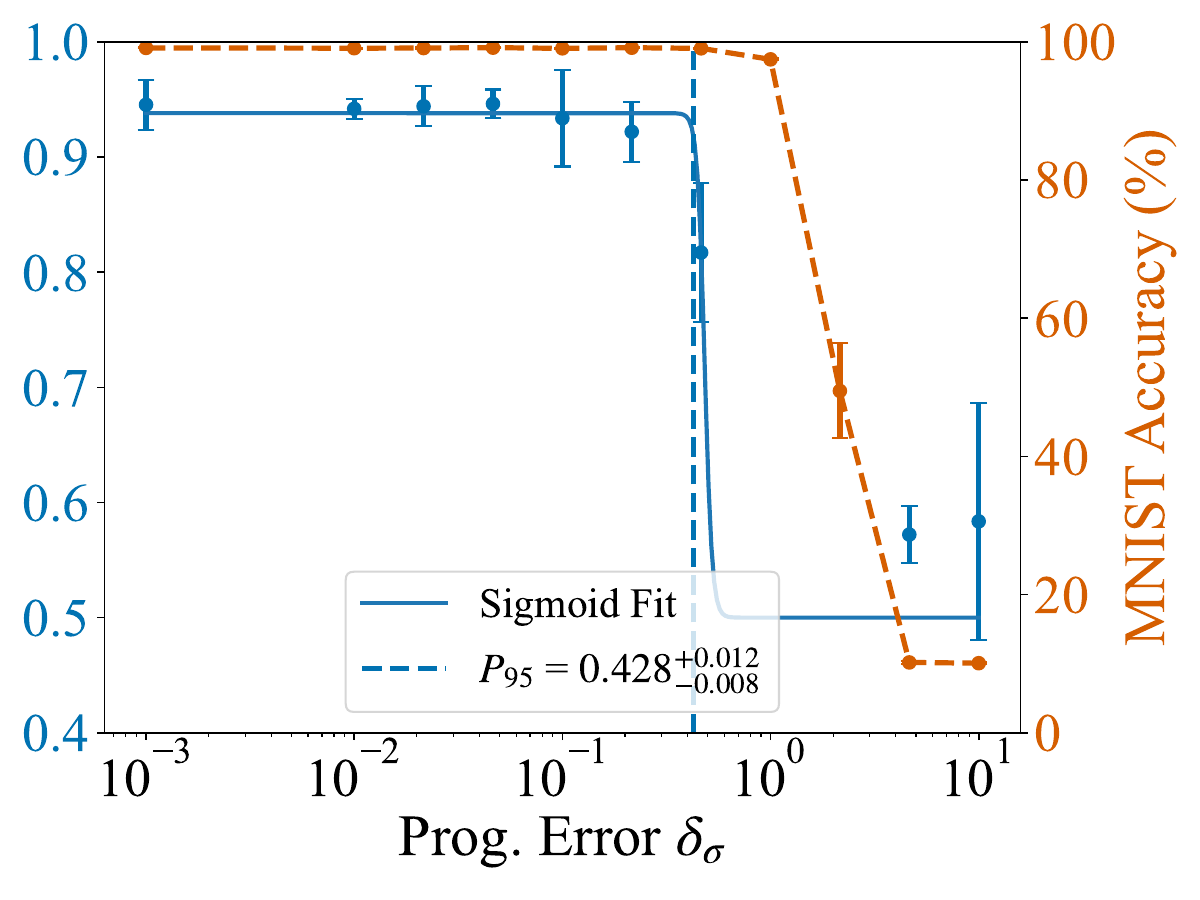}
        \includegraphics[width=\linewidth]
            {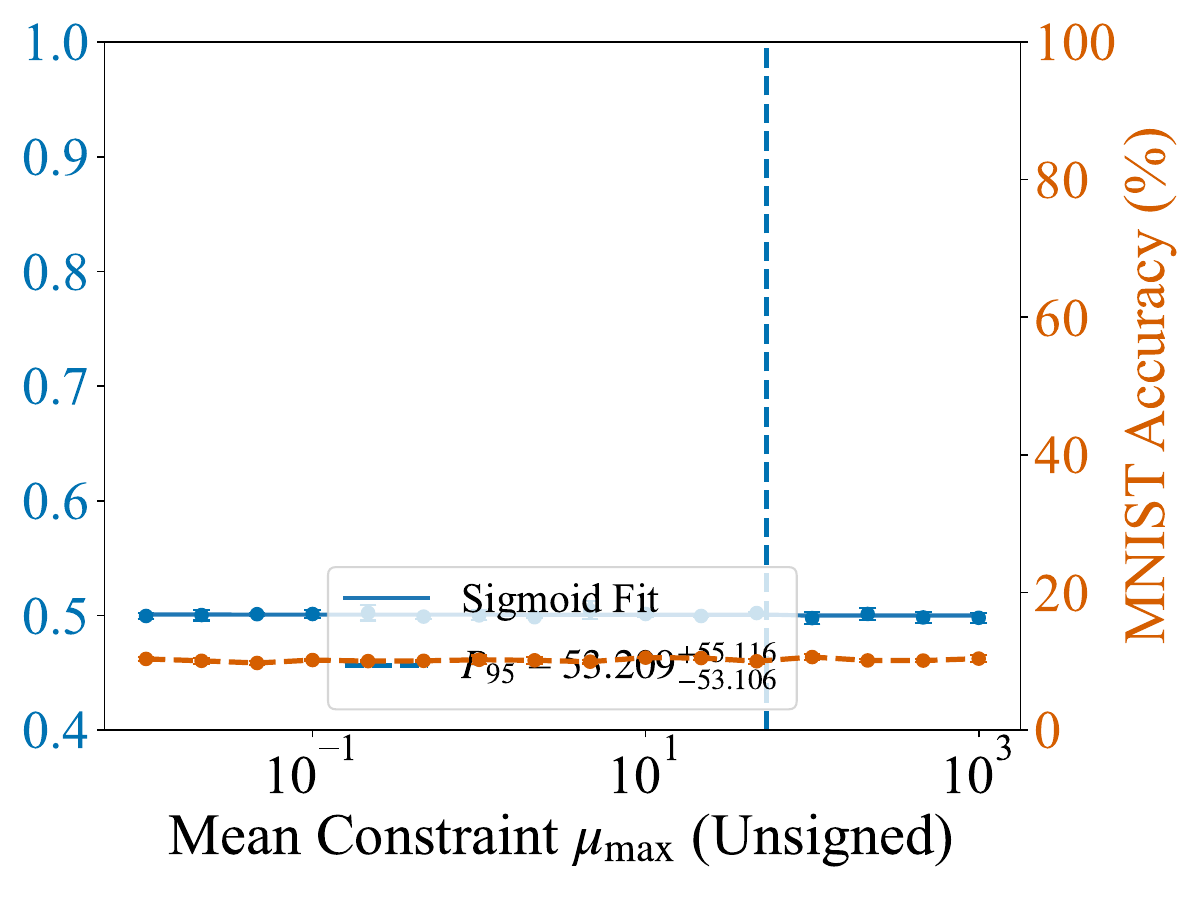}
        \includegraphics[width=\linewidth]
            {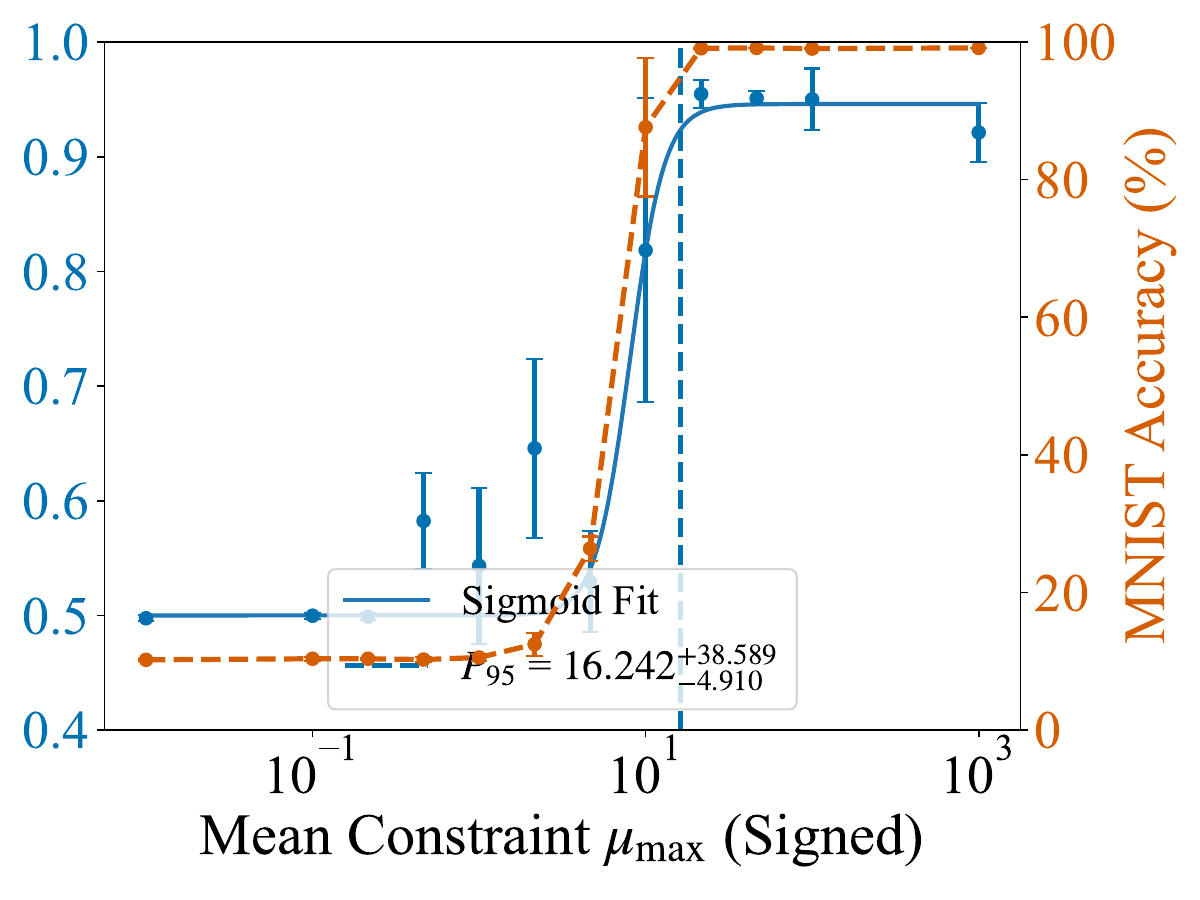}
        \includegraphics[width=\linewidth]
            {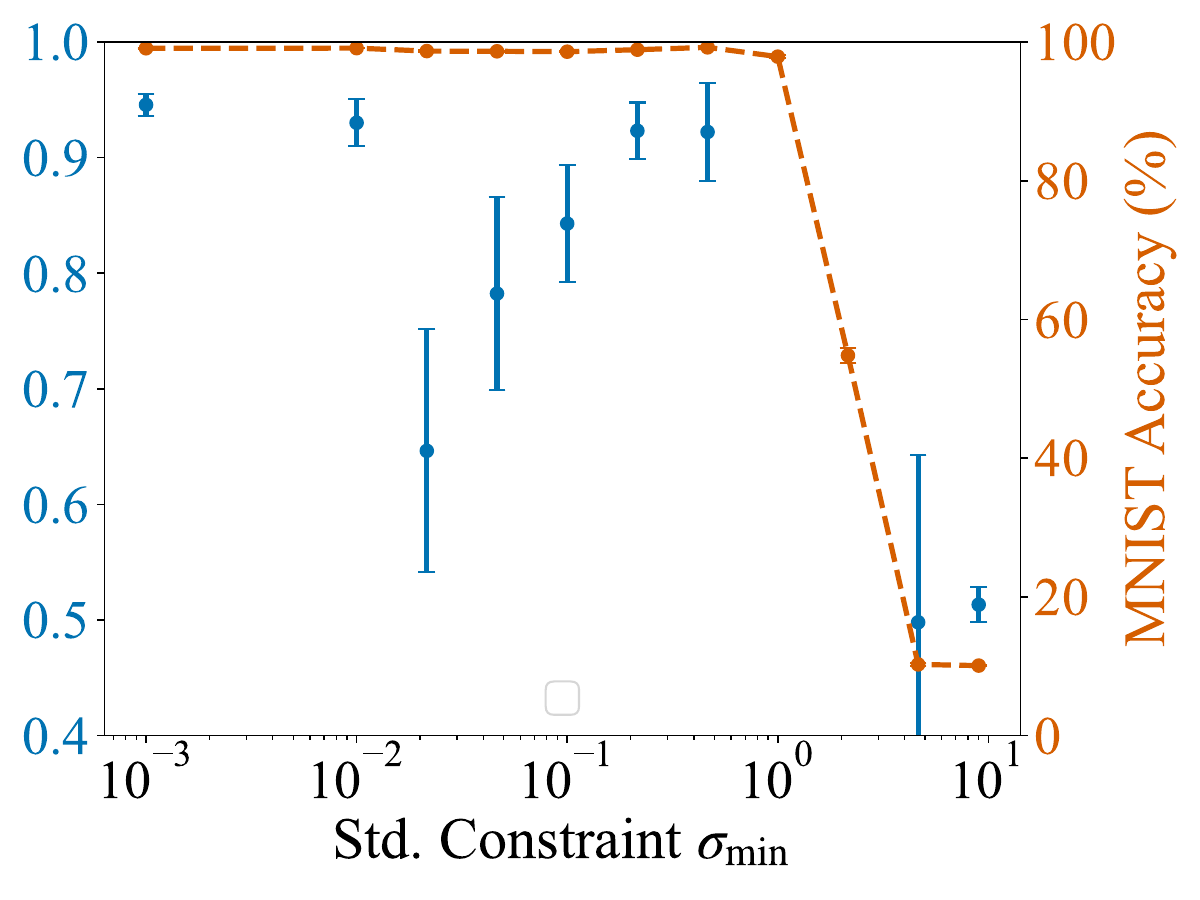}
        \includegraphics[width=\linewidth]
            {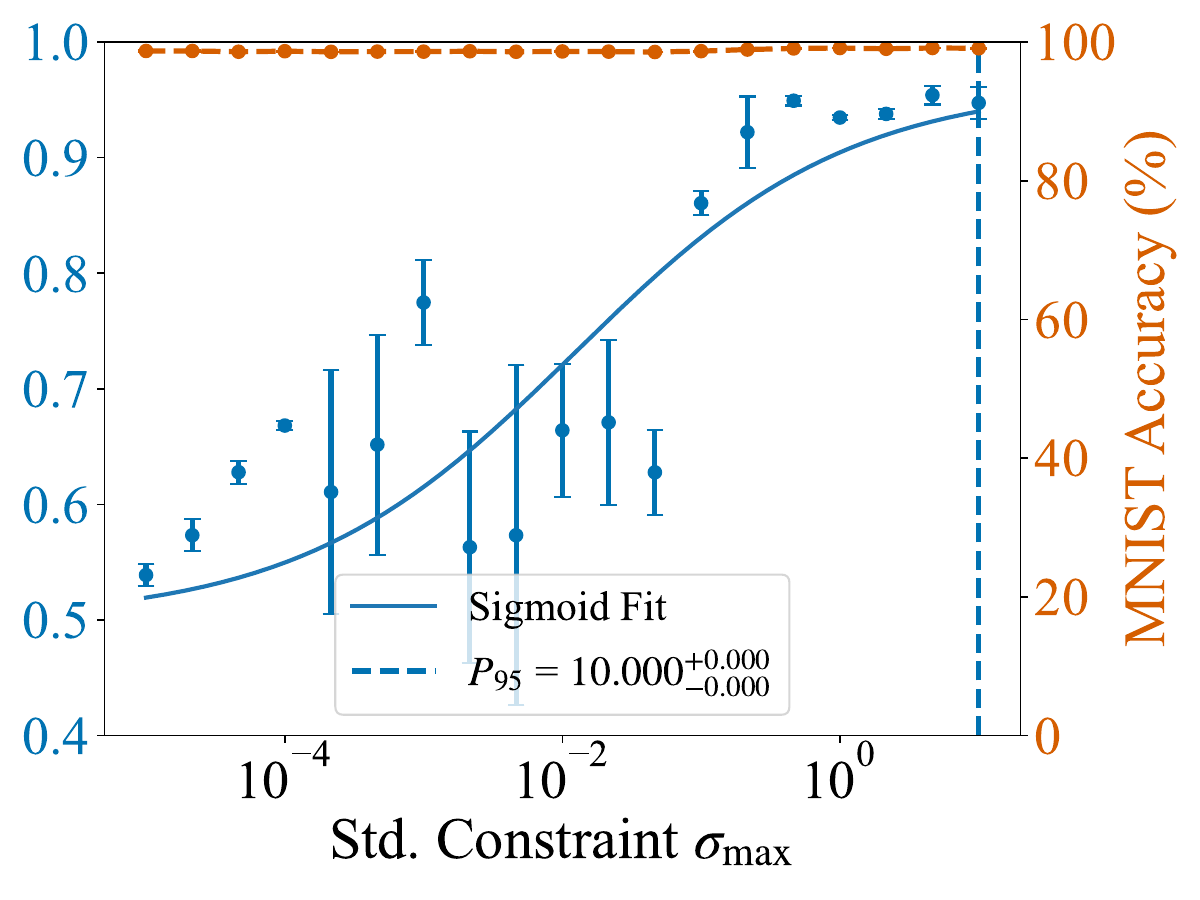}
    \end{subfigure}

    \caption{Ablation study of individual hardware constraints under
    hardware-aware training. Results are reported for LeNet on Dirty-MNIST.}
    \label{fig:AB-dmnist}
\end{figure*}

Figure~\ref{fig:AB-dmnist} presents the results for the simpler LeNet on Dirty-MNIST task with HW-aware training.  Compared to the weight-space stochasticity, activation-space stochasticity exhibits noticeably less stable uncertainty estimates. This observation further supports the arguments in Section~\ref{sec:ablation}  that weight-space formulations provide more reliable uncertainty quantification.

\subsection{Fully constrained study on CIFAR-10}
\begin{figure*}[tb!]
	\centering
	
	\begin{subfigure}[b]{0.49\textwidth}
		\centering
		\includegraphics[width=\linewidth]{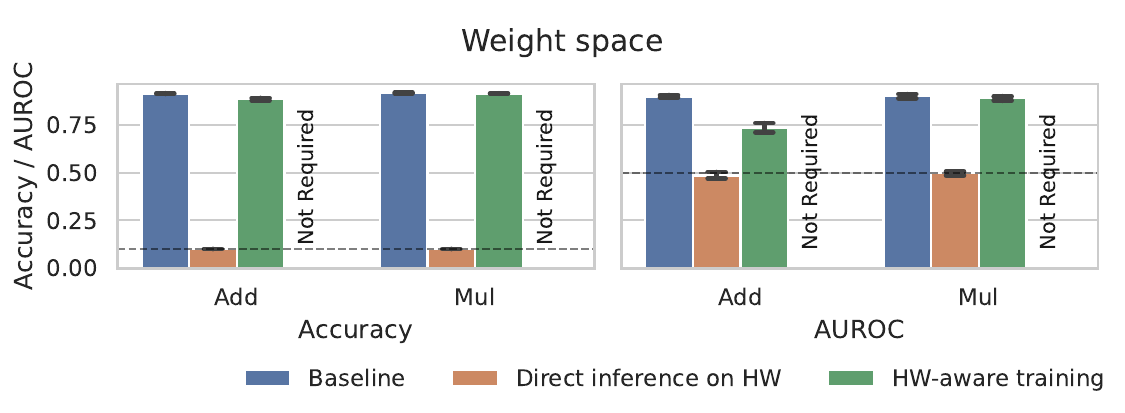}
		\label{fig:5-full-CIFAR_b}
	\end{subfigure}
	\hfill
	\begin{subfigure}[b]{0.49\textwidth}
		\centering
		\includegraphics[width=\linewidth]{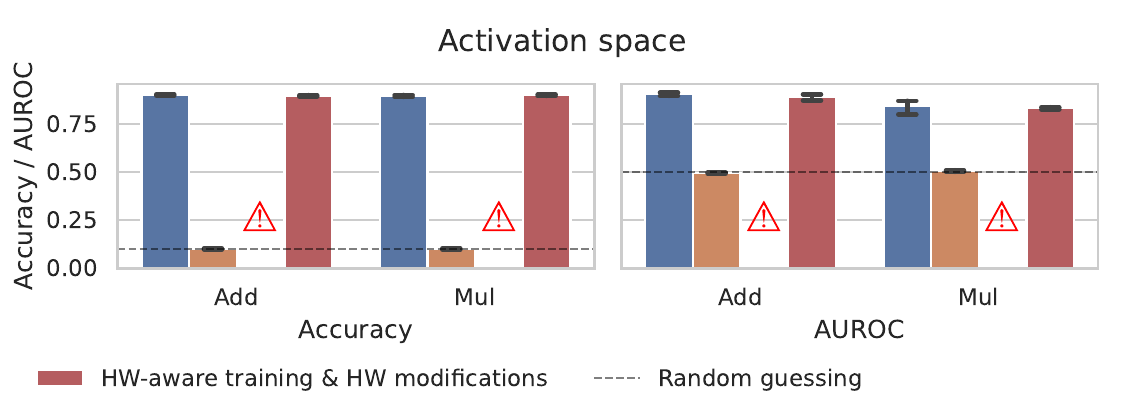}
		\label{fig:5-full-CIFAR_a}
	\end{subfigure}
	
	\caption{
		Co-design study for an example accelerator based on the \ATWI{} constraints,
		with equivalent settings to Figure \ref{fig:5-full-CINIC}, but on CIFAR-10 instead of CINIC-10.
		For weight-space stochasticity, the constraints lie within the stable operating range of Table~\ref{tab:ablation}, requiring only hardware-aware training.
		For activation-space stochasticity, hardware-aware training does not converge until the relevant hardware constraints are relaxed according to Table~\ref{tab:ablation}.
	}
	\label{fig:5-full-CIFAR}
\end{figure*}

In addition to Figure~\ref{fig:5-full-CINIC} from the main text, experiments under all constraints simultaneously were also conducted on CIFAR-10.
The results are shown in Figure~\ref{fig:5-full-CIFAR}.
In general the observations from these results are equivalent to those of Figure~\ref{fig:5-full-CINIC} and the main text.
However, these results additionally demonstrate that the constraints boundaries obtained in Table \ref{tab:ablation} hold validity across varying dataset complexity.

\newcommand{\AblationMissingPlotHeight}{0.75\linewidth}
\newcommand{\AblationMissingPlot}{%
    \par\noindent\begin{minipage}[c][\AblationMissingPlotHeight][c]{0.90\linewidth}
        \centering No experiments with unsigned constraints converged; training produced NaNs, consistent with exploding gradients.
    \end{minipage}\par%
}

\begin{figure*}[p]
    \centering

    \begin{subfigure}[t]{0.235\textwidth}
        \centering
        \caption{W-Add}
        \includegraphics[width=\linewidth]
            {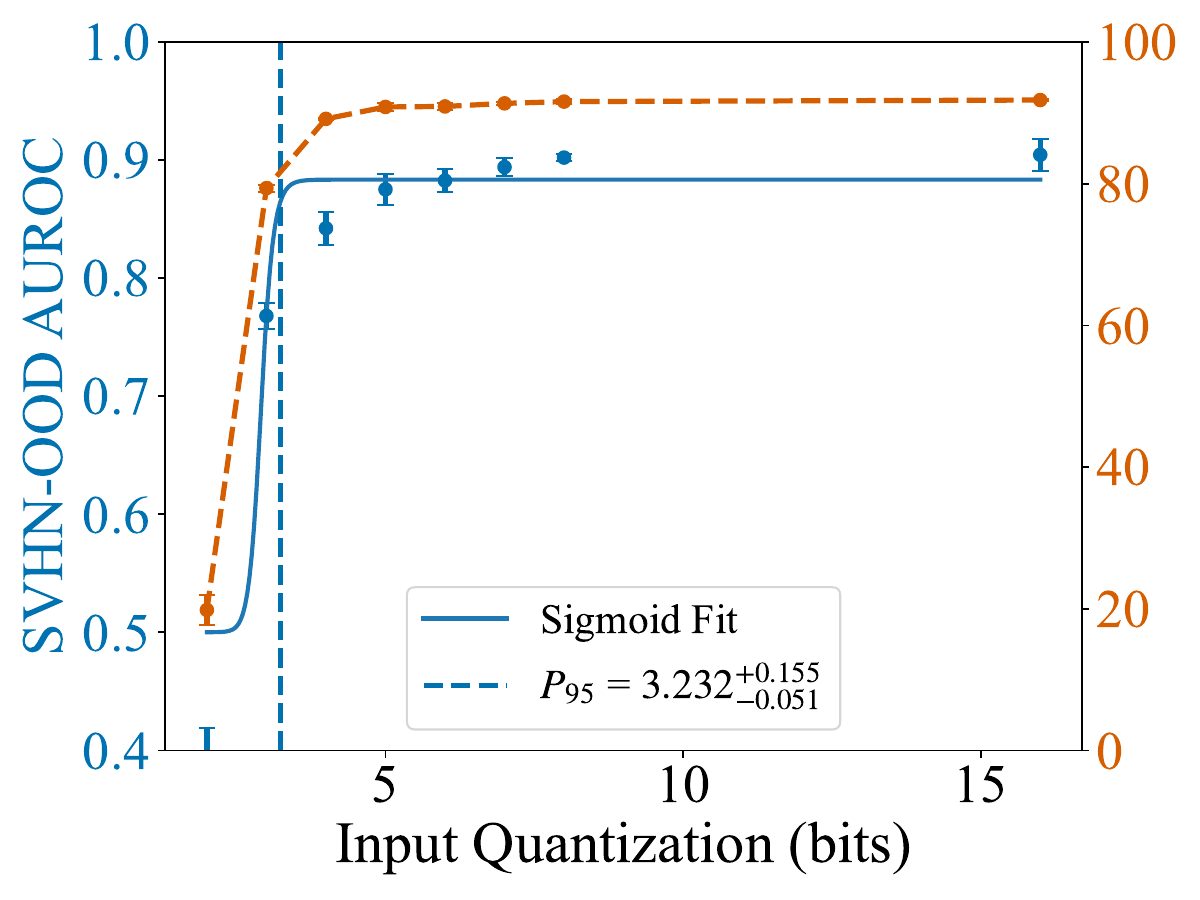}
        \includegraphics[width=\linewidth]
            {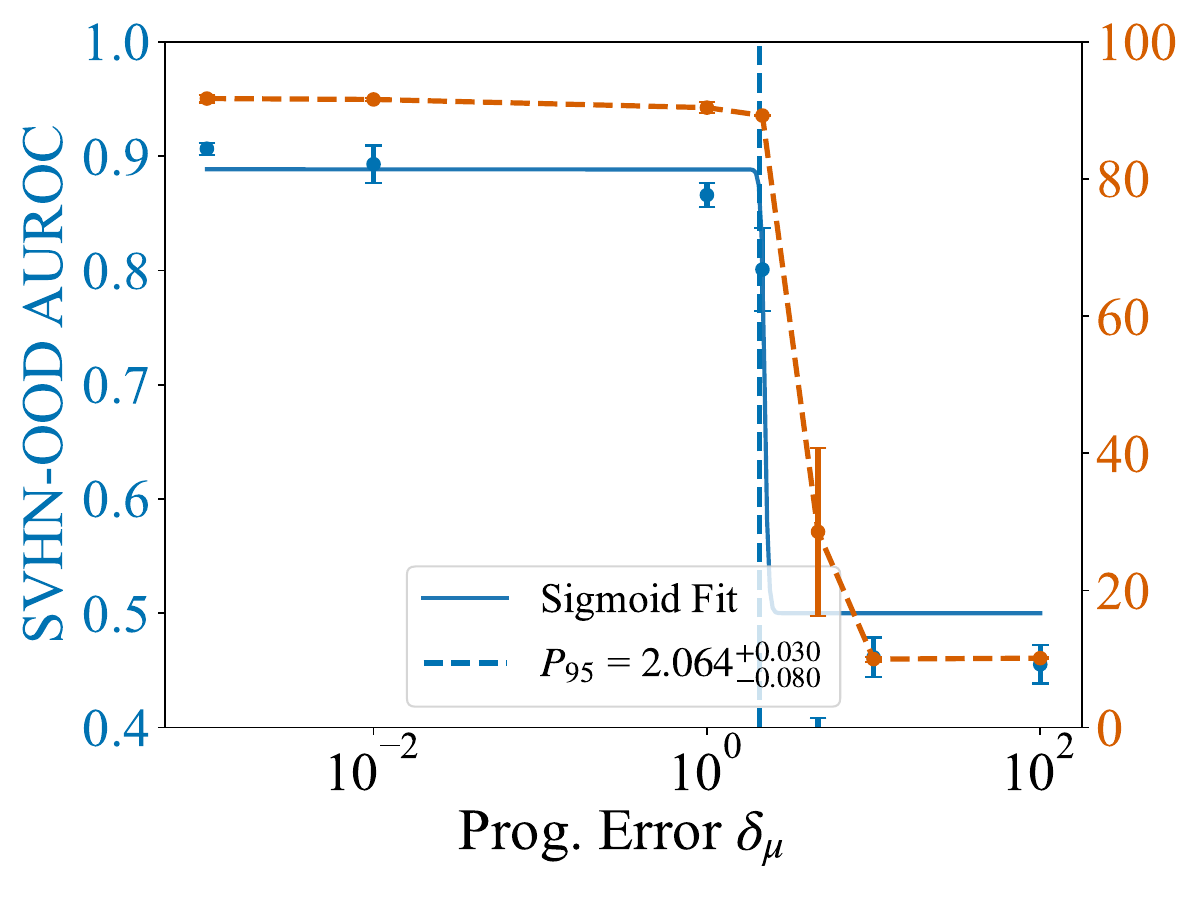}
        \includegraphics[width=\linewidth]
            {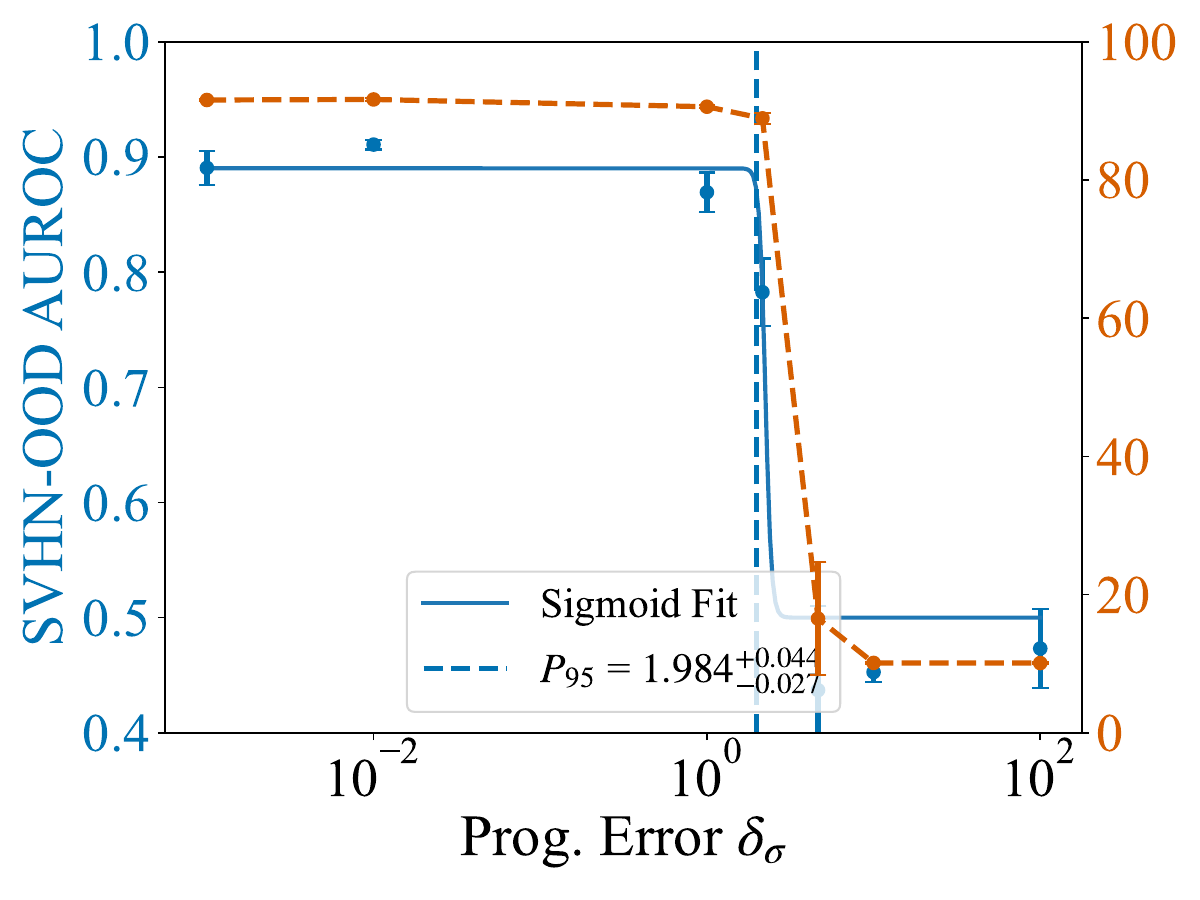}
        \includegraphics[width=\linewidth]
            {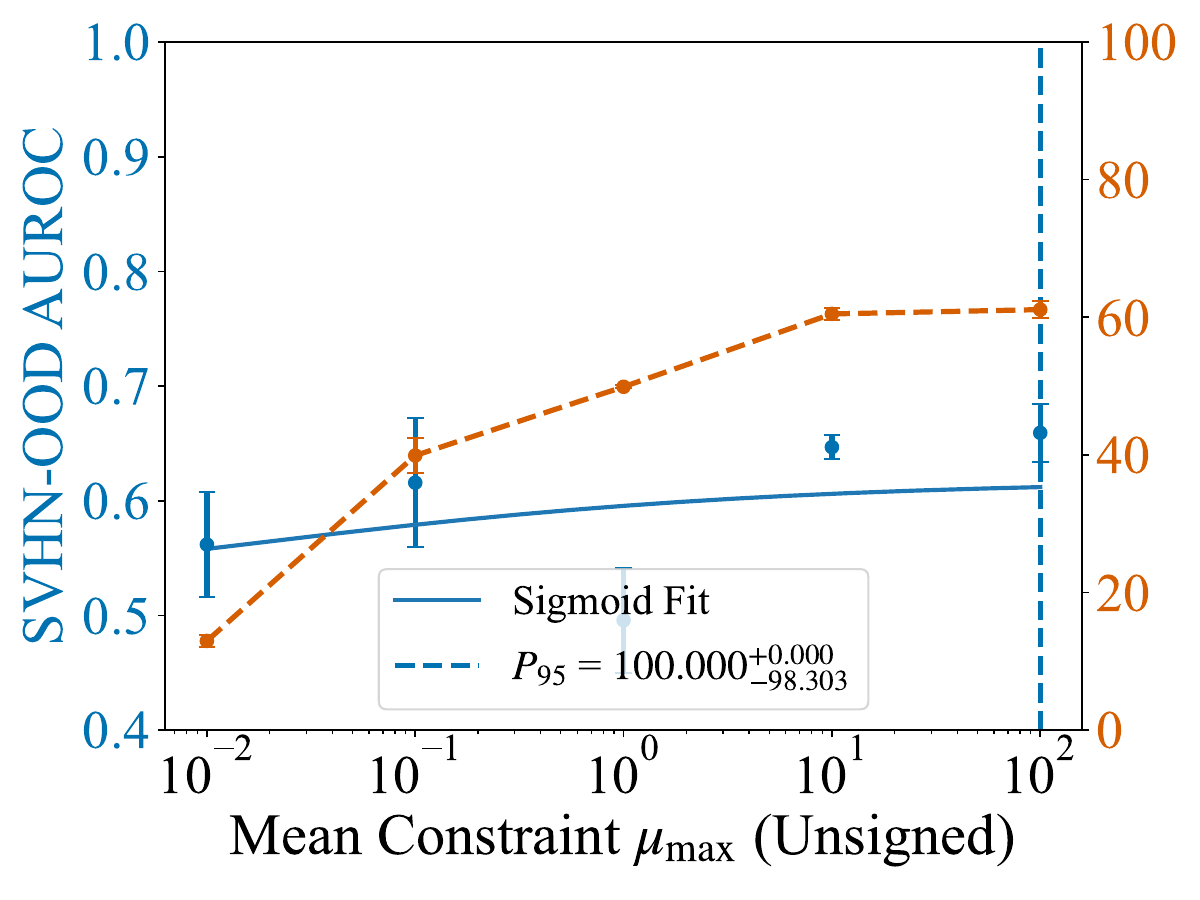}
        \includegraphics[width=\linewidth]
            {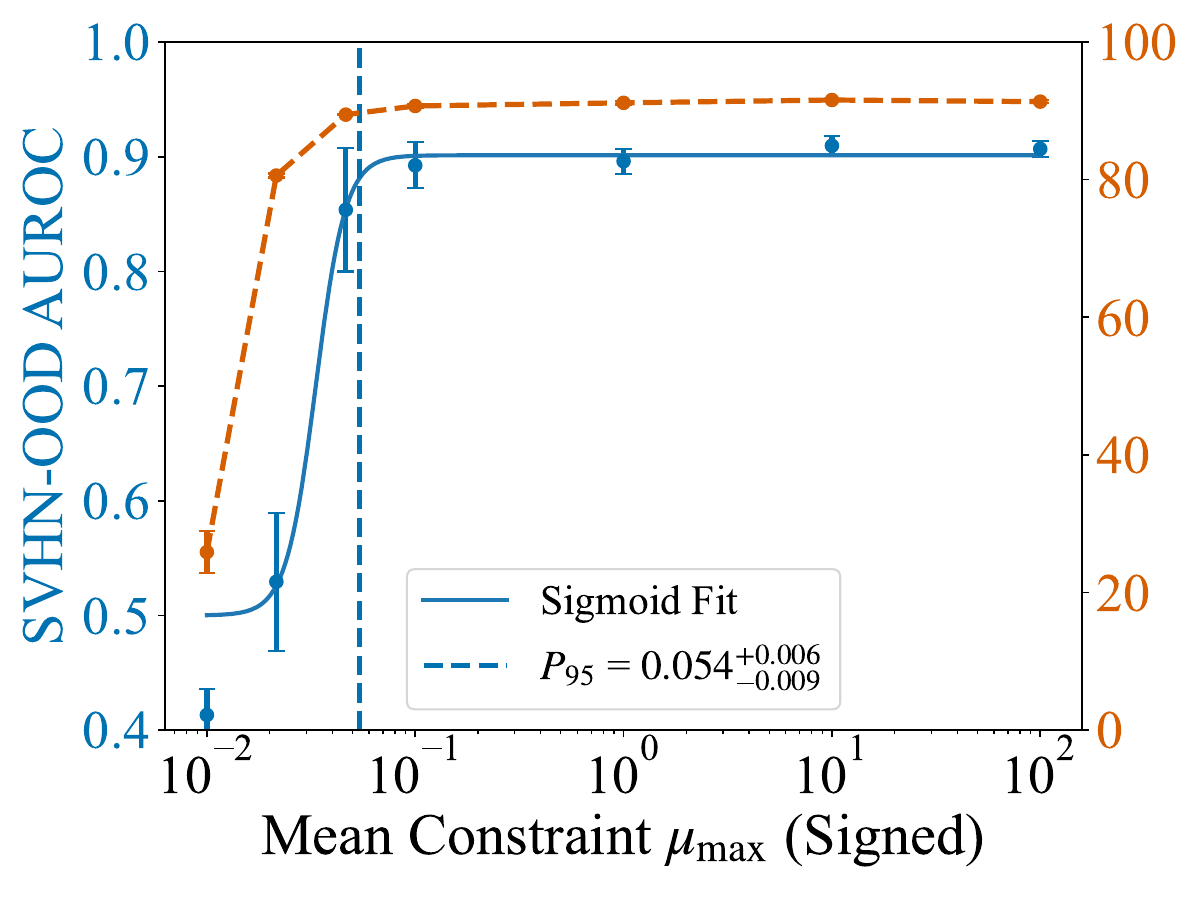}
        \includegraphics[width=\linewidth]
            {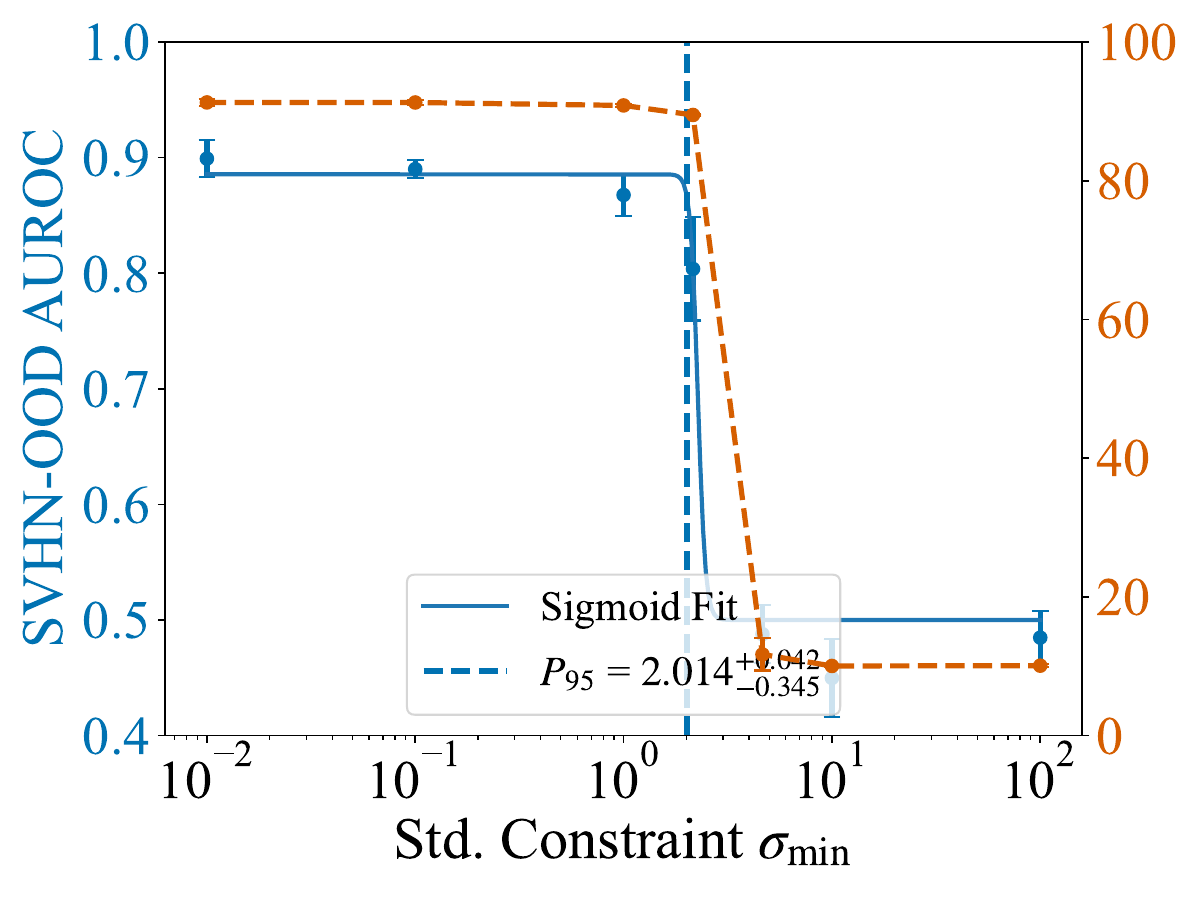}
        \includegraphics[width=\linewidth]
            {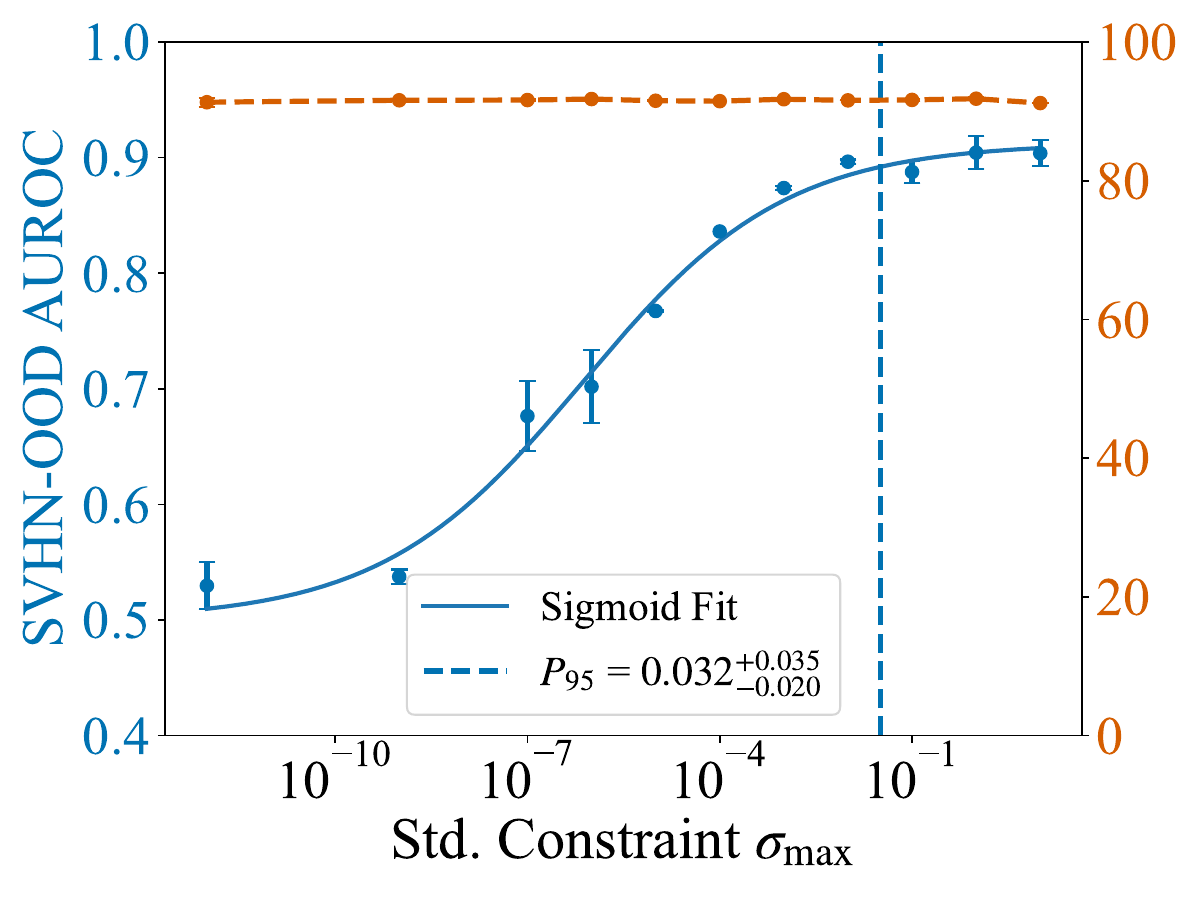}
    \end{subfigure}
    \hfill
    \begin{subfigure}[t]{0.235\textwidth}
        \centering
        \caption{W-Mul}
        \includegraphics[width=\linewidth]
            {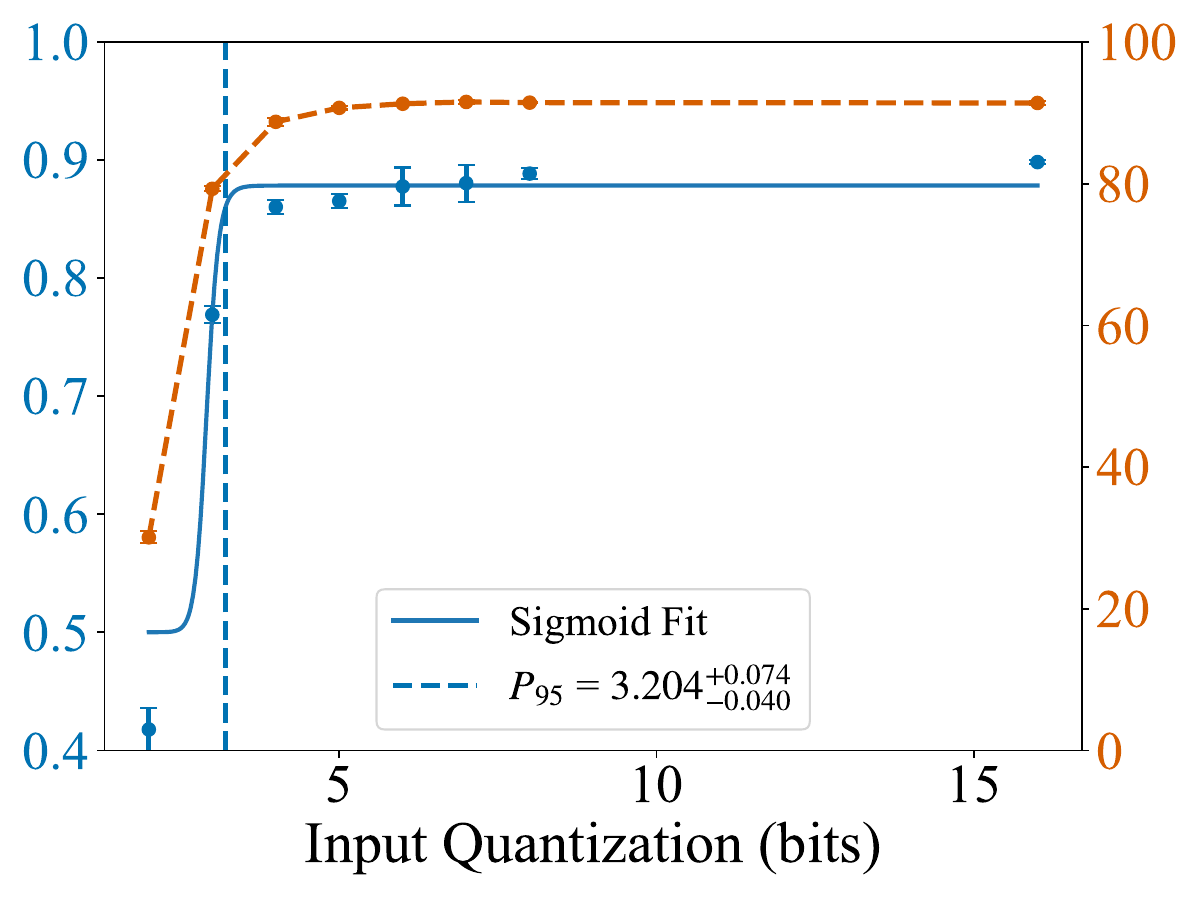}
        \includegraphics[width=\linewidth]
            {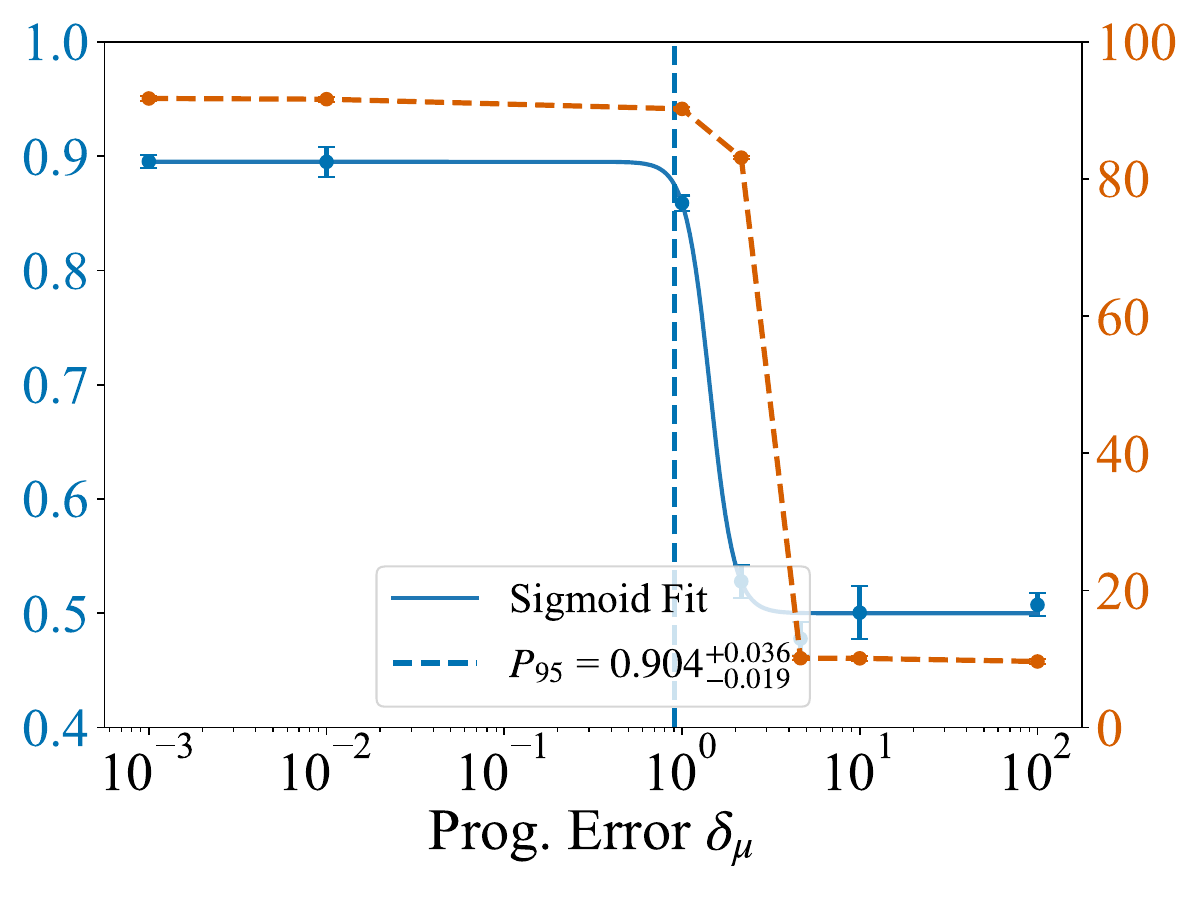}
        \includegraphics[width=\linewidth]
            {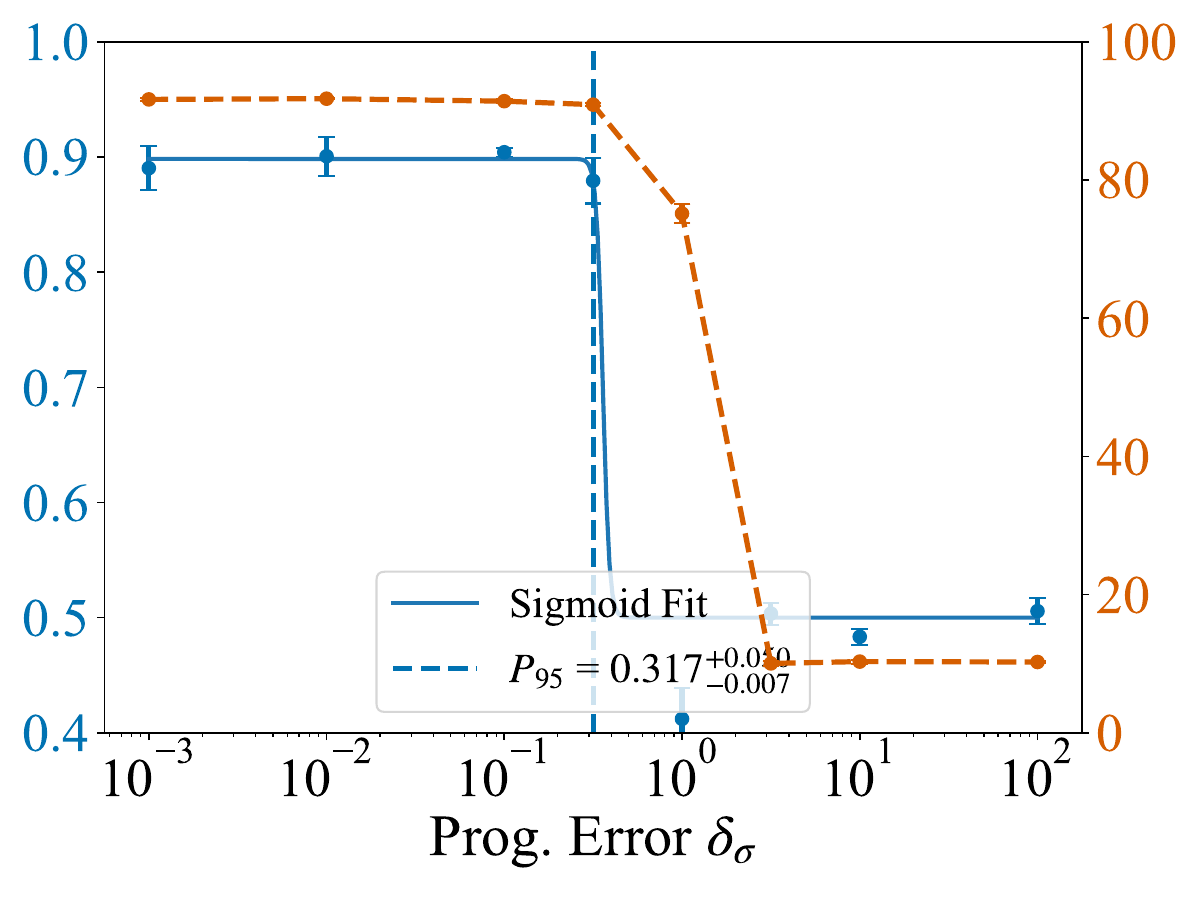}
        \includegraphics[width=\linewidth]
            {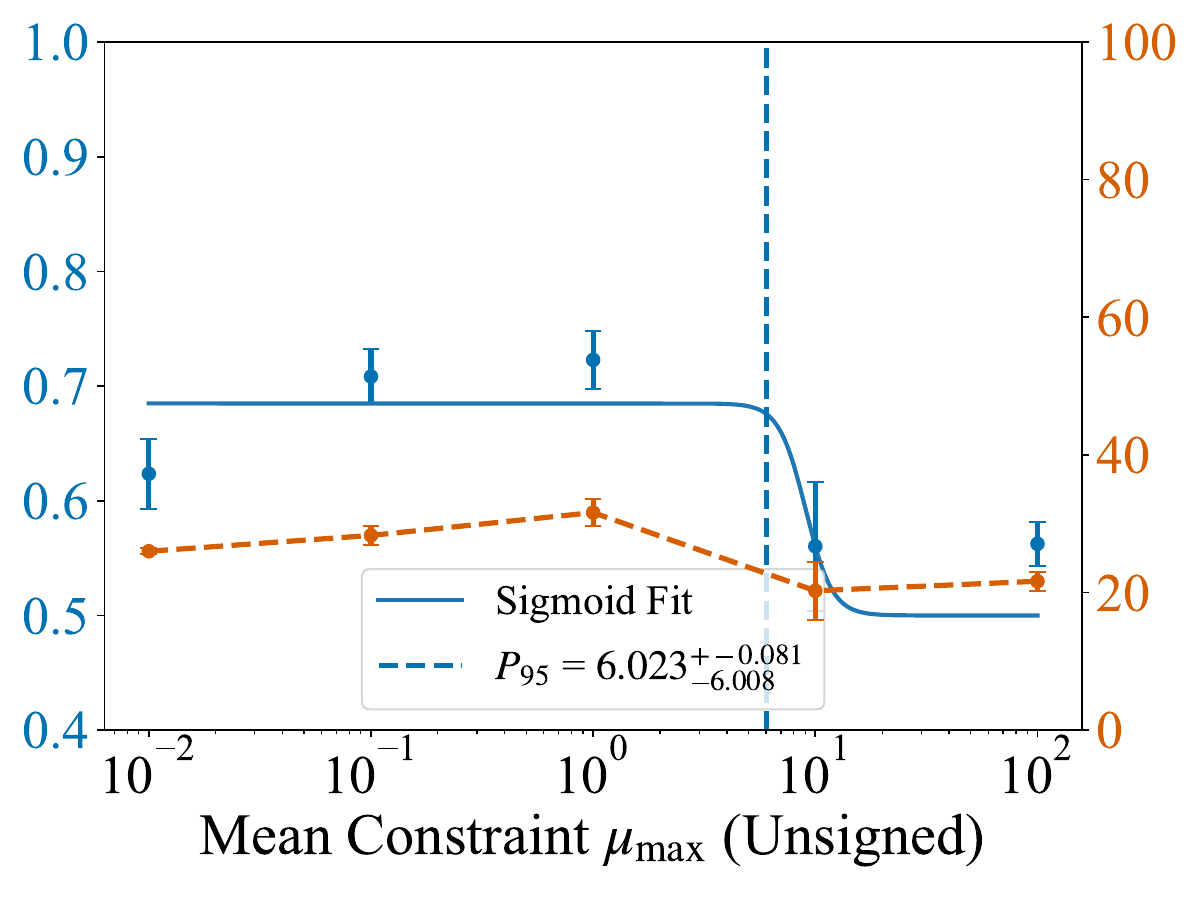}
        \includegraphics[width=\linewidth]
            {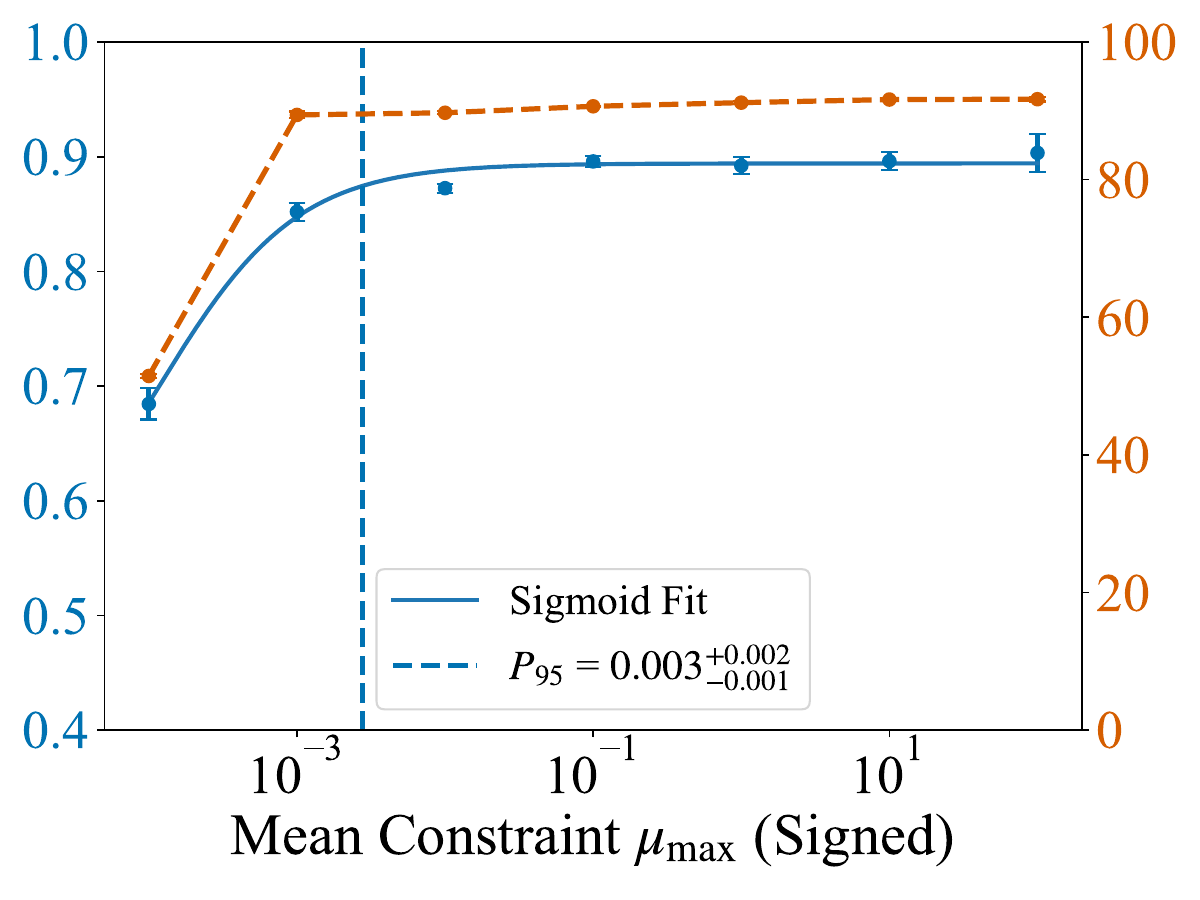}
        \includegraphics[width=\linewidth]
            {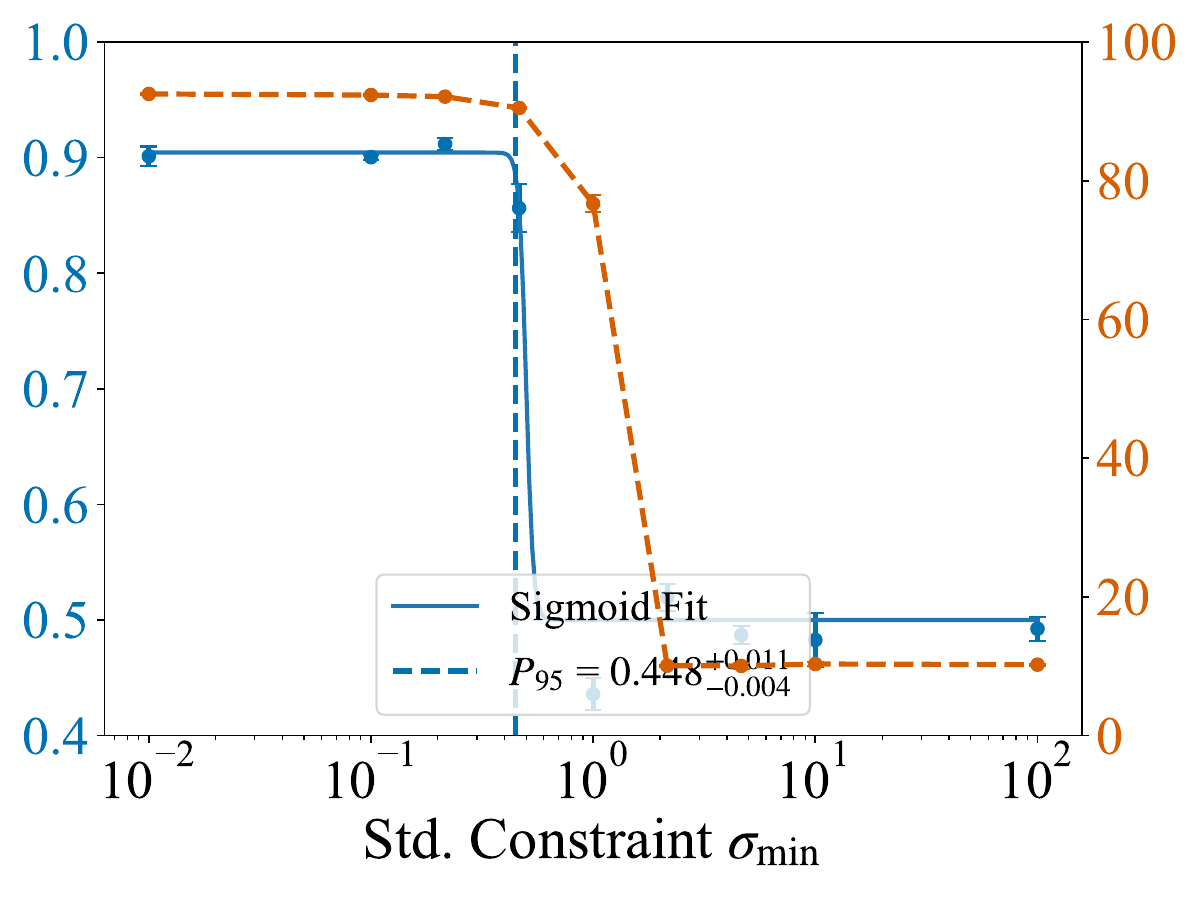}
        \includegraphics[width=\linewidth]
            {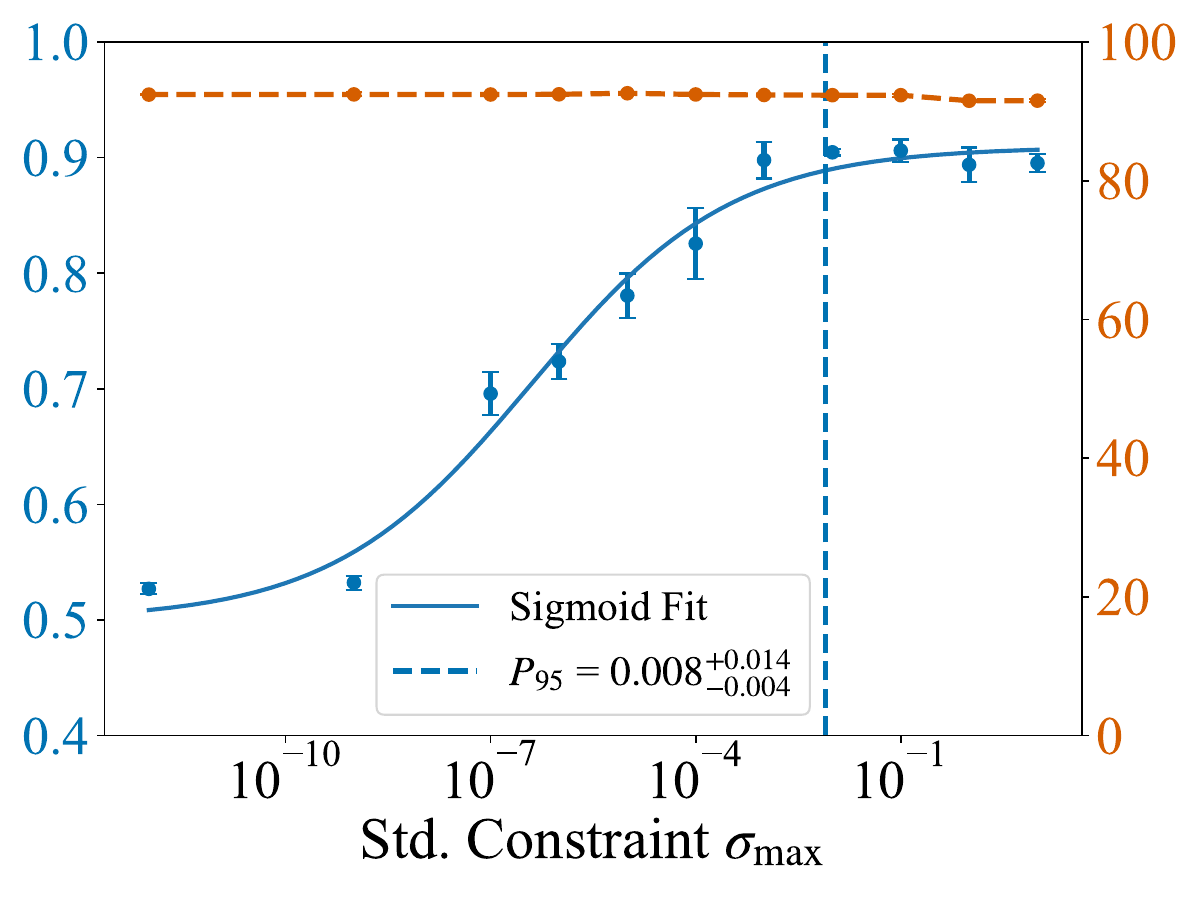}
    \end{subfigure}
    \hfill
    \begin{subfigure}[t]{0.235\textwidth}
        \centering
        \caption{A-Add}
        \includegraphics[width=\linewidth]
            {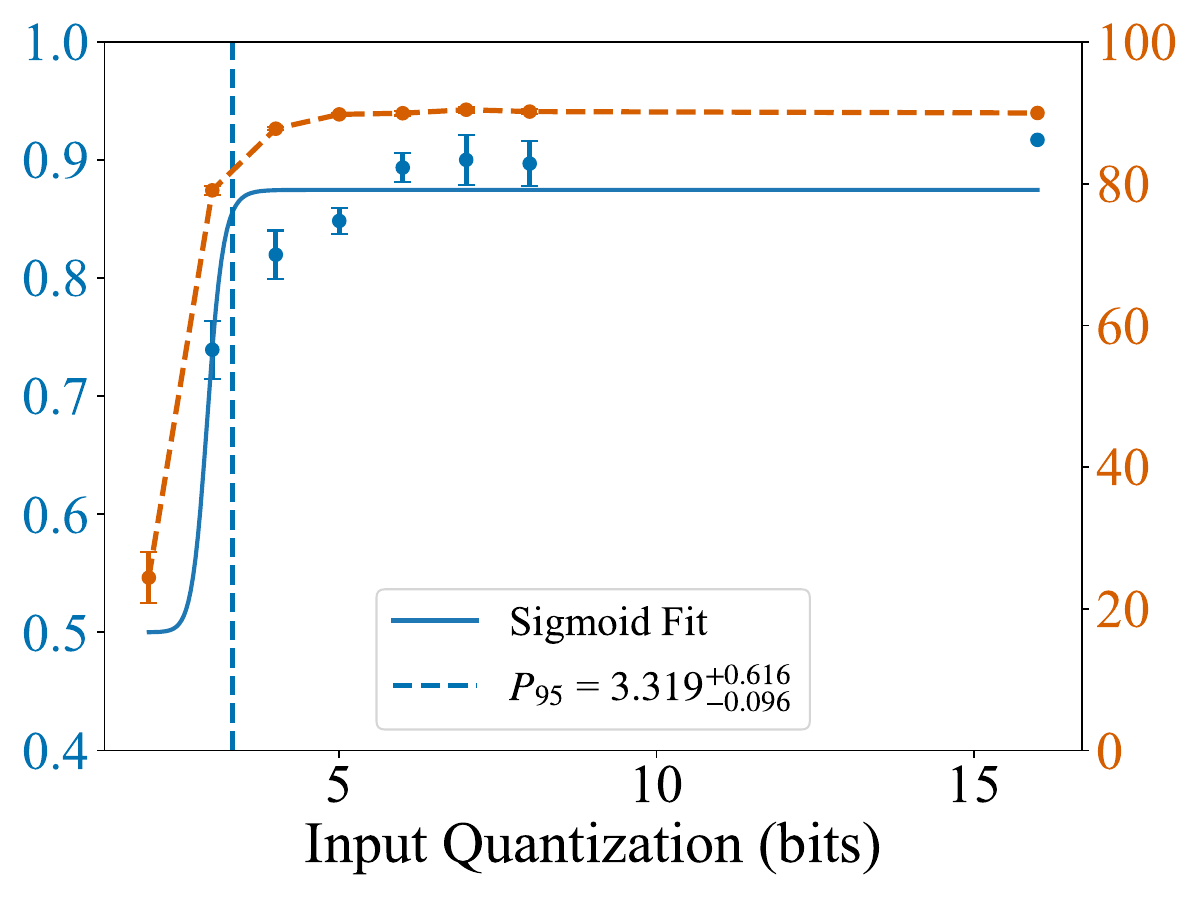}
        \includegraphics[width=\linewidth]
            {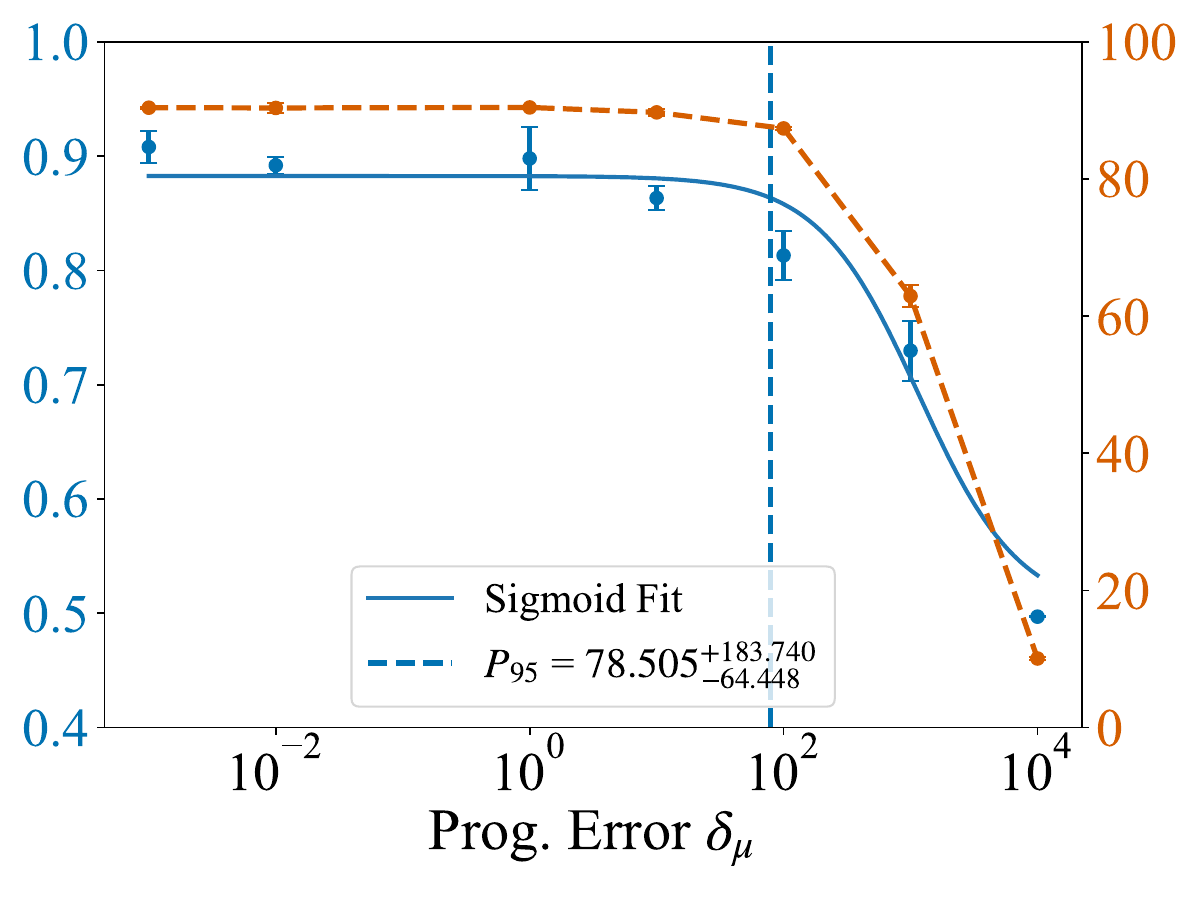}
        \includegraphics[width=\linewidth]
            {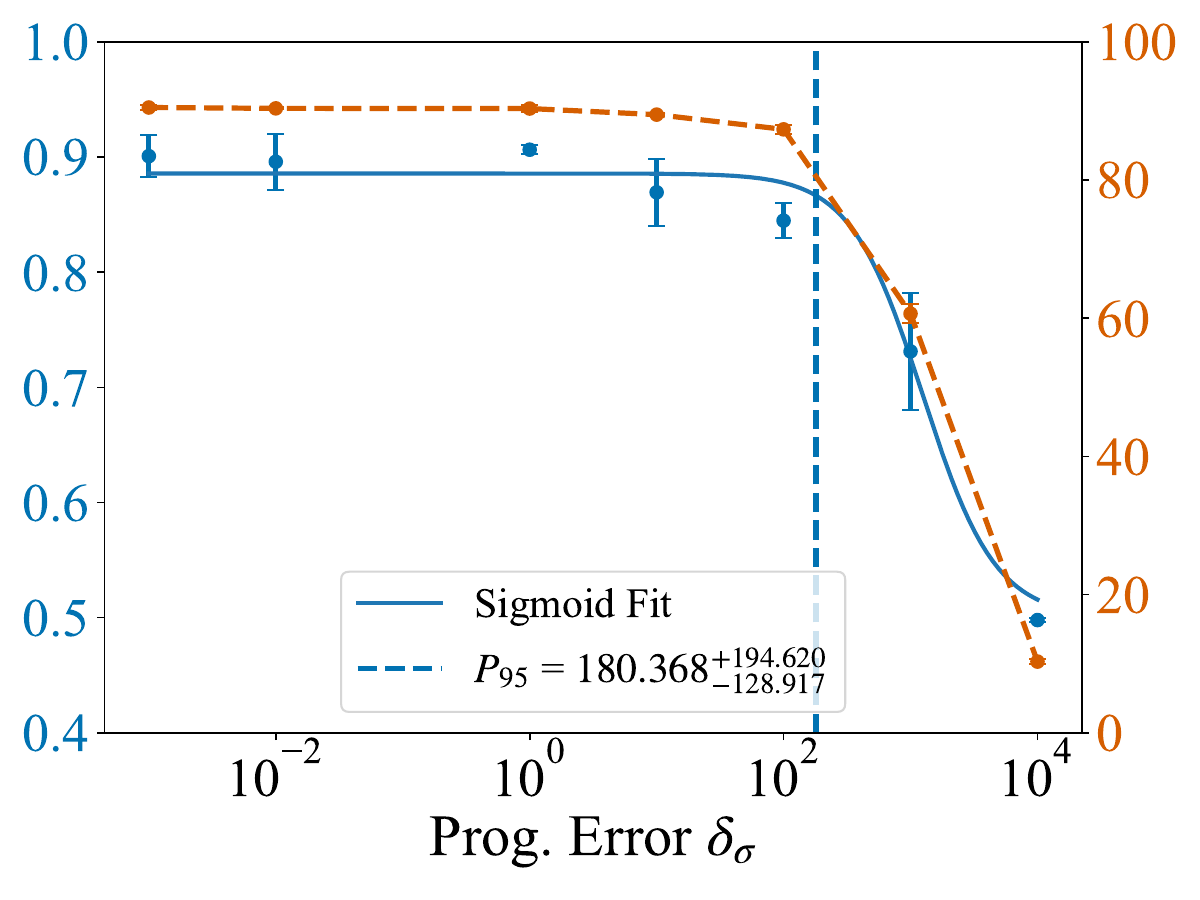}
        \includegraphics[width=\linewidth]
            {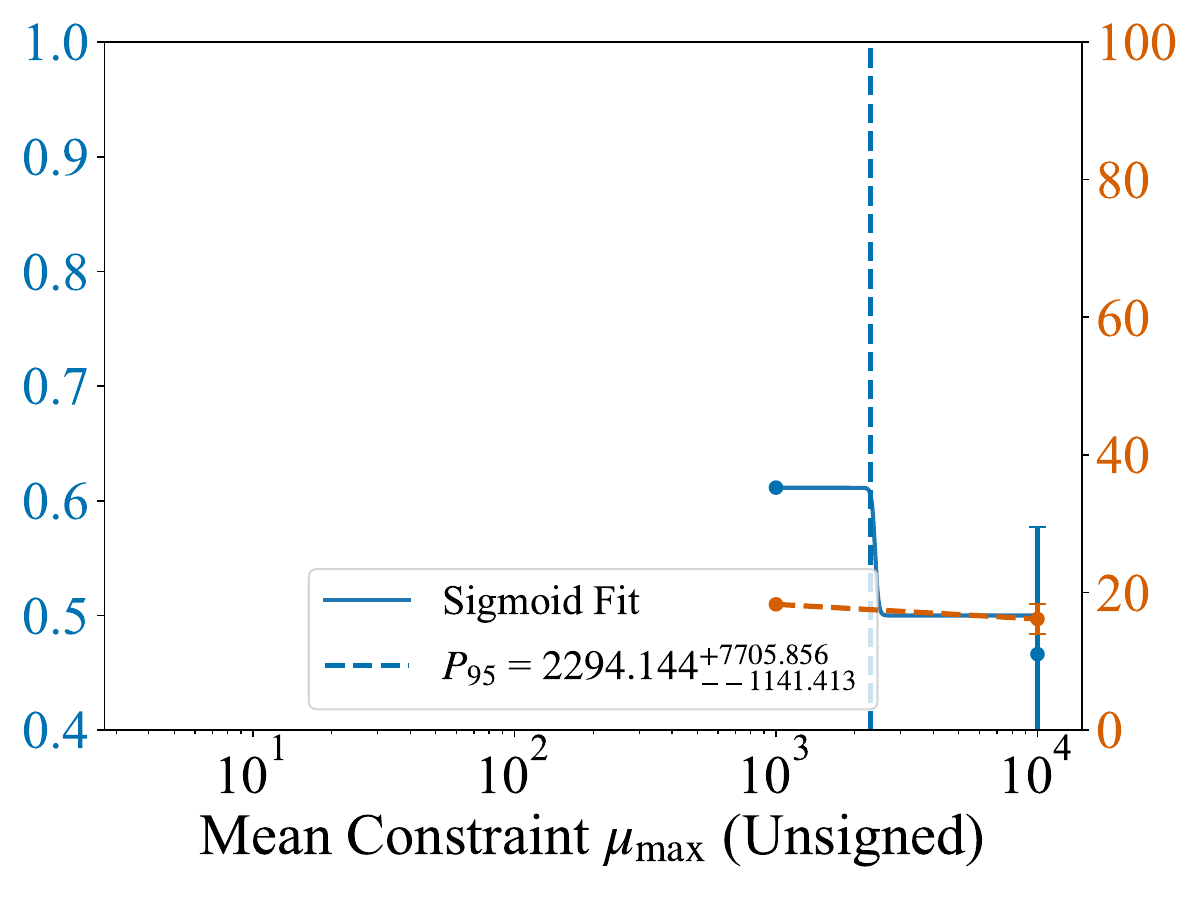}
        \includegraphics[width=\linewidth]
            {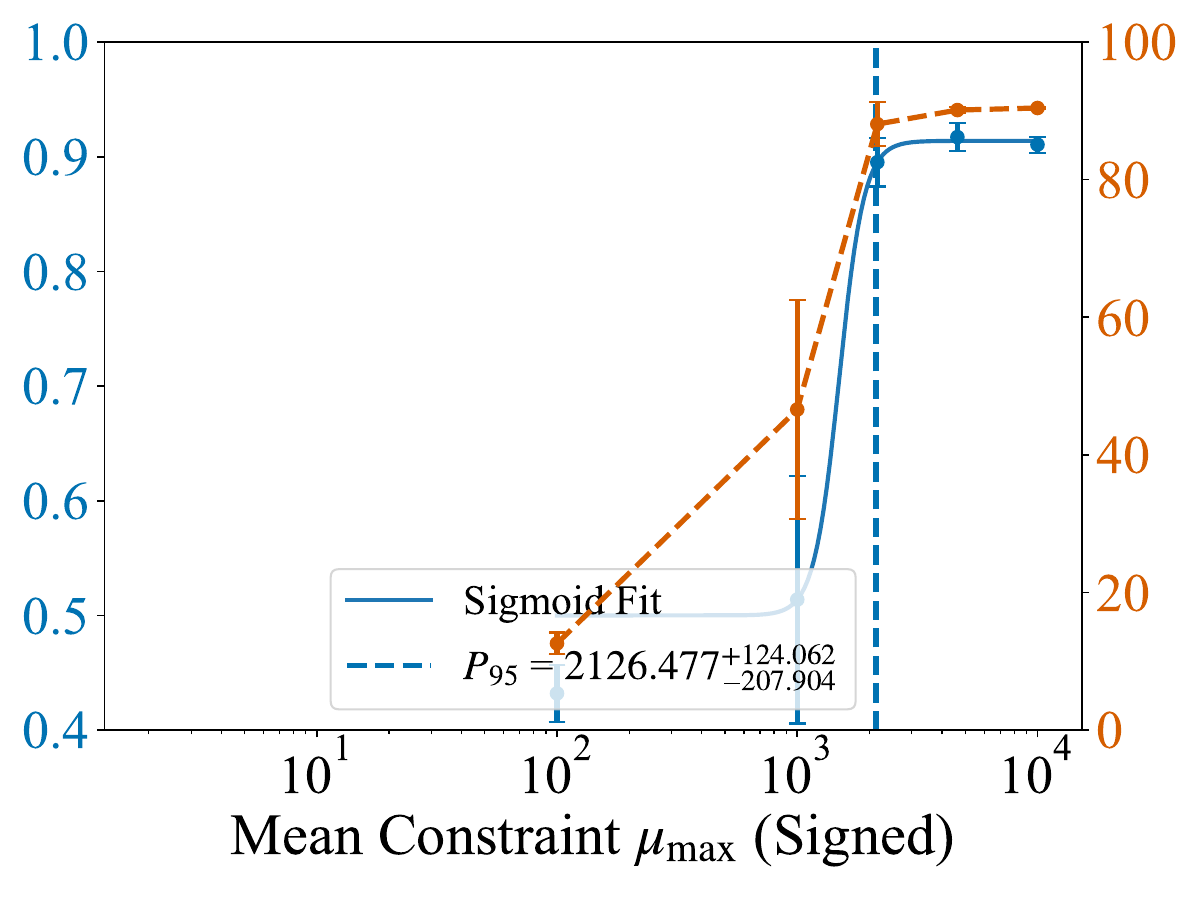}
        \includegraphics[width=\linewidth]
            {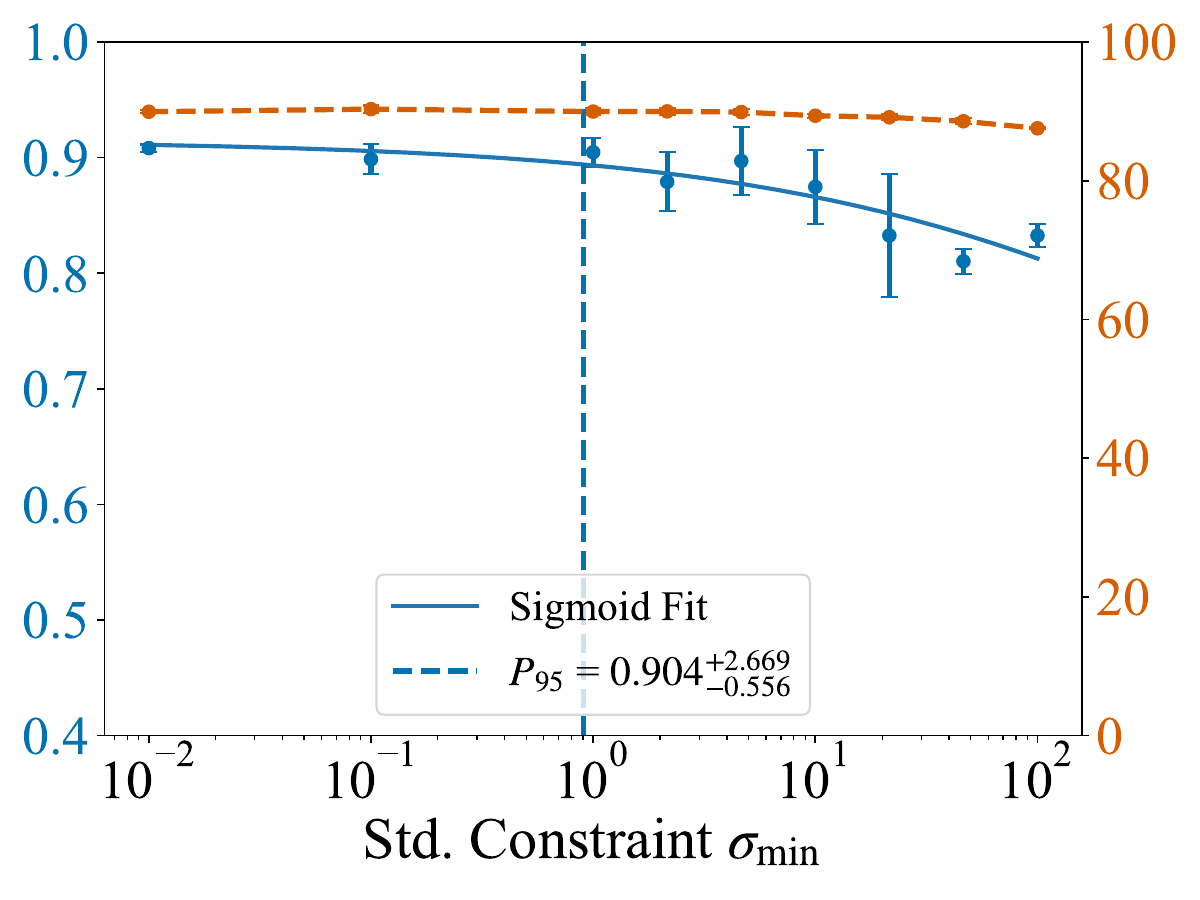}
        \includegraphics[width=\linewidth]
            {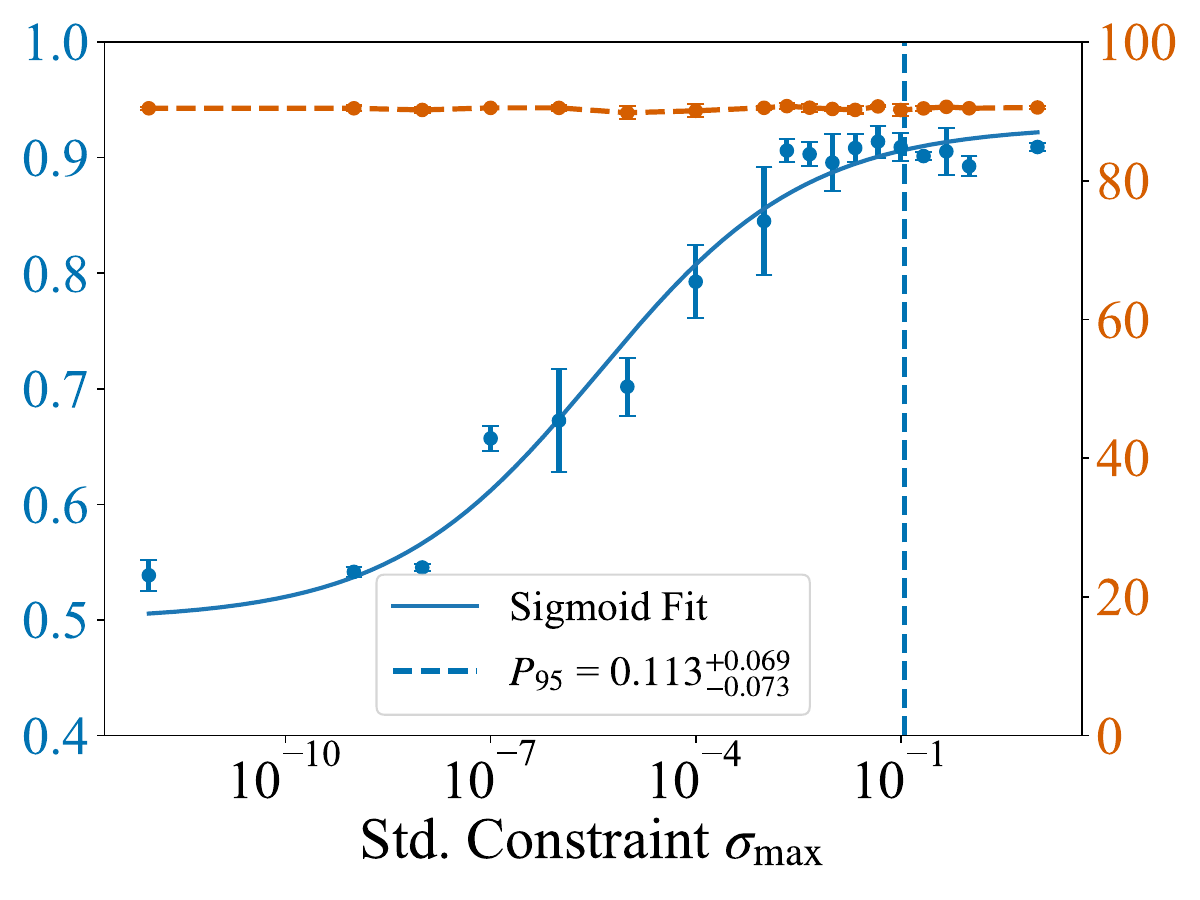}
    \end{subfigure}
    \hfill
    \begin{subfigure}[t]{0.235\textwidth}
        \centering
        \caption{A-Mul}
        \includegraphics[width=\linewidth]
            {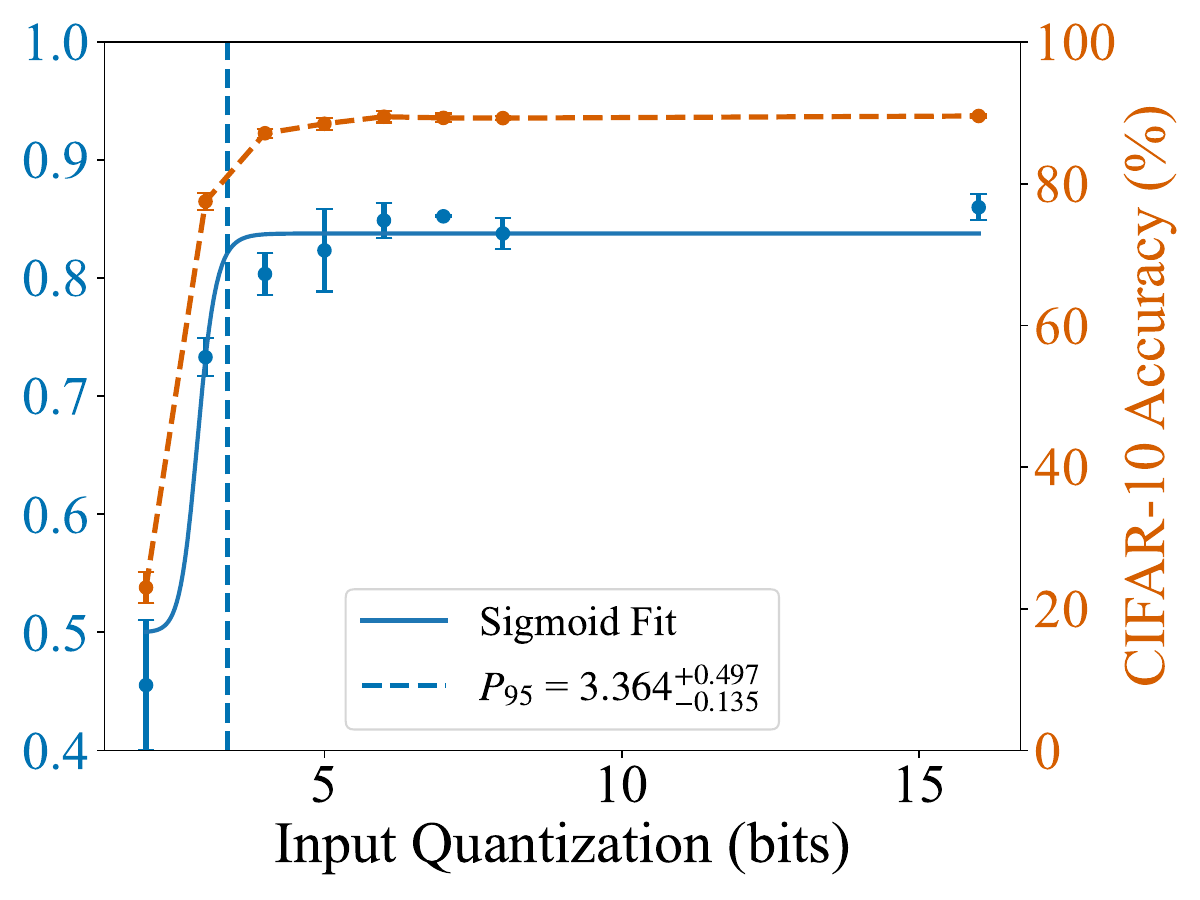}
        \includegraphics[width=\linewidth]
            {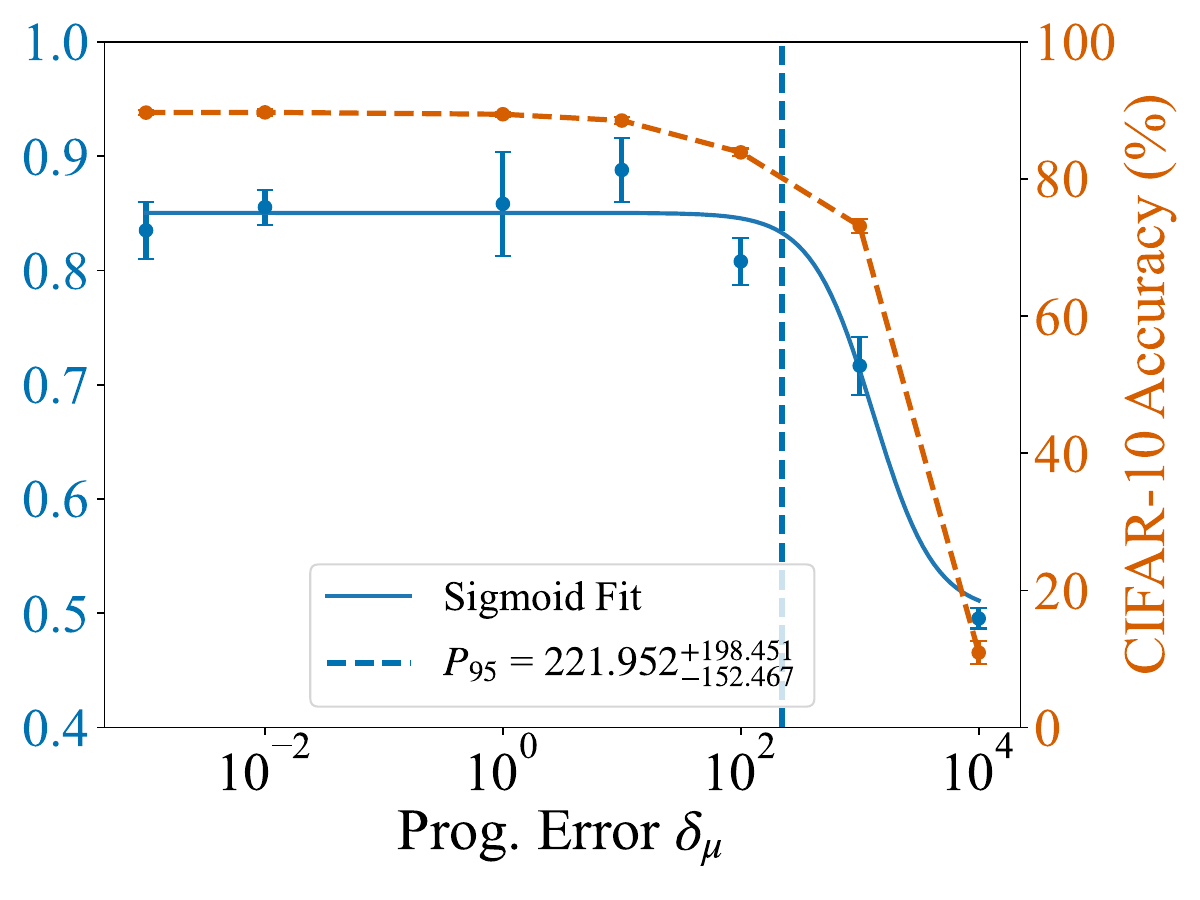}
        \includegraphics[width=\linewidth]
            {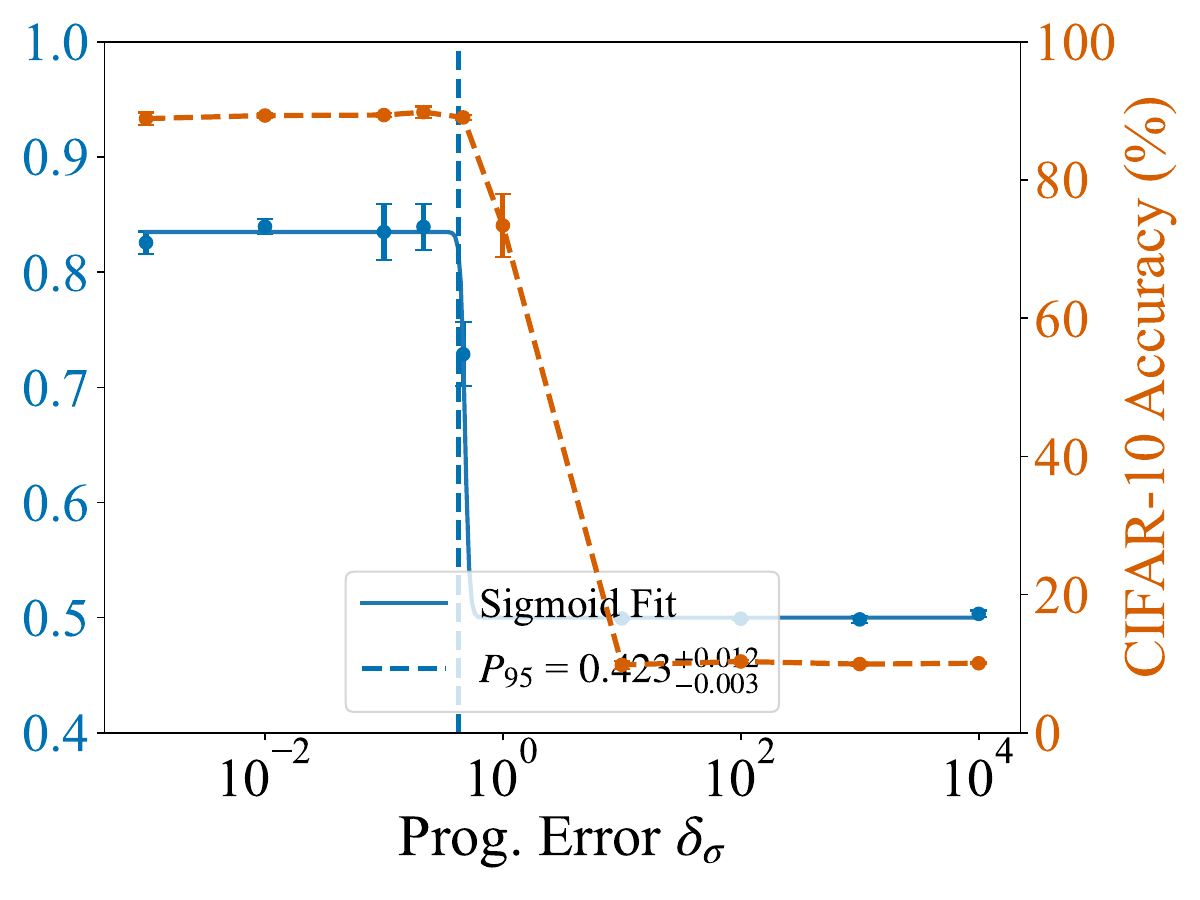}
        \AblationMissingPlot
        \includegraphics[width=\linewidth]
            {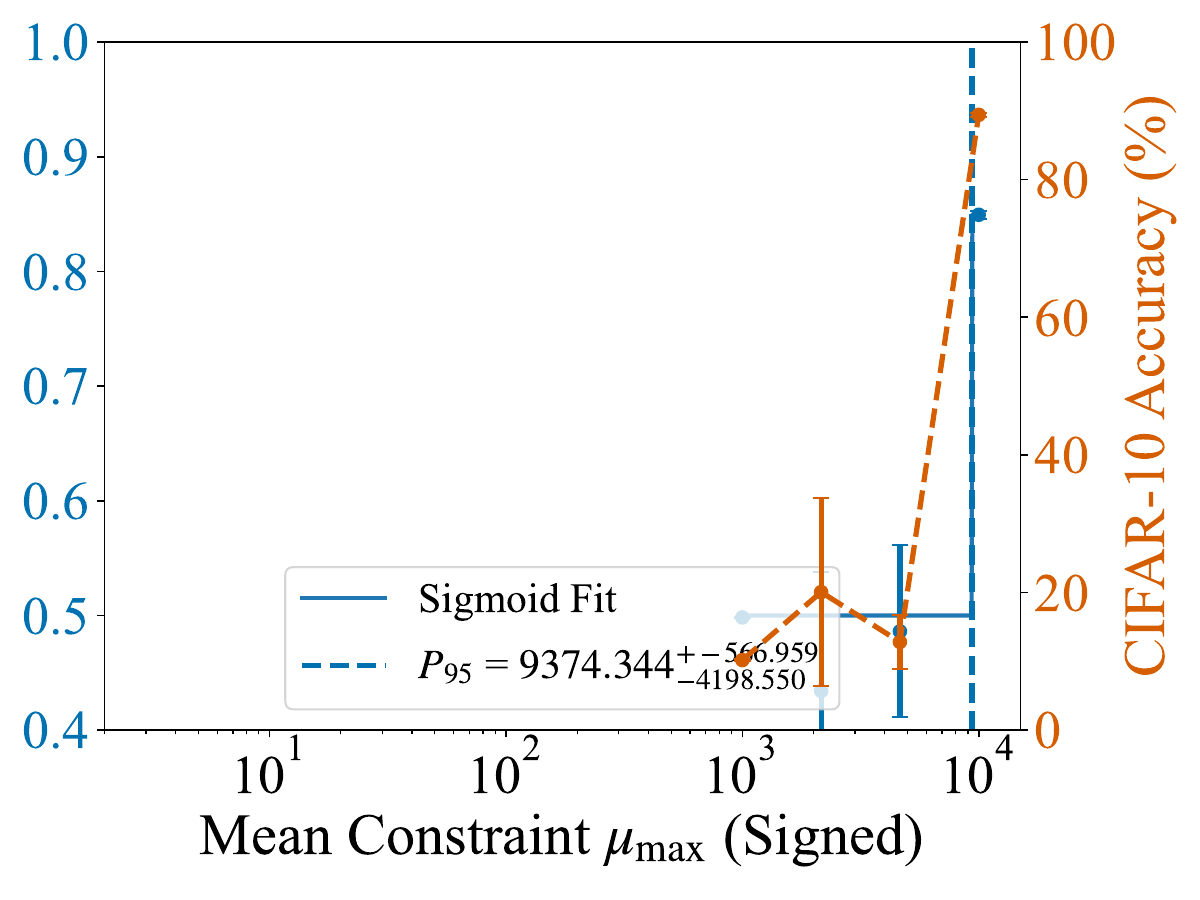}
        \includegraphics[width=\linewidth]
            {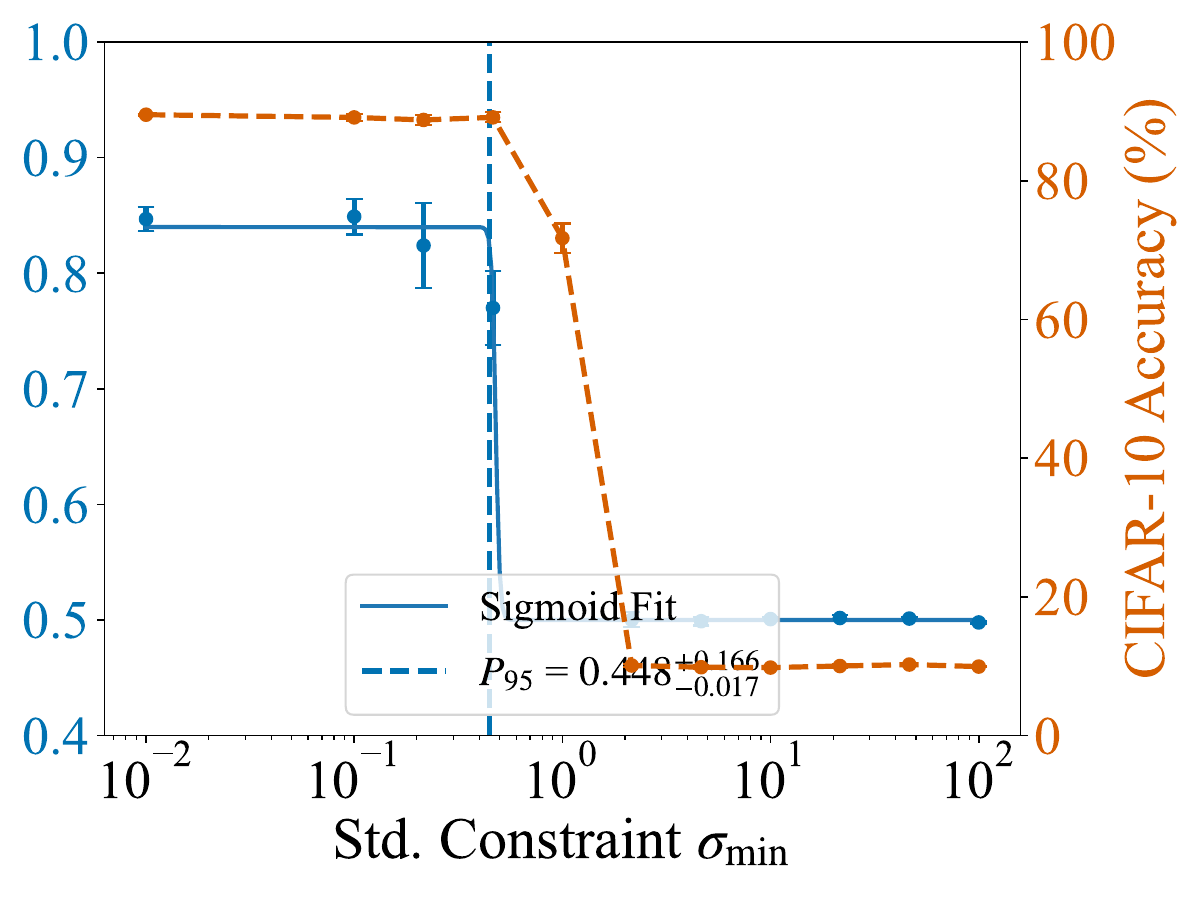}
        \includegraphics[width=\linewidth]
            {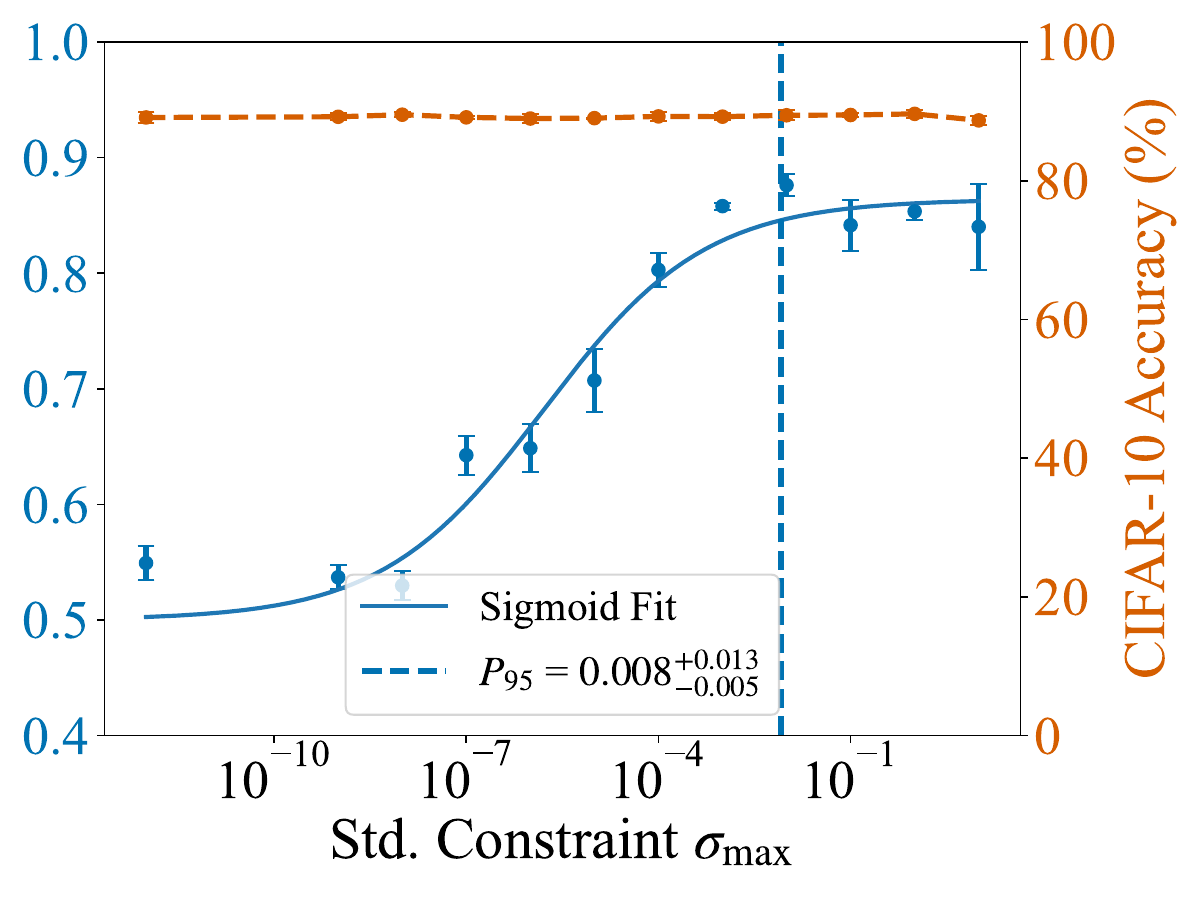}
    \end{subfigure}

    \caption{Ablation study of individual hardware constraints under
    hardware-aware training. Results are reported for ResNet-18 on CIFAR-10.}
    \label{fig:AB-cifar}
\end{figure*}

